# Translation of Black-Box Clinical Prediction Models into Standalone Transparent Nomograms: Temporal External Validation in Heart Transplantation

Henry Pigot[1], Paulo J.G. Lisboa[2], Sandra Ortega-Martorell[2], Ivan Olier[2], Joseph Mahon[2], Johan Nilsson[1,3*]

[1]Department of Translational Medicine, Artificial Intelligence and Bioinformatics in Cardiothoracic Sciences Research Unit, Lund University, Lund, Sweden

[2]Artificial Intelligence and Digital Technologies Research Institute, Liverpool John Moores University, Liverpool, UK

[3]Department of Thoracic and Vascular Surgery, Skåne University Hospital, Lund, Sweden

*Corresponding Author: Johan Nilsson, MD, PhD

Entrégatan 7, SE-22185 Lund, Sweden

E-mail: johan.nilsson@med.lu.se

## Summary

We convert black-box clinical prediction models for tabular data into standalone nomograms that can be audited term by term. PRiSM (Partial Responses in Structured Models) takes the shape of each effect and interaction from the source model, not merely which variables mattered, and lets the outcome select and weight them. We tested this in 50,356 heart transplant recipients, with validation in a later era than training. Nomograms from all 5 source models — a public clinical risk score, logistic regression, neural networks, random forests and extreme gradient boosting — met a prespecified noninferiority criterion for discrimination before any further simplification, and generally preserved calibration and clinical net benefit. Those from the 3 machine-learning models showed no detectable difference in discrimination from de novo generalized additive and explainable boosting models, exceeded neural additive models, and carried fewer terms than the explainable boosting model. PRiSM is released as an open-source Python package.

# Introduction

Clinical prediction models are increasingly used to support medical decision-making, including diagnosis, prognosis, treatment selection, and organ allocation[1-3]. However, for clinical decision support, predictive accuracy alone is not sufficient. Models must also be interpretable, well-calibrated, and transparent enough to permit a rigorous audit of their performance and the relationships they capture between predictors and outcomes[2,4-6].

In heart transplantation, decisions depend on balancing donor, recipient, and perioperative factors, as well as transplant timing[3,7,8]. Where such decisions rest on a model's output, transparency means being able to check what the model has learned about each factor before acting on its estimate[9]. Yet achieving reliable predictions remains challenging. Even a fundamental outcome such as 1-year mortality is difficult to predict, a difficulty further exacerbated by temporal changes in recipient profiles and treatment patterns[10,11].

Conventional interpretable methods, such as logistic regression, provide transparency but generally rely on linear relationships on the logit scale and require interactions to be specified in advance[12,13]. In contrast, more complex black-box models, including tree ensembles and neural networks, can automatically learn nonlinear clinical patterns and interactions from the data, but this advantage comes at the expense of reduced explainability and limited auditability[14-16].

Existing approaches differ in what they carry over from a trained model: an explanation of it, its predicted values, or only its ranking of variables. Post hoc methods, such as Shapley additive explanations (SHAP) and local interpretable model-agnostic explanations (LIME), and effect visualizations, such as partial-dependence and accumulated-local-effects plots, describe a model that remains in production and generate no risk estimates of their own[17-21]. Distillation fits a transparent surrogate to the black box's predictions, so what the surrogate reports answers to the model rather than to the outcome; the surrogates range from an additive model of main effects and pairwise interactions[22] to a neural additive network[23] to per-feature point scores read from SHAP attributions[24]. Interpretable tree approximations return a deployable model but apply only to decision forests, and they spread a variable's effect across branches instead of presenting it as a readable response[25]. Intrinsically interpretable models — generalized additive models, explainable boosting machines, neural additive models, component-selection methods, and rule ensembles — are transparent by construction, but are ordinarily fitted de novo to the outcome, so they carry over neither what a deployed model has learned nor the validation behind it[26-32]. Fitted instead to a black box's predictions, they become distillation surrogates and answer to the model again. In none of these does the observed outcome select or reweight effects transferred from the source model.

A recurring approach in recent clinical work is the machine-learning nomogram. The nomogram itself is a well-established graphical tool that calculates a risk estimate by summing points assigned to predictor values, and these studies do yield a deployable one but the trained model serves only as an instrument of variable selection. A gradient-boosted ensemble is fitted and its predictors ranked by SHAP attribution, and a fresh multivariable logistic regression, linear on the logit and without interactions, is then fitted to the outcome and drawn as the nomogram[33,34]. Where a predictor's learned shape is consulted at all, it is reduced to a cut-point read off the source model's partial-dependence plot, about which the continuous predictor is then dichotomized[33].

Whether a trained clinical black-box model can be mapped directly into a standalone transparent nomogram, and that nomogram then refined into a shorter one, without sacrificing performance under temporal external validation, remains an open question. We therefore applied PRiSM (Partial Responses in Structured Models), a model-agnostic framework that converts probabilistic black-box classifiers for any tabular data into standalone nomograms based on main effects and pairwise interactions[35-37].

In this framework the source model's predictions are written as a sum of partial responses, each a potentially nonlinear effect of one predictor or of a pair. The partial responses are read from the source model itself, by separating its behavior into one- and two-variable effects (a functional analysis of variance, or functional ANOVA), using the same constructions as partial-dependence and individual-conditional-expectation plots[19,38-40]; no intermediate surrogate is fitted whose own fidelity would then have to be established. These responses become the new predictors, and a sparse generalized additive model fitted to the observed outcome selects and weights them[26]. What transfers is the shape of each effect the source model learned, while the outcome decides which of those effects survive and how heavily they count. Because the prediction is then a plain sum of partial-response values, it can be deployed as a standalone nomogram; and because only the model's input-output behavior is used, the same procedure applies to any probabilistic classifier, including a published risk score that can no longer be retrained. The decomposition is exact for source models with no higher-order structure, so its fidelity can be checked against a known answer: logistic regression yields only univariate effects, and the published Index for Mortality Prediction After Cardiac Transplantation (IMPACT) score yields the thresholds that define it. An optional second step relearns the retained shapes from the outcome, so that a heavily simplified nomogram need not inherit shapes read from a source model that its few terms no longer represent; what it carries over is then the scaffold of selected variables and interactions, not the shapes themselves.

We assessed PRiSM across 5 model families for prediction of 1-year mortality after heart transplantation, using a prespecified noninferiority criterion together with calibration and decision-curve analysis; all 5 nomograms met the criterion in temporal external validation before any further simplification. We release the framework as an open-source Python package so that the conversion can be applied to other models and datasets[41].

## Methods

We performed a temporal external-validation study of PRiSM for prediction of 1-year mortality after heart transplantation. PRiSM was applied in two phases: initial translation of the source model into a transparent nomogram, followed by an optional refinement phase using a structured-network model, the Partial Response Network (PRN). The primary analyses evaluated maintenance of source-model performance after translation, whereas sparse-model analyses examined how far the translated nomograms could be simplified, what the refinement phase contributed, and whether performance relative to the source model was maintained. Five source models were evaluated: IMPACT, logistic regression (LR), multilayer perceptron (MLP), random forest (RF), and XGBoost (XGB). Three de novo additive comparators — generalized additive model (GAM), neural additive model (NAM), and explainable boosting machine (EBM) — were trained on the same temporal split[26,28,29]. Further details regarding the methods can be found in the supplemental information.

**Study design and population**

We evaluated PRiSM for prediction of 1-year post-transplant mortality using national registry data from the Scientific Registry of Transplant Recipients (SRTR). Adults (≥18 years) undergoing isolated heart transplantation during 2000–2022 were eligible. Data were extracted from the SRTR on June 30, 2024. Patients with follow-up shorter than 365 days and no recorded death within that interval were excluded to ensure complete ascertainment of the primary outcome, defined as all-cause mortality within 365 days of transplantation; this exclusion applied to 1,081 patients, 2.1% of those otherwise eligible. The final cohort comprised 50,356 recipients.

Records were ordered by transplant year and allocated to 3 nonoverlapping cohorts: training (2000–2016), tuning (2017–2019), and external validation (2020–2022); baseline characteristics of the 3 cohorts are given in Table 1. The training cohort was used for model development, preprocessing, imputation-model fitting, and PRiSM decomposition; the tuning cohort for hyperparameter selection; and the temporal external-validation cohort for final evaluation only. The study design, model development, and validation were conducted in accordance with the TRIPOD+AI checklist (Table S1)[4]. Complete exclusion criteria are provided in Table S2. The study was approved by the Swedish Ethical Review Authority (2016/987) and adhered to the Ethics Statement of the International Society for Heart and Lung Transplantation. Because the study is a secondary analysis of de-identified registry data with no patient contact, individual informed consent was not obtained. SRTR data were de-identified before release to investigators[42]. Recipient sex was the value recorded in the registry, a single binary field on the Organ Procurement and Transplantation Network (OPTN) transplant registration forms; the registry does not document whether it reflects sex assigned at birth or current sex, nor whether it was self-reported or entered by center staff, and it records no gender identity.

**Predictor selection and preprocessing**

Predictors were selected through expert clinical review of 1,293 available registry variables, followed by ranking of the 56 resulting candidates by mean absolute SHAP value in the training cohort; the 22 highest-ranked predictors were retained, and the final predictor set was fixed before model fitting. Continuous predictors were centered and scaled, categorical variables were dummy encoded, and missing values were imputed with MissForest using a training-cohort imputation model applied unchanged to later cohorts to avoid information leakage[43]. A leakage-audit and additional selection, preprocessing, and missing data details are provided in Document S1, supplemental methods.

**Source model development**

IMPACT was applied as a pretrained clinical score; the other 4 source models were trained on the training cohort, with hyperparameters optimized on the tuning cohort to maximize the area under the receiver-operating-characteristic curve (AUROC)[12,14-16,44]. Discrimination alone was used as the tuning objective because it is the primary endpoint; calibration was assessed as a secondary endpoint without recalibration, so tuning the source models for calibration would have confounded that comparison. Implementation details, hyperparameter spaces, and model-specific tuning procedures are provided in Document S1 and Tables S3-S8.

**PRiSM methodology**

PRiSM (Partial Responses in Structured Models; Figure 1) is a model-agnostic framework that operates on the input-output behavior of a trained probabilistic binary classifier for tabular data[35]. Rather than assigning feature attributions to individual predictions, PRiSM decomposes the trained source model itself into additive partial responses and assembles the selected responses into a standalone nomogram.

In Phase I, the trained source model was decomposed by functional ANOVA into univariate main effects and bivariate interactions on the log-odds scale[38]. Each partial response was averaged over the observed distribution of the remaining predictors rather than fixing them at reference values, so that it was evaluated across the predictor combinations seen in the data (the Lebesgue method; Document S1). Only univariate and bivariate terms were retained, following the main-effects-plus-selected-pairwise-interactions design established for intelligible additive models[27]; a PRiSM nomogram therefore represents what the source model does with one or two variables at a time, and not what it does with three or more variables jointly. The least absolute shrinkage and selection operator (LASSO) was used to select and weight predictively informative partial responses. The decomposition uses only the source model's input-output behavior and is independent of the outcome; within Phase I, the outcome enters only at LASSO selection and weighting. Supervising selection on the observed outcome rather than on the source model's predictions distinguishes PRiSM from knowledge distillation: the aim is not to reproduce the black box but to retain the effects that predict the observed outcome, and to leave the remaining higher-order structure aside (Document S1). The final nomogram was defined as the intercept plus the weighted sum of the selected partial responses and was intended as a standalone transparent predictor, not an exact reproduction of the source model's probability estimates.

A term is a single selected partial response, either a univariate main effect or a bivariate interaction. A categorical predictor counts as one variable throughout, contributing one term rather than one per level, so selection ranged over 22 univariate and 231 bivariate candidates for the 4 trained source models, and over 12 and 66 for IMPACT, which was applied with its own published predictor set. Partial responses whose fitted coefficient was 0.1 or less in absolute value on the log-odds scale were treated as unselected, a noise filter fixed before model evaluation rather than a tuned hyperparameter (Document S1).

Two LASSO strategies were used. The penalty, lambda, sets how aggressively terms are removed: the stronger the penalty, the fewer terms the nomogram carries. The baseline strategy selected a value of lambda that preserved at least 99.8% of the maximum tuning-cohort AUROC while minimizing the number of selected terms, referred to throughout as the baseline lambda. The sparse strategy instead selected lambda to produce the smallest Phase I nomogram (i.e., with the fewest terms) whose tuning-cohort retention contrast was nonnegative, that is, whose tuning-cohort AUROC reached $0.90 \times \text{AUROC}_{\text{source,tuning}} + 0.05$; this value is referred to throughout as the sparse lambda. This is the criterion defined below for the primary endpoint, evaluated on the tuning cohort. No external-validation quantity was used in lambda selection. Full LASSO and decomposition configurations are provided in Table S9.

In Phase II, an optional refinement re-estimated the selected responses in a model that admits no interactions above second order. The Phase I responses are read from a source model whose behavior also reflects those higher-order interactions; refitting the same terms in a model that cannot represent them removes much of that residual influence, and precludes overfitting through unconstrained interaction complexity. Partial responses selected in Phase I defined the architecture of a structured neural network, the PRN[35], in which each subnetwork receives only one or two variables corresponding to a selected effect. The PRN therefore inherits from Phase I only which effects are permitted — the scaffold of selected variables and interactions — and relearns their response shapes from the observed outcome. The PRN was trained on the observed outcome using the same training data as the source model, after which functional ANOVA decomposition was again applied, followed by LASSO selection to generate PRN-refined nomograms. Additional implementation details are provided in Document S1 and Tables S10 and S11.

Each nomogram is named by two choices: Phase I alone or Phase I followed by Phase II refinement, and the lambda used for Phase I selection, baseline or sparse. All 4 combinations were produced for each of the 5 source models. Two carry the analyses reported here: the Phase I nomogram at the baseline lambda, termed the baseline PRiSM nomogram, and the Phase II nomogram at the sparse lambda, termed the sparse PRN nomogram; these bracket the trade between fidelity to the source model and parsimony. The sparse PRiSM nomogram is reported alongside them as the comparator from which each sparse PRN nomogram was refined (Table 2). The fourth combination, the Phase II nomogram at the baseline lambda, is not reported here, since it neither carries the source model's learned shapes, which Phase II relearns, nor is the shortest nomogram available; it nonetheless met the noninferiority criterion for all 5 source models, at term counts at or below those of the corresponding baseline PRiSM nomogram (Tables S12-S14).

**Primary and secondary endpoints**

The primary endpoint was noninferiority in temporal external-validation AUROC. The criterion was predefined as preservation of at least 90% of the source model's discrimination above chance, a percent-retention criterion of the type the FDA noninferiority guidance designates M2, following its synthesis approach[45]. For each nomogram the retention-adjusted contrast was T = AUROC_nomogram - 0.90 × AUROC_source - 0.05, which exceeds zero exactly when more than 90% of the source model's discrimination above chance is retained. Noninferiority was concluded if the lower bound of the two-sided 95% confidence interval for T exceeded zero. T is the quantity Table 2 reports; the difference in AUROC between the nomogram and its source model (nomogram minus source model) is a descriptive quantity reported in Table S14. The confidence-interval requirement applies to this external-validation evaluation only; the sparse lambda selection described above applied the same retention criterion to tuning-cohort point estimates. Secondary endpoints were calibration slope, calibration-in-the-large, and observed-to-expected ratio[46]. Decision-curve analysis quantified net benefit across threshold probabilities of 5% to 20%[47]. Confidence intervals for AUROC were estimated with the DeLong method, and those for the difference in AUROC and for T from the same paired DeLong covariance matrix, which preserves the pairing of patients evaluated by both a source model and its nomogram; those for the calibration slope and calibration-in-the-large with the Wald method; and those for the observed-to-expected ratio with a seeded percentile bootstrap over 1,000 resamples. Models were evaluated on the temporal external-validation cohort without recalibration. Model training and PRiSM conversion were implemented in Python 3.12. Statistical evaluation was performed in R version 4.5.2. Additional statistical analyses, software implementation details, data and code availability, and methodological considerations are provided in Document S1.

**De novo additive comparators**

Translated nomograms were compared with 3 additive models trained de novo on the same temporal split: GAM, NAM, and EBM[26,28,29]. As with the source models, comparator hyperparameters were selected to maximize the tuning-cohort AUROC. Discrimination was compared in the external-validation cohort as the difference in AUROC between each nomogram and each comparator (nomogram minus comparator), using paired DeLong tests. No equivalence margin was set for these comparisons, so they can detect a difference but cannot establish equivalence. Implementation details are provided in Document S1.

**Sensitivity and subgroup analyses**

Two sensitivity analyses were performed. The conversion was repeated under 3 random seeds with the source models and the imputed data held fixed, which varied only the stochastic steps of the pipeline and not the decomposition itself. The conversion was also repeated with transplant year added to the predictor set, at the same seed for both configurations. Both analyses are described in Document S1.

As exploratory analyses, performance was examined within subgroups of the external-validation cohort defined by recipient sex and by race or ethnic group (White, Black, Hispanic, Other); neither variable is among the 22 predictors of the trained source models, although both appear in the published IMPACT definition. Discrimination was compared between each group and its reference group, calibration was compared across groups, and the effect of PRiSM translation on any between-group difference in AUROC was assessed. These analyses were not adjusted for multiplicity. Estimators and additional details are provided in Document S1.

## Results

### Cohort characteristics

The final cohort comprised 50,356 heart transplant recipients partitioned chronologically into training (2000–2016; $n$ = 32,543), tuning (2017–2019; $n$ = 8,558), and external validation (2020–2022; $n$ = 9,255) cohorts. Overall, 5,042 patients died within 365 days, a 1-year mortality of 10.0% (95% CI, 9.8 to 10.3); deaths numbered 3,451 (10.6%) in the training cohort, 732 (8.6%) in the tuning cohort, and 859 (9.3%) in the external-validation cohort. Most baseline characteristics were stable across cohorts, but the external-validation cohort showed a temporal shift after the 2018 United Network for Organ Sharing (UNOS) allocation-policy revision, with higher recipient acuity and greater use of mechanical circulatory support at transplantation: pretransplant ICU admission increased from 29.5% in the training cohort to 55.1% in the validation cohort, intraaortic balloon pump use from 5.7% to 27.9%, and extracorporeal membrane oxygenation from 0.6% to 6.1% (Table 1). Additional cohort characteristics and temporal between-cohort differences are provided in Table S15.

### Predictive noninferiority and preservation of source model performance

The 5 source models showed modest discrimination for 1-year mortality after heart transplantation in temporal external validation, with AUROC values ranging from 0.619 to 0.646 (Table 2); these are the reference values every nomogram below is measured against. The main finding was that baseline PRiSM nomograms preserved the predictive performance of their source models (Figure 2A). All 5 baseline nomograms met the prespecified noninferiority criterion in temporal external validation, retaining more than 90% of their source model's discrimination above chance (Table 2). These nomograms carried 9 terms for IMPACT, 21 for LR, 22 for RF, 26 for XGB, and 30 for MLP (Table 2). The IMPACT and LR nomograms were entirely univariate, whereas the nomograms of the trained nonlinear models retained pairwise interactions as well: 20 univariate and 10 bivariate terms for MLP, 14 and 8 for RF, and 19 and 7 for XGB. The retention contrast T ranged from +0.012 for XGB to +0.023 for IMPACT, and the smallest lower confidence bound across the 5 nomograms was +0.006, for RF (Table 2). Descriptively, discrimination changed little in absolute terms: the differences in AUROC ranged from -0.003 for XGB to +0.011 for IMPACT (Table S14). Calibration was generally preserved after PRiSM conversion, with several models showing improved calibration metrics (Table 2). Decision-curve analysis showed similar net benefit for the source models and the baseline PRiSM nomograms across threshold probabilities of 5% to 20% (Figure 2B-C). The calibration and decision-curve comparisons are descriptive; no formal between-model testing was performed for these secondary endpoints.

**Sparse PRN simplification**

Results for the sparse PRN models were more model dependent than those for the baseline PRiSM nomograms. Sparse PRN nomograms contained 5 to 11 terms, the same counts as the sparse PRiSM nomograms they were refined from; the Phase II selection removed none of the Phase I terms, so refinement changed the shapes of the retained effects and not which effects were retained. The sparse Phase I selection was univariate for all 5 source models, so every sparse PRiSM and sparse PRN nomogram inherited a scaffold containing no interactions. In the validation cohort, the prespecified noninferiority criterion was met for LR and XGB, but not for IMPACT, MLP, or RF (Table 2).

The relevant comparator for these models is the sparse PRiSM nomogram from which each was refined: none of the 5 met the noninferiority criterion in external validation (Table 2). Phase II refinement therefore recovered noninferiority that sparsification of the Phase I selection had lost, for 2 of 5 source models. Of the 3 sparse PRN nomograms that did not meet the criterion, RF alone lost retention outright, with T at -0.004; IMPACT and MLP both retained more than 90% of their source model's discrimination above chance on the point estimate, at T = +0.013 and +0.011, but with confidence intervals too wide to exclude zero (Table 2). Decision-curve analysis for the sparse PRN nomograms showed net benefit similar to that of the source models across threshold probabilities of 5% to 20% (Figure 2D).

The regularization paths show the mechanism behind these results (Figure 3). For MLP (Figure 3A), the univariate main effects were selected first and tuning-cohort discrimination plateaued shortly after, near the baseline penalty; weakening the penalty beyond it added only more bivariate interaction terms, which raised training AUROC without improving tuning AUROC and so widened the gap between them, the signature of overfitting. The LR path selected only univariate terms at every penalty, as a linear source model has no interaction structure to recover (Figure 3B). After Phase II refinement of the sparse Phase I selection, tuning AUROC was largely insensitive to the penalty and the gap between training and tuning AUROC stayed small, because the network is confined to the 1- and 2-variable effects fixed in Phase I and cannot reintroduce further interactions (Figure 3C, D). Full discrimination and calibration results for the source models and all reported nomograms are provided in Tables S12-S14 and S16.

**Recovery of known model structure**

LR produced linear response profiles, whereas MLP, XGB, and RF showed nonlinear responses (Figure 4A). For IMPACT, whose points are defined by published thresholds, the Phase I partial responses for total bilirubin and creatinine clearance stepped at those thresholds (bilirubin at 17.1, 34.2, and 68.4 umol/L and creatinine clearance at 30 and 50 mL/min; Figure 4A), while the categorical IMPACT predictors that survived selection were recovered as their defined groups (Figure S1A).

All 20 translated nomograms are available through an open-access interactive web interface. Complete sparse PRiSM and sparse PRN nomograms for all 5 source models are provided in Figures S1 and S2, the complete baseline PRiSM nomogram for MLP in Figure S3, and a bedside risk calculation for a synthetic patient using the sparse PRN MLP nomogram in Figure S4.

**Comparator analysis**

In the external-validation cohort, the 3 de novo additive comparators achieved AUROCs of 0.647 for EBM, 0.645 for GAM, and 0.615 for NAM. For the nomograms derived from MLP, XGB, and RF, differences against EBM and GAM had confidence intervals including zero at the baseline lambda, and all 3 exceeded NAM (Table 3). The sparse PRN nomograms did the same for MLP and XGB, at 10 and 11 terms, while the 8-term RF nomogram fell 0.023 AUROC (95% CI, -0.036 to -0.011) below EBM and 0.021 (95% CI, -0.033 to -0.009) below GAM and did not separate from NAM. At the baseline lambda, the nomograms derived from IMPACT and LR fell 0.015 to 0.022 AUROC below EBM and GAM and did not separate from NAM, consistent with the discrimination available in their source models. Over the same 22 clinical variables, the baseline PRiSM nomograms of the 4 trained source models carried 21 to 30 terms and their sparse PRN nomograms 8 to 11, against 42 terms for EBM, of which 20 were pairwise interactions, and 22 each for GAM and NAM (Tables 2 and 3). Full comparator results are provided in Tables S17-S19.

**Sensitivity and subgroup analyses**

Noninferiority conclusions were identical across all seeds and all 4 nomogram combinations (Table S20), and agreed for 19 of the 20 nomograms with and without transplant year as a predictor; the single exception was a sparse PRN nomogram whose confidence bound fell short of zero by less than 0.001 AUROC (Table S24). Full results for sensitivity analyses are provided in Document S1, supplemental methods, and Tables S20-S25.

In exploratory subgroup analyses of the external-validation cohort, discrimination was lower in Black and in Hispanic than in White recipients for every model evaluated, including the 5 source models and the 3 de novo additive comparators, by 0.030 to 0.069 AUROC in Black recipients and by 0.005 to 0.053 in Hispanic recipients (Table S26). PRiSM translation did not alter these differences: across every nomogram and group compared, the largest change was -0.031 AUROC (95% CI, -0.063 to +0.001), for the baseline PRiSM nomogram derived from IMPACT in Hispanic recipients, and no confidence interval excluded zero. Performance by recipient sex was similar throughout, differing by at most 0.014 AUROC in either direction. No nomogram showed heterogeneity in calibration across groups (Document S1).

## Discussion

In this temporal external-validation study, 5 source models trained or implemented using tabular clinical data were converted into standalone baseline PRiSM nomograms, all of which met the noninferiority criterion relative to their respective source models. Further simplification was constrained: no sparse PRiSM nomogram met the criterion in external validation, and Phase II PRN refinement recovered it for 2 of the 5 source models, with substantial reductions in the number of terms.

The contribution of this study is evidential and practical rather than methodological. The PRiSM framework has been described previously[35]; what had not been tested was whether its nomograms could stand in for a clinical black box under temporal external validation, and the framework had not been available as a packaged tool that others can install and apply. This study supplies both: the validation evidence reported here, and an open-source Python implementation released alongside it[41]. PRiSM carries a trained black-box classifier's learned nonlinear effects and pairwise interactions into a standalone transparent nomogram that can be inspected term by term and used directly to make predictions. It creates a candidate replacement predictor that can preserve the source model's predictive performance while making its risk logic legible, and without giving up feature attribution. The distinction between replacing a model and accounting for it decides whether the object under review is the deployed model itself or a separate account of it[5,6,21,48]. This is the position argued by Rudin, that high-stakes decisions should rest on models that are interpretable by design rather than on explanations layered over black boxes[49].

Discrimination was modest across all 5 source models (AUROC, 0.62 to 0.65), consistent with previously reported performance for prediction of 1-year mortality after heart transplantation. This likely reflects the absence of important biological, surgical, and institutional factors in the available data, as well as changes in treatment practices over time[11]. Modest discrimination of this kind is a recurring finding in clinical machine learning and a reason to temper expectations of predictive performance alone[50]. Where the predictive signal is limited, the case for transparency is correspondingly stronger: a risk estimate without an interpretable clinical structure should not be used uncritically for decisions with major consequences.

In a PRiSM nomogram, each effect enters as a separate non-overlapping term, univariate or bivariate, with an independently readable contribution, so clinicians can work through it one concept at a time, and a concept whose contribution appears biologically implausible can in principle be challenged and adjusted when clinically justified, as concept-bottleneck models allow by design[51]; term-level intervention was not implemented or evaluated here. Auditability at the level of individual terms is what work on interpretable and explainable machine learning has moved toward[37], and what current guidance on AI in health care treats as a prerequisite for reliable use[5,6,52]. In heart transplantation, where assessment of candidates and organ offers requires multidisciplinary judgment rather than automation, the relevant use is human-in-the-loop: a nomogram that can be read term by term gives the team something to examine together, in discussion and in teaching[7,8,53]. All 20 nomograms are published as an open-access interactive interface (https://aibcts.github.io/prism-nomogram/web_nomogram/index.html), where a reader can set predictor values and see each term's contribution to the points total and the predicted risk.

Heart-transplant prediction models may overestimate performance when temporal leakage is not addressed, and time-based validation gives more clinically realistic estimates than random cross-validation[10]; the temporal split used here follows that reasoning. The case-mix shift the validation cohort carries (Table 1) is therefore part of the test rather than a complication of it.

Translating the source models into baseline PRiSM nomograms also clarified their internal structure. The two known-structure controls behaved as expected, returning only univariate effects for LR (Figure 3B) and, for IMPACT, its published thresholds (Figure 4A) and categories (Figure S1A). By contrast, in the baseline nomograms of MLP, XGB, and RF, several continuous predictors contributed nonlinearly to risk, including creatinine clearance, total bilirubin, and ischemic time. The baseline PRiSM nomogram for the MLP also exposed selected pairwise interactions, including an interaction between pretransplant creatinine clearance and recipient clinical status at transplantation, suggesting that the risk contribution of renal dysfunction may be context- dependent. These model-derived patterns are best viewed as hypothesis- generating rather than causal. The baseline PRiSM nomograms retained a modest number of interactions (10 for MLP, 8 for RF, and 7 for XGB), whereas all 5 sparse PRN nomograms retained none. The regularization paths show why (Figure 3A): the baseline selection already sits at the point where discrimination plateaus, so the interactions beyond it add nothing on the tuning cohort. That few interactions survive simplification is informative about the source models, not a limitation of the translation. What sparsification buys is the amount to be worked through: the sparse PRN MLP nomogram presents 10 univariate terms, against 20 univariate and 10 bivariate in the baseline nomogram, the latter read jointly in two variables (Figures 4B and S3).

The two phases serve different ends. Phase I carries the source model's learned responses and is the representation in which its structure can be read and audited; at the baseline penalty it preserved discrimination for all 5 source models without relearning the shapes of the transferred partial-response functions. Phase II serves simplification. Under aggressive sparsification few Phase I responses are retained, and they are read from a source model whose higher-order behavior they do not represent. Re-estimating them in a network confined to the selected univariate and bivariate effects recovered noninferiority for 2 of the 5 source models. Both are candidates for deployment in the source model's place: the baseline PRiSM nomogram where auditable fidelity to the source model is the priority and noninferiority is required across model families, and the sparse PRN nomogram where parsimony matters most and the criterion is met for the family in question. The two are marked points on a single regularization path (Figure 3A) rather than a fixed pair: intermediate penalties give intermediate nomograms, so the number of terms can be chosen to suit the intended use.

The additive comparator models (EBM, GAM, and NAM), each tuned on the same cohort and against the same objective as the source models, discriminated in the same range as the source models and the nomograms translated from them; on this task, discrimination did not require a black-box model. The comparators reached that discrimination at a different size: EBM required 42 terms, against 30 for the baseline MLP nomogram and 10 for its sparse PRN refinement, with no detectable difference in discrimination from either (Table 3). These models, however, answer a different question: they provide alternatives trained de novo, but they do not translate or replace an already trained source model. The approaches that do begin from a trained model take only a variable ranking from it[33,34], or decompose a surrogate fitted to its predictions rather than the model itself[23]. The closest prior attempt to use both sources of information keeps them apart: an additive surrogate is fitted to the model's predictions, a second additive model is fitted independently to the observed outcome, and the two are compared to audit the black box[22]. PRiSM instead combines them in one deployable instrument, taking the response shapes from the model and the selection and weighting from the outcome, and compressing the source model's learned risk structure into a clinically manageable nomogram. This inverts the objective of distillation: the higher-order remainder discarded by second-order truncation is set aside rather than treated as error to be minimized.

Several limitations apply. Missing data were handled with single imputation, which does not propagate uncertainty[54]; serum albumin was 57.0% missing in the tuning cohort, so lambda selection depended in part on imputed values (detailed rationale provided in Document S1, supplemental methods). Sensitivity analyses showed consistent results across random seeds, but held the imputed data fixed. The framework is currently limited to tabular binary classification, and generalizability beyond the US SRTR registry requires further international validation[52]. Because each partial response is obtained by averaging the source model over the observed data, predictor combinations that occur rarely could in principle distort which terms are selected[38]; that the baseline nomograms retained their source models' discrimination under temporal external validation argues against a material effect here (Document S1). The lower discrimination in Black and in Hispanic than in White recipients appeared in every model evaluated, including the de novo comparators, and PRiSM translation left it unchanged. A difference this general points to the predictors and the cohort rather than to any modelling step. Sex was available only as a single binary registry field, so the subgroup analysis by sex could not distinguish sex from gender. The noninferiority criterion assessed fidelity to the source model rather than absolute clinical adequacy. Finally, while this retrospective study establishes technical and translational feasibility, future prospective testing is required to evaluate real-world workflow integration, user behavior, and effects on patient outcomes before clinical implementation is claimed.

Across 5 source-model families, every baseline PRiSM nomogram met the prespecified noninferiority criterion in temporal external validation of 1-year mortality after heart transplantation, and PRiSM is released here for the first time as an installable open-source package. Simplifying further traded discrimination for brevity, at a cost the regularization path makes visible before a nomogram is chosen. The baseline nomograms carry over the main effects and pairwise interactions the source model learned, and all nomograms are fitted to the observed outcome rather than to the source model's predictions. Whether a nomogram read term by term changes what a clinical team decides is a question that can now be asked of a specific, deployable model.

## Resource availability

### Lead contact

Requests for further information and resources should be directed to and will be fulfilled by the lead contact, Johan Nilsson (johan.nilsson@med.lu.se).

### Materials availability

This study did not generate new unique reagents.

**Data and code availability**

The SRTR data analyzed in this study were supplied by UNOS as the contractor for the OPTN under a data-use agreement and cannot be redistributed by the authors; access requires an application to SRTR. A synthetic dataset that emulates the SRTR schema is provided with the code for testing.

PRiSM is available as an open-source Python package (BSD 3-Clause) at https://github.com/AIBCTS/PRiSM, together with demonstration notebooks and the configurable pipeline used here; the version used in this study is archived[41]. All 20 nomograms generated in this study are available through an interactive web interface at https://aibcts.github.io/prism-nomogram/web_nomogram/index.html, where users can adjust predictor values and inspect their effects on points and predicted 1-year mortality risk. This interface is provided for transparency and exploration of model behavior; clinical implementation was not evaluated in this study.

Any additional information required to reanalyze the data reported in this paper is available from the lead contact upon request.

Also see Document S1, supplemental methods.

## Supplemental information

Document S1. Supplemental methods, Figures S1-S4, and Tables S1-S26

## Acknowledgments

The authors thank the United Network for Organ Sharing (UNOS) for access to the SRTR registry data. The data used in this research were supplied by UNOS as the contractor for the Organ Procurement and Transplantation Network (OPTN). The interpretation and reporting of these data are the responsibility of the authors and should in no way be seen as an official policy of, or interpretation by, the U.S. Government, the Department of Health and Human Services, the Health Resources and Services

Administration, UNOS, or the OPTN. Figure 1 was prepared in Apple Keynote, and Figures 2, 3, and 4 were assembled in Adobe Illustrator, with only cosmetic adjustments for readability, including layout, colors, and font size. These adjustments did not affect the data, estimates, confidence intervals, model outputs, or interpretation.

Financial support was provided by the Swedish Heart-Lung Foundation (20220591 and 20230545), the Swedish Research Council (2019-00487, 2022-00683, and 2023-03184), Skåne County Council (2022-Projekt0168), donation funds from Skåne University Hospital, Hans-Gabriel and Alice Trolle-Wachtmeister's Foundation for Medical Research, and Hjelms Family Foundation for Medical Research. The LMK Foundation partly funds Henry Pigot's post-doctoral position.

The funding agencies had no role in the design and conduct of the study; collection, management, analysis, or interpretation of the data; preparation, review, or approval of the manuscript; or the decision to submit the manuscript for publication.

## Author contributions

HP developed the Python codebase, wrote Document S1, and prepared the initial manuscript draft under JN's supervision. PL originated the PRiSM methodology in MATLAB, interpreted the results, and guided the software translation. JM performed an initial MATLAB-to-Python translation. SOM and IO contributed to the development of the original MATLAB PRiSM algorithm. JN conceived and led the study, acquired the project funding, and designed the research methodology. He developed the statistical analysis code (R, Stata), the predictor-selection code (Python), and the interactive web-based nomograms. He performed the formal statistical analysis, interpreted the results, and was responsible for the final drafting of the main manuscript, including the preparation of figures and tables. He provided overall project supervision. All authors reviewed and edited the manuscript and approved the final version.

## Declaration of interests

The authors declare no competing interests.

## Declaration of generative AI and AI-assisted technologies in the writing process

During the preparation of this work, the authors used OpenAI ChatGPT, Google Gemini, and Anthropic Claude to refine language, and support editorial consistency. After using these tools, the authors reviewed and edited the content as needed and take full responsibility for the content of the published article. No generative AI or AI-assisted technologies were used to generate primary data, define outcomes, perform statistical analyses, conduct independent inference, alter results, or generate figures.

## References

1. Aleksova, N., Alba, A.C., Molinero, V.M., Connolly, K., Orchanian-Cheff, A., Badiwala, M., Ross, H.J., and Posada, J.G.D. (2020). Risk prediction models for survival after heart transplantation: A systematic review. Am. J. Transplant. *20*, 1137-1151. 10.1111/ajt.15708.
2. van Smeden, M., Reitsma, J.B., Riley, R.D., Collins, G.S., and Moons, K.G. (2021). Clinical prediction models: diagnosis versus prognosis. J. Clin. Epidemiol. *132*, 142-145. 10.1016/j.jclinepi.2021.01.009.
3. Stehlik, J., Stevenson, L.W., Edwards, L.B., Crespo-Leiro, M.G., Delgado, J.F., Dorent, R., Frigerio, M., Macdonald, P., MacGowan, G.A., Nanni Costa, A., et al. (2014). Organ allocation around the world: insights from the ISHLT International Registry for Heart and Lung Transplantation. J. Heart Lung Transplant. *33*, 975-984. 10.1016/j.healun.2014.08.001.
4. Collins, G.S., Moons, K.G.M., Dhiman, P., Riley, R.D., Beam, A.L., Van Calster, B., Ghassemi, M., Liu, X., Reitsma, J.B., van Smeden, M., et al. (2024). TRIPOD+AI statement: updated guidance for reporting clinical prediction models that use regression or machine learning methods. BMJ *385*, e078378. 10.1136/bmj-2023-078378.
5. Haug, C.J., and Harrison, E.M. (2026). Which Human-in-the-Loop? Why Context, Culture, and Health Systems Matter. NEJM AI *3*. 10.1056/AIe2600084.
6. U.S. Food and Drug Administration (2024). Transparency for machine learning-enabled medical devices: guiding principles. Jun. https://www.fda.gov/media/179269/download.
7. Jain, R., Kransdorf, E.P., Cowger, J., Jeevanandam, V., and Kobashigawa, J.A. (2025). Donor Selection for Heart Transplantation in 2025. JACC Heart Fail. *13*, 389-401. 10.1016/j.jchf.2024.09.016.
8. Peled, Y., Ducharme, A., Kittleson, M., Bansal, N., Stehlik, J., Amdani, S., Saeed, D., Cheng, R., Clarke, B., Dobbels, F., et al. (2024). International Society for Heart and Lung Transplantation Guidelines for the Evaluation and Care of Cardiac Transplant Candidates-2024. J. Heart Lung Transplant. *43*, 1529-1628 e1554. 10.1016/j.healun.2024.05.010.
9. Reis, B.Y., and Cava, W.G.L. (2026). Design principles for integrated AI alignment. Patterns *7*. 10.1016/j.patter.2026.101587.
10. Miller, R.J.H., Sabovčik, F., Cauwenberghs, N., Vens, C., Khush, K.K., Heidenreich, P.A., Haddad, F., and Kuznetsova, T. (2022). Temporal shift and predictive performance of machine learning for heart transplant outcomes. J. Heart Lung Transplant. *41*, 928-936. 10.1016/j.healun.2022.03.019.
11. Mohammadi, I., Farahani, S., Karimi, A., Jahanian, S., Firouzabadi, S.R., Alinejadfard, M., Fatemi, A., Hajikarimloo, B., and Akhlaghpasand, M. (2025). Mortality prediction of heart transplantation using machine learning models: a systematic review and meta-analysis. Front. Artif. Intell. *8*. 10.3389/frai.2025.1551959.
12. Hosmer, D.W., Lemeshow, S., and Sturdivant, R.X. (2013). Applied Logistic Regression (John Wiley & Sons).
13. Harrell, F.E., Jr. (2001). Regression modeling strategies with applications to linear models, logistic regression, and survival analysis. (Springer-Verlag).
14. Bishop, C.M. (1995). Neural Networks for Pattern Recognition (Oxford University Press).

15. Breiman, L. (2001). Random Forests. Mach. Learn. *45*, 5-32. 10.1023/A:1010933404324.
16. Chen, T., and Guestrin, C. (2016). XGBoost: A Scalable Tree Boosting System. Proceedings of the 22nd ACM SIGKDD International Conference on Knowledge Discovery and Data Mining, 2016/08/13/. (Association for Computing Machinery), pp. 785-794.
17. Lundberg, S.M., and Lee, S.-I. (2017). A unified approach to interpreting model predictions. Advances in neural information processing systems, 2017. I. Guyon, U.V. Luxburg, S. Bengio, H. Wallach, R. Fergus, S. Vishwanathan, and R. Garnett, eds. (Curran Associates, Inc.), pp. 4765-4774.
18. Ribeiro, M.T., Singh, S., and Guestrin, C. (2016). "Why Should I Trust You?": Explaining the Predictions of Any Classifier. Proceedings of the 22nd ACM SIGKDD International Conference on Knowledge Discovery and Data Mining, 2016/08/13/. (Association for Computing Machinery), pp. 1135-1144.
19. Friedman, J.H. (2001). Greedy function approximation: A gradient boosting machine. Ann. Stat. *29*, 1189-1232. 10.1214/aos/1013203451.
20. Apley, D.W., and Zhu, J. (2020). Visualizing the Effects of Predictor Variables in Black Box Supervised Learning Models. J. R. Stat. Soc. Ser. B. Stat. Methodol. *82*, 1059-1086. 10.1111/rssb.12377.
21. Roychowdhury, S., Lanfranchi, V., and Mazumdar, S. (2025). Evaluating explanation performance for clinical decision support systems for non-imaging data: A systematic literature review. Comput. Biol. Med. *197*, 110944. 10.1016/j.compbiomed.2025.110944.
22. Tan, S., Caruana, R., Hooker, G., and Lou, Y. (2018). Distill-and-Compare: Auditing Black-Box Models Using Transparent Model Distillation. Proceedings of the 2018 AAAI/ACM Conference on AI, Ethics, and Society, 2018/12/27/. (Association for Computing Machinery), pp. 303-310.
23. Köhler, D., Rügamer, D., Boyle, L.J., Maloney, K.O., and Schmid, M. (2025). Achieving interpretable machine learning by functional decomposition of black-box models into explainable predictor effects. npj Artif. Intell. *1*, 34. 10.1038/s44387-025-00033-7.
24. Oh, M.-Y., Kim, H.-S., Jung, Y.M., Lee, H.-C., Lee, S.-B., and Lee, S.M. (2025). Machine Learning–Based Explainable Automated Nonlinear Computation Scoring System for Health Score and an Application for Prediction of Perioperative Stroke: Retrospective Study. J. Med. Internet Res. *27*, e58021. 10.2196/58021.
25. Sagi, O., and Rokach, L. (2021). Approximating XGBoost with an interpretable decision tree. Inf. Sci. *572*, 522-542. 10.1016/j.ins.2021.05.055.
26. Hastie, T.J., and Tibshirani, R.J. (1990). Generalized Additive Models (CRC Press).
27. Lou, Y., Caruana, R., Gehrke, J., and Hooker, G. (2013). Accurate intelligible models with pairwise interactions. Proceedings of the 19th ACM SIGKDD international conference on Knowledge discovery and data mining, 2013/08/11/. (Association for Computing Machinery), pp. 623-631.
28. Nori, H., Jenkins, S., Koch, P., and Caruana, R. (2019). InterpretML: A Unified Framework for Machine Learning Interpretability. 10.48550/arXiv.1909.09223.
29. Agarwal, R., Melnick, L., Frosst, N., Zhang, X., Lengerich, B., Caruana, R., and Hinton, G. (2021). Neural Additive Models: Interpretable Machine Learning with Neural Nets. In Advances in Neural Information Processing Systems 34 (NeurIPS 2021), pp. 4699-4711.
30. Zhang, H.H., and Lin, Y. (2006). Component Selection and Smoothing for Nonparametric Regression in Exponential Families. Stat. Sin. *16*, 1021-1041.

31. Friedman, J.H., and Popescu, B.E. (2008). Predictive learning via rule ensembles. Ann. Appl. Stat. *2*, 916-954. 10.1214/07-AOAS148.
32. Kato, H., Hanada, H., and Takeuchi, I. (2023). Safe RuleFit: Learning Optimal Sparse Rule Model by Meta Safe Screening. IEEE Trans. Pattern Anal. Mach. Intell. *45*, 2330-2343. 10.1109/TPAMI.2022.3167993.
33. Chen, H., Yang, F., Duan, Y., Yang, L., and Li, J. (2024). A novel higher performance nomogram based on explainable machine learning for predicting mortality risk in stroke patients within 30 days based on clinical features on the first day ICU admission. BMC Med. Inform. Decis. Mak. *24*, 161. 10.1186/s12911-024-02547-7.
34. Yim, W.Y., Li, Y., Hou, J., Chen, Y., Xiong, T., Li, C., Lai, J., Peng, Y., Geng, B., Wu, Y., et al. (2026). Bridging machine learning and clinical practice: a multicentre nomogram for 90-day graft failure risk stratification in heart transplantation. Open Heart *13*. 10.1136/openhrt-2025-003790.
35. Walters, B., Ortega-Martorell, S., Olier, I., and Lisboa, P.J.G. (2023). How to Open a Black Box Classifier for Tabular Data. Algorithms *16*, 181. 10.3390/a16040181.
36. Lisboa, P.J.G., Jayabalan, M., Ortega-Martorell, S., Olier, I., Medved, D., and Nilsson, J. (2022). Enhanced survival prediction using explainable artificial intelligence in heart transplantation. Sci. Rep. *12*, 19525. 10.1038/s41598-022-23817-2.
37. Lisboa, P.J.G., Saralajew, S., Vellido, A., Fernández-Domenech, R., and Villmann, T. (2023). The coming of age of interpretable and explainable machine learning models. Neurocomputing *535*, 25-39. 10.1016/j.neucom.2023.02.040.
38. Hooker, G. (2007). Generalized functional anova diagnostics for high-dimensional functions of dependent variables. J. Comput. Graph. Stat. *16*, 709-732.
39. Lengerich, B., Tan, S., Chang, C.-H., Hooker, G., and Caruana, R. (2020). Purifying Interaction Effects with the Functional ANOVA: An Efficient Algorithm for Recovering Identifiable Additive Models. Proceedings of the Twenty Third International Conference on Artificial Intelligence and Statistics, 2020/06/03/. (PMLR), pp. 2402-2412.
40. Goldstein, A., Kapelner, A., Bleich, J., and Pitkin, E. (2015). Peeking Inside the Black Box: Visualizing Statistical Learning With Plots of Individual Conditional Expectation. J. Comput. Graph. Stat. *24*, 44-65. 10.1080/10618600.2014.907095.
41. Pigot, H., Lisboa, P., and Nilsson, J. (2026). PRiSM (Partial Responses in Structured Models): a Python framework for converting probabilistic binary classifiers into auditable nomograms, version 0.1.1 (Zenodo). https://doi.org/10.5281/zenodo.19632957.
42. Leppke, S., Leighton, T., Zaun, D., Chen, S.-C., Skeans, M., Israni, A.K., Snyder, J.J., and Kasiske, B.L. (2013). Scientific Registry of Transplant Recipients: collecting, analyzing, and reporting data on transplantation in the United States. Transplant. Rev. *27*, 50-56. 10.1016/j.trre.2013.01.002.
43. Stekhoven, D.J., and Bühlmann, P. (2012). MissForest—non-parametric missing value imputation for mixed-type data. Bioinformatics *28*, 112-118. 10.1093/bioinformatics/btr597.
44. Weiss, E.S., Allen, J.G., Arnaoutakis, G.J., George, T.J., Russell, S.D., Shah, A.S., and Conte, J.V. (2011). Creation of a Quantitative Recipient Risk Index for Mortality Prediction After Cardiac Transplantation (IMPACT). Ann. Thorac. Surg. *92*, 914-922. 10.1016/j.athoracsur.2011.04.030.
45. U.S. Food and Drug Administration (2016). Non-inferiority clinical trials to establish effectiveness: guidance for industry. Center for Drug Evaluation and Research (CDER) and Center for Biologics Evaluation and Research (CBER). November 2016. https://www.fda.gov/regulatory-information/search-fda-guidance-documents/non-inferiority-clinical-trials.

46. Van Calster, B., McLernon, D.J., van Smeden, M., Wynants, L., Steyerberg, E.W., and on behalf of Topic Group 'Evaluating diagnostic tests and prediction models' of the STRATOS initiative (2019). Calibration: the Achilles heel of predictive analytics. BMC Med. *17*, 230. 10.1186/s12916-019-1466-7.
47. Van Calster, B., Wynants, L., Verbeek, J.F.M., Verbakel, J.Y., Christodoulou, E., Vickers, A.J., Roobol, M.J., and Steyerberg, E.W. (2018). Reporting and Interpreting Decision Curve Analysis: A Guide for Investigators. Eur. Urol. *74*, 796-804. 10.1016/j.eururo.2018.08.038.
48. Abbas, Q., Jeong, W., and Lee, S.W. (2025). Explainable AI in Clinical Decision Support Systems: A Meta-Analysis of Methods, Applications, and Usability Challenges. Healthcare (Basel) *13*. 10.3390/healthcare13172154.
49. Rudin, C. (2019). Stop explaining black box machine learning models for high stakes decisions and use interpretable models instead. Nat. Mach. Intell. *1*, 206-215. 10.1038/s42256-019-0048-x.
50. Chen, J.H., and Asch, S.M. (2017). Machine Learning and Prediction in Medicine - Beyond the Peak of Inflated Expectations. N. Engl. J. Med. *376*, 2507-2509. 10.1056/NEJMp1702071.
51. Koh, P.W., Nguyen, T., Tang, Y.S., Mussmann, S., Pierson, E., Kim, B., and Liang, P. (2020). Concept Bottleneck Models. Proceedings of the 37th International Conference on Machine Learning, 2020/11/21/. (PMLR), pp. 5338-5348.
52. Lekadir, K., Frangi, A.F., Porras, A.R., Glocker, B., Cintas, C., Langlotz, C.P., Weicken, E., Asselbergs, F.W., Prior, F., Collins, G.S., et al. (2025). FUTURE-AI: international consensus guideline for trustworthy and deployable artificial intelligence in healthcare. BMJ *388*, e081554. 10.1136/bmj-2024-081554.
53. Copeland, H., Knezevic, I., Baran, D.A., Rao, V., Pham, M., Gustafsson, F., Pinney, S., Lima, B., Masetti, M., Ciarka, A., et al. (2023). Donor heart selection: Evidence-based guidelines for providers. J. Heart Lung Transplant. *42*, 7-29. 10.1016/j.healun.2022.08.030.
54. Ware, J.H., Harrington, D., Hunter, D.J., and D'Agostino, R.B. (2012). Missing Data. N. Engl. J. Med. *367*, 1353-1354. 10.1056/NEJMsm1210043.

## Figure legends

### Figure 1. Overview of the PRiSM framework

PRiSM (Partial Responses in Structured Models) converts a trained binary prediction model for tabular clinical data into a transparent standalone nomogram. In Phase I, the source model is broken down on the log-odds scale into individual predictor effects and selected pairwise interactions. Least absolute shrinkage and selection operator (LASSO) regularization is then used to retain the effects included in the translated nomogram. In Phase II, the effects selected in Phase I define a Partial Response Network (PRN), a structured neural network in which each subnetwork receives only one or two input variables corresponding to a single selected effect. The PRN is trained on the training-cohort data and outcomes, then broken down again and reselected to generate a PRN-refined nomogram. Source models and PRN models were trained or implemented in the training cohort ($n$ = 32,543); hyperparameter, lambda, and model-selection decisions were made in the tuning cohort ($n$ = 8,558); and final evaluation was performed in the temporal external-validation cohort ($n$ = 9,255). ANOVA denotes analysis of variance. See also Tables S9-S11.

### Figure 2. Predictive noninferiority and decision-curve performance of translated nomograms

All results are from the temporal external-validation cohort (2020–2022; $n$ = 9,255). (A) Retention-adjusted contrast T = AUROC_nomogram - 0.90 × AUROC_source - 0.05 for each translated nomogram. Positive values indicate that the translated nomogram retained more than 90% of its source model's discrimination above chance, and negative values indicate that it retained less. Horizontal lines are two-sided 95% confidence intervals for T, derived from the paired DeLong covariance matrix of the two AUROCs. The vertical line at zero is the noninferiority boundary: a confidence interval lying entirely to its right meets the criterion. (B-D) Decision-curve analysis for the source models (B), the baseline PRiSM nomograms (C), and the sparse PRN nomograms (D), across threshold probabilities of 5% to 20%; treat-all and treat-none strategies are shown for comparison. Net benefit is plotted as a point estimate. AUROC denotes area under the receiver-operating-characteristic curve; PRiSM, Partial Responses in Structured Models; PRN, Partial Response Network; IMPACT, Index for Mortality Prediction After Cardiac Transplantation; LR, logistic regression; MLP, multilayer perceptron; RF, random forest; and XGB, extreme gradient boosting. See also Tables S12-S14 and S16.

**Figure 3. LASSO regularization paths for Phase I and Phase II**

For the multilayer-perceptron and logistic-regression source models, each panel plots, against the LASSO penalty lambda, tuning-cohort and training-cohort AUROC on the left axis and the number of selected terms on the right axis, with the total resolved into its univariate and bivariate components. Lambda decreases from left to right, so terms accumulate rightward. Solid curves are tuning-cohort AUROC ($n$ = 8,558) and dashed curves training-cohort AUROC ($n$ = 32,543). (A and B) Phase I decomposition of the source model, for the multilayer perceptron (A) and logistic regression (B), with vertical lines marking both selection points along a single path: the sparse lambda and, at weaker penalization, the baseline lambda. (C and D) Phase II PRN refinement following the sparse Phase I selection, for the multilayer perceptron (C) and logistic regression (D). The PRN is trained on the terms selected at the sparse lambda, and LASSO selection is repeated on the re-decomposed PRN responses, so only the baseline lambda applied at this stage is marked. AUROC denotes area under the receiver-operating-characteristic curve; LASSO, least absolute shrinkage and selection operator; LR, logistic regression; MLP, multilayer perceptron; and PRN, Partial Response Network. See also Tables S10 and S11.

**Figure 4. Representative partial-response functions and a sparse PRN nomogram**

Responses were derived from the models fitted in the training cohort ($n$ = 32,543). (A) Representative partial-response functions for 3 continuous predictors in the PRiSM and sparse PRN models: total bilirubin, creatinine clearance, and ischemic time. Curves are shown for the source-model families to illustrate how Phase II PRN refinement alters the translated response shapes while preserving the permitted scaffold of variables and interactions selected in Phase I. For the IMPACT source model, the Phase I ("PRiSM") bilirubin and creatinine-clearance responses step at the thresholds IMPACT defines (bilirubin 17.1, 34.2, and 68.4 umol/L [1.0, 2.0, and 4.0 mg/dL]; creatinine clearance 30 and 50 mL/min), recovering its published additive structure. (B) The sparse multilayer-perceptron (MLP) PRN nomogram derived after Phase II refinement. The top scale indicates points assigned to each predictor value, and the bottom scale converts total points to predicted mortality risk. PRiSM denotes Partial Responses in Structured Models; PRN, Partial Response Network; IMPACT, Index for Mortality Prediction After Cardiac Transplantation; LR, logistic regression; MLP, multilayer perceptron; RF, random forest; and XGB, extreme gradient boosting. See also Figures S1-S4.

## Figures

Figures 1-4 are supplied as separate files (Figure1.pdf through Figure4.pdf). Titles and legends appear under Figure legends above.

## Tables

**Table 1. Baseline characteristics of the study cohorts and predictors used in the source models.**

| Characteristic | Training Cohort (*n*=32,543) | Tuning Cohort (*n*=8,558) | Temporal External-Validation Cohort (*n*=9,255) |
|---|---|---|---|
| **Recipient Characteristics** | | | |
| Recipient Age, yr — median [IQR] | 55.0 [46.0, 62.0] | 57.0 [46.2, 63.0] | 57.0 [46.0, 63.0] |
| Female Sex — no. (%)[b] | 7,913 (24.3) | 2,310 (27.0) | 2,435 (26.3) |
| Race or Ethnic Group — no. (%)[b] | | | |
| White | 22,760 (69.9) | 5,412 (63.2) | 5,414 (58.5) |
| Black | 6,073 (18.7) | 1,980 (23.1) | 2,409 (26.0) |
| Hispanic/Latino | 2,507 (7.7) | 772 (9.0) | 968 (10.5) |
| Other | 1,203 (3.7) | 394 (4.6) | 464 (5.0) |
| Body Mass Index, kg/m$^2$ — median [IQR][a] | 26.6 [23.4, 30.0] | 27.4 [24.0, 31.1] | 27.3 [24.1, 31.1] |
| *Missing* | *579 (1.8)* | *160 (1.9)* | *218 (2.4)* |
| Height, cm — median [IQR][a] | 175.3 [167.6, 180.3] | 175.0 [167.6, 180.3] | 175.0 [167.6, 180.3] |
| *Missing* | *335 (1.0)* | *106 (1.2)* | *126 (1.4)* |
| Primary Cause of Heart Failure — no. (%) | | | |
| Nonischemic cardiomyopathy | 16,282 (50.0) | 5,201 (60.8) | 5,786 (62.5) |
| Ischemic cardiomyopathy | 13,125 (40.3) | 2,501 (29.2) | 2,510 (27.1) |
| Congenital | 836 (2.6) | 290 (3.4) | 356 (3.8) |
| Graft Failure | 978 (3.0) | 234 (2.7) | 254 (2.7) |
| Valvular Heart Disease | 608 (1.9) | 98 (1.1) | 88 (1.0) |
| Other | 714 (2.2) | 234 (2.7) | 261 (2.8) |
| Status at Transplant — no. (%)[a] | | | |
| ICU | 9,603 (29.5) | 3,370 (39.4) | 5,099 (55.1) |
| Hospitalized (not ICU) | 5,723 (17.6) | 1,299 (15.2) | 1,306 (14.1) |
| Not Hospitalized | 17,217 (52.9) | 3,889 (45.4) | 2,850 (30.8) |
| Duration on Waiting List, days — median [IQR][a] | 91.0 [27.0, 256.0] | 73.0 [18.0, 257.0] | 31.0 [9.0, 150.0] |

| Characteristic | Training Cohort (*n*=32,543) | Tuning Cohort (*n*=8,558) | Temporal External-Validation Cohort (*n*=9,255) |
|---|---|---|---|
| **Laboratory Values — median [IQR]** | | | |
| Creatinine Clearance, mL/min | 76.6 [57.8, 99.8] | 78.6 [59.0, 102.3] | 78.5 [59.0, 103.6] |
| *Missing* | *937 (2.9)* | *261 (3.0)* | *361 (3.9)* |
| Total Bilirubin, mg/dL | 0.8 [0.5, 1.2] | 0.7 [0.4, 1.0] | 0.7 [0.5, 1.1] |
| *Missing* | *1,204 (3.7)* | *82 (1.0)* | *96 (1.0)* |
| Serum Albumin, g/dL[a] | 3.7 [3.3, 4.1] | 3.8 [3.4, 4.2] | 3.8 [3.3, 4.1] |
| *Missing* | *8,435 (25.9)* | *4,880 (57.0)* | *202 (2.2)* |
| **Comorbidities and Mechanical Support — no. (%)** | | | |
| Diabetes Mellitus[a] | | | |
| No | 23,984 (74.2) | 6,083 (71.1) | 6,442 (69.6) |
| Type I | 616 (1.9) | 89 (1.0) | 106 (1.1) |
| Type II | 5,557 (17.2) | 2,285 (26.7) | 2,599 (28.1) |
| Other | 2,159 (6.7) | 97 (1.1) | 107 (1.2) |
| Dialysis Dependence | 1,165 (3.6) | 419 (4.9) | 528 (5.7) |
| Infection (within 2 weeks prior to transplant) | 3,551 (11.3) | 818 (9.6) | 1,042 (11.3) |
| *Missing* | *1,046 (3.2)* | *20 (0.2)* | *39 (0.4)* |
| History of Prior Blood Transfusions[a] | 7,189 (23.3) | 1,657 (19.4) | 1,452 (15.8) |
| *Missing* | *1,679 (5.2)* | *30 (0.4)* | *50 (0.5)* |
| Implantable Cardioverter Defibrillator (ICD)[a] | 21,263 (66.4) | 6,318 (74.6) | 6,246 (68.3) |
| *Missing* | *509 (1.6)* | *89 (1.0)* | *116 (1.3)* |
| Ventilator Support | 681 (2.1) | 119 (1.4) | 201 (2.2) |
| ECMO | 184 (0.6) | 239 (2.8) | 560 (6.1) |
| IABP[b] | 1,866 (5.7) | 1,419 (16.6) | 2,580 (27.9) |
| VAD | | | |
| None | 20,066 (64.7) | 4,807 (56.2) | 6,051 (65.4) |
| LVAD | 8,274 (26.7) | 3,512 (41.0) | 2,986 (32.3) |
| RVAD | 64 (0.2) | 26 (0.3) | 40 (0.4) |
| BiVAD | 1,060 (3.4) | 213 (2.5) | 178 (1.9) |
| Unspecified | 1,569 (5.1) | 0 (0.0) | 0 (0.0) |
| *Missing* | *1,510 (4.6)* | — | — |
| HCV Seropositivity[a] | | | |
| Negative | 29,364 (92.1) | 8,206 (96.1) | 8,913 (96.4) |
| Not Done | 1,872 (5.9) | 131 (1.5) | 119 (1.3) |
| Positive | 642 (2.0) | 202 (2.4) | 218 (2.4) |
| *Missing* | *665 (2.0)* | *19 (0.2)* | *5 (0.1)* |
| **Donor Characteristics** | | | |
| Donor Age, yr — median [IQR][a] | 30.0 [22.0, 41.0] | 31.0 [24.0, 40.0] | 32.0 [25.0, 39.0] |
| Donor Weight, kg — median [IQR][a] | 79.4 [69.0, 91.0] | 80.5 [70.0, 93.5] | 81.7 [70.7, 95.0] |
| *Missing* | *415 (1.3)* | *163 (1.9)* | *219 (2.4)* |
| Total Ischemic Time, hr — median [IQR][a] | 3.2 [2.5, 3.8] | 3.2 [2.5, 3.8] | 3.5 [2.9, 4.0] |
| *Missing* | *1,428 (4.4)* | *139 (1.6)* | *80 (0.9)* |
| History of Heavy Alcohol Use — no. (%)[a] | 5,457 (17.0) | 1,515 (18.2) | 1,781 (20.0) |
| *Missing* | *511 (1.6)* | *211 (2.5)* | *363 (3.9)* |

Values are presented as medians with interquartile ranges for continuous variables and as counts and percentages for categorical variables. Variables carrying no footnote letter were used in all 5 source models. Missingness is reported for variables with at least 1% missing data in any cohort; percentages for these variables are calculated on the nonmissing observations within each cohort, whereas the missingness percentages themselves use the full cohort as the denominator. In the race or ethnic group category, "Other" included Asian, Native American, Pacific Islander, and multiracial categories. BiVAD included left ventricular assist device plus right ventricular assist device and total artificial heart. ECMO denotes extracorporeal membrane oxygenation; HCV, hepatitis C virus; IABP, intraaortic balloon pump; ICD, implantable cardioverter–defibrillator; ICU, intensive care unit; IMPACT, Index for Mortality Prediction After Cardiac Transplantation; IQR, interquartile range; LVAD, left ventricular assist device; RVAD, right ventricular assist device; and VAD, ventricular assist device.

[a]Used in the logistic-regression, multilayer perceptron, random-forest, and XGBoost models.

[b]Used in the IMPACT model only.

See also Table S15.

**Table 2. Performance of source models and translated nomograms in the temporal external-validation cohort.**

| Model | Version | No. of Variables or Terms | AUROC (95% CI) | Retention Contrast T (95% CI) | Calibration Slope (95% CI) | CITL (95% CI) | Observed:Expected Ratio (95% CI) |
|---|---|---|---|---|---|---|---|
| IMPACT | Source | 12 | 0.619 (0.599, 0.639) | — | 0.602 (0.502, 0.701) | -0.535 (-0.608, -0.463) | 0.642 (0.603, 0.683) |
| | PRiSM | 9 | 0.630 (0.610, 0.649) | +0.023[a] (+0.011, +0.034) | 0.988 (0.837, 1.139) | -0.170 (-0.241, -0.099) | 0.861 (0.809, 0.915) |
| | Sparse PRiSM | 5 | 0.601 (0.581, 0.621) | -0.006 (-0.021, +0.010) | 1.360 (1.119, 1.600) | -0.137 (-0.207, -0.067) | 0.885 (0.831, 0.941) |
| | Sparse PRN | 5 | 0.620 (0.600, 0.639) | +0.013 (-0.004, +0.029) | 0.964 (0.812, 1.115) | -0.117 (-0.188, -0.046) | 0.902 (0.846, 0.958) |
| LR | Source | 22 | 0.620 (0.600, 0.640) | — | 0.806 (0.685, 0.927) | -0.226 (-0.297, -0.154) | 0.823 (0.774, 0.875) |
| | PRiSM | 21 | 0.625 (0.605, 0.645) | +0.017[a] (+0.012, +0.022) | 0.703 (0.600, 0.805) | -0.198 (-0.270, -0.126) | 0.845 (0.795, 0.899) |
| | Sparse PRiSM | 9 | 0.611 (0.591, 0.631) | +0.004 (-0.007, +0.015) | 1.179 (0.988, 1.371) | -0.146 (-0.216, -0.075) | 0.878 (0.826, 0.935) |
| | Sparse PRN | 9 | 0.625 (0.605, 0.645) | +0.017[a] (+0.005, +0.029) | 0.905 (0.770, 1.040) | -0.151 (-0.223, -0.080) | 0.876 (0.824, 0.932) |
| MLP | Source | 22 | 0.645 (0.625, 0.665) | — | 0.739 (0.643, 0.834) | -0.253 (-0.326, -0.181) | 0.808 (0.760, 0.859) |
| | PRiSM | 30 | 0.646 (0.626, 0.666) | +0.016[a] (+0.010, +0.021) | 0.784 (0.682, 0.885) | -0.199 (-0.271, -0.127) | 0.845 (0.795, 0.898) |
| | Sparse PRiSM | 10 | 0.636 (0.616, 0.655) | +0.005 (-0.007, +0.017) | 1.246 (1.073, 1.419) | -0.120 (-0.190, -0.049) | 0.899 (0.845, 0.956) |
| | Sparse PRN | 10 | 0.641 (0.621, 0.661) | +0.011 (0.000, +0.021) | 0.972 (0.842, 1.101) | -0.131 (-0.203, -0.060) | 0.891 (0.839, 0.948) |
| RF | Source | 22 | 0.642 (0.622, 0.661) | — | 0.730 (0.624, 0.836) | -0.248 (-0.320, -0.176) | 0.810 (0.760, 0.862) |
| | PRiSM | 22 | 0.641 (0.621, 0.660) | +0.013[a] (+0.006, +0.020) | 0.891 (0.768, 1.014) | -0.099 (-0.171, -0.028) | 0.917 (0.861, 0.975) |
| | Sparse PRiSM | 8 | 0.621 (0.601, 0.641) | -0.006 (-0.016, +0.004) | 0.977 (0.824, 1.129) | -0.142 (-0.212, -0.071) | 0.883 (0.829, 0.938) |
| | Sparse PRN | 8 | 0.624 (0.604, 0.644) | -0.004 (-0.015, +0.008) | 0.907 (0.772, 1.041) | -0.174 (-0.245, -0.103) | 0.859 (0.808, 0.913) |
| XGB | Source | 22 | 0.646 (0.627, 0.666) | — | 0.811 (0.705, 0.917) | -0.156 (-0.228, -0.084) | 0.875 (0.824, 0.929) |
| | PRiSM | 26 | 0.644 (0.624, 0.663) | +0.012[a] (+0.008, +0.017) | 0.781 (0.678, 0.884) | -0.158 (-0.231, -0.086) | 0.873 (0.822, 0.929) |
| | Sparse PRiSM | 11 | 0.635 (0.615, 0.654) | +0.003 (-0.005, +0.012) | 1.053 (0.906, 1.200) | -0.112 (-0.183, -0.041) | 0.906 (0.851, 0.962) |
| | Sparse PRN | 11 | 0.646 (0.627, 0.666) | +0.014[a] (+0.005, +0.024) | 1.019 (0.887, 1.151) | -0.117 (-0.188, -0.046) | 0.902 (0.849, 0.959) |

All results are from the temporal external-validation cohort (2020–2022; $n$ = 9,255). No. of variables or terms denotes input variables for source models and nomogram terms for all other rows. The retention contrast is T = AUROC_nomogram - 0.90 × AUROC_source - 0.05, which exceeds zero exactly when the nomogram retains more than 90% of its source model's discrimination above chance. The difference in AUROC between each nomogram and its source model, a descriptive quantity, is reported in Table S14. Confidence intervals for AUROC were estimated with the DeLong method and those for T from the same paired DeLong covariance matrix; confidence intervals for the calibration slope and CITL with the Wald method; and confidence intervals for the observed-to-expected ratio with a seeded percentile bootstrap over 1,000 resamples. A calibration slope of 1, CITL of 0, and an observed-to-expected ratio of 1 indicate perfect calibration. PRiSM denotes the Phase I nomogram at the baseline lambda; Sparse PRiSM, the Phase I nomogram at the sparse lambda, from which each sparse PRN nomogram was refined; and Sparse PRN, its Phase II refinement. The fourth combination, Phase II at the baseline lambda, is reported in Tables S12-S14. AUROC denotes area under the receiver-operating-characteristic curve; CITL, calibration-in-the-large; IMPACT, Index for Mortality Prediction After Cardiac Transplantation; LR, logistic regression; MLP, multilayer perceptron; PRiSM, Partial Responses in Structured Models; PRN, Partial Response Network; RF, random forest; and XGB, extreme gradient boosting.

[a]Noninferiority met: the lower bound of the two-sided 95% confidence interval for T exceeded zero. For the MLP sparse PRN nomogram the lower bound does not exceed zero (-0.0003) but rounds to 0.000 at the three decimal places shown.

See also Tables S12-S14 and S16.

**Table 3. Translated nomogram discrimination versus de novo additive comparators in temporal external validation.**

| Source model | Version | No. of Terms | Nomogram AUROC (95% CI) | Difference in AUROC vs EBM (95% CI) | Difference in AUROC vs GAM (95% CI) | Difference in AUROC vs NAM (95% CI) |
|---|---|---|---|---|---|---|
| MLP | PRiSM | 30 | 0.646 (0.626, 0.666) | -0.001 (-0.006, +0.005) | +0.001 (-0.003, +0.005) | +0.032 (+0.019, +0.044) |
| | Sparse PRiSM | 10 | 0.636 (0.616, 0.655) | -0.011 (-0.023, +0.001) | -0.009 (-0.020, +0.002) | +0.021 (+0.004, +0.038) |
| | Sparse PRN | 10 | 0.641 (0.621, 0.661) | -0.006 (-0.016, +0.004) | -0.004 (-0.013, +0.005) | +0.026 (+0.011, +0.042) |
| XGB | PRiSM | 26 | 0.644 (0.624, 0.663) | -0.003 (-0.009, +0.003) | -0.001 (-0.007, +0.005) | +0.029 (+0.017, +0.042) |
| | Sparse PRiSM | 11 | 0.635 (0.615, 0.654) | -0.012 (-0.022, -0.002) | -0.010 (-0.020, 0.000) | +0.020 (+0.005, +0.035) |
| | Sparse PRN | 11 | 0.646 (0.627, 0.666) | -0.001 (-0.011, +0.009) | +0.001 (-0.007, +0.010) | +0.032 (+0.016, +0.047) |
| RF | PRiSM | 22 | 0.641 (0.621, 0.660) | -0.006 (-0.016, +0.003) | -0.004 (-0.014, +0.006) | +0.026 (+0.012, +0.040) |
| | Sparse PRiSM | 8 | 0.621 (0.601, 0.641) | -0.026 (-0.039, -0.013) | -0.024 (-0.037, -0.010) | +0.007 (-0.010, +0.023) |
| | Sparse PRN | 8 | 0.624 (0.604, 0.644) | -0.023 (-0.036, -0.011) | -0.021 (-0.033, -0.009) | +0.009 (-0.006, +0.025) |
| IMPACT | PRiSM | 9 | 0.630 (0.610, 0.649) | -0.017 (-0.032, -0.003) | -0.015 (-0.030, 0.000) | +0.015 (-0.003, +0.033) |
| | Sparse PRiSM | 5 | 0.601 (0.581, 0.621) | -0.046 (-0.063, -0.029) | -0.044 (-0.061, -0.026) | -0.014 (-0.032, +0.005) |
| | Sparse PRN | 5 | 0.620 (0.600, 0.639) | -0.028 (-0.043, -0.011) | -0.025 (-0.041, -0.009) | +0.005 (-0.013, +0.023) |
| LR | PRiSM | 21 | 0.625 (0.605, 0.645) | -0.022 (-0.031, -0.013) | -0.020 (-0.028, -0.012) | +0.010 (-0.003, +0.023) |
| | Sparse PRiSM | 9 | 0.611 (0.591, 0.631) | -0.036 (-0.050, -0.021) | -0.034 (-0.047, -0.020) | -0.003 (-0.020, +0.014) |
| | Sparse PRN | 9 | 0.625 (0.605, 0.645) | -0.022 (-0.035, -0.010) | -0.020 (-0.032, -0.008) | +0.010 (-0.006, +0.027) |

All comparisons are in the temporal external-validation cohort (2020–2022; $n$ = 9,255). Reference external-validation AUROCs: EBM 0.647 (95% CI, 0.627 to 0.667; 42 terms, of which 20 pairwise interactions), GAM 0.645 (95% CI, 0.625 to 0.664; 22 terms), NAM 0.615 (95% CI, 0.595 to 0.635; 22 terms). IMPACT and LR are transparent controls: their nomograms reproduce a source model that is itself below these comparators, so a difference in those rows reflects the source model rather than the translation. Positive differences favor PRiSM. The nomogram versions are those of Table 2. No equivalence margin was set for these comparisons, so they can detect a difference but cannot establish equivalence. Confidence intervals for AUROC were estimated with the DeLong method and those for the difference in AUROC from the paired DeLong covariance matrix. AUROC denotes area under the receiver-operating-characteristic curve; EBM, explainable boosting machine; GAM, spline-based generalized additive model; IMPACT, Index for Mortality Prediction After Cardiac Transplantation; LR, logistic regression; MLP, multilayer perceptron; NAM, neural additive model; PRiSM, Partial Responses in Structured Models; PRN, Partial Response Network; RF, random forest; and XGB, extreme gradient boosting.

See also Tables S17-S19.

Figure 1

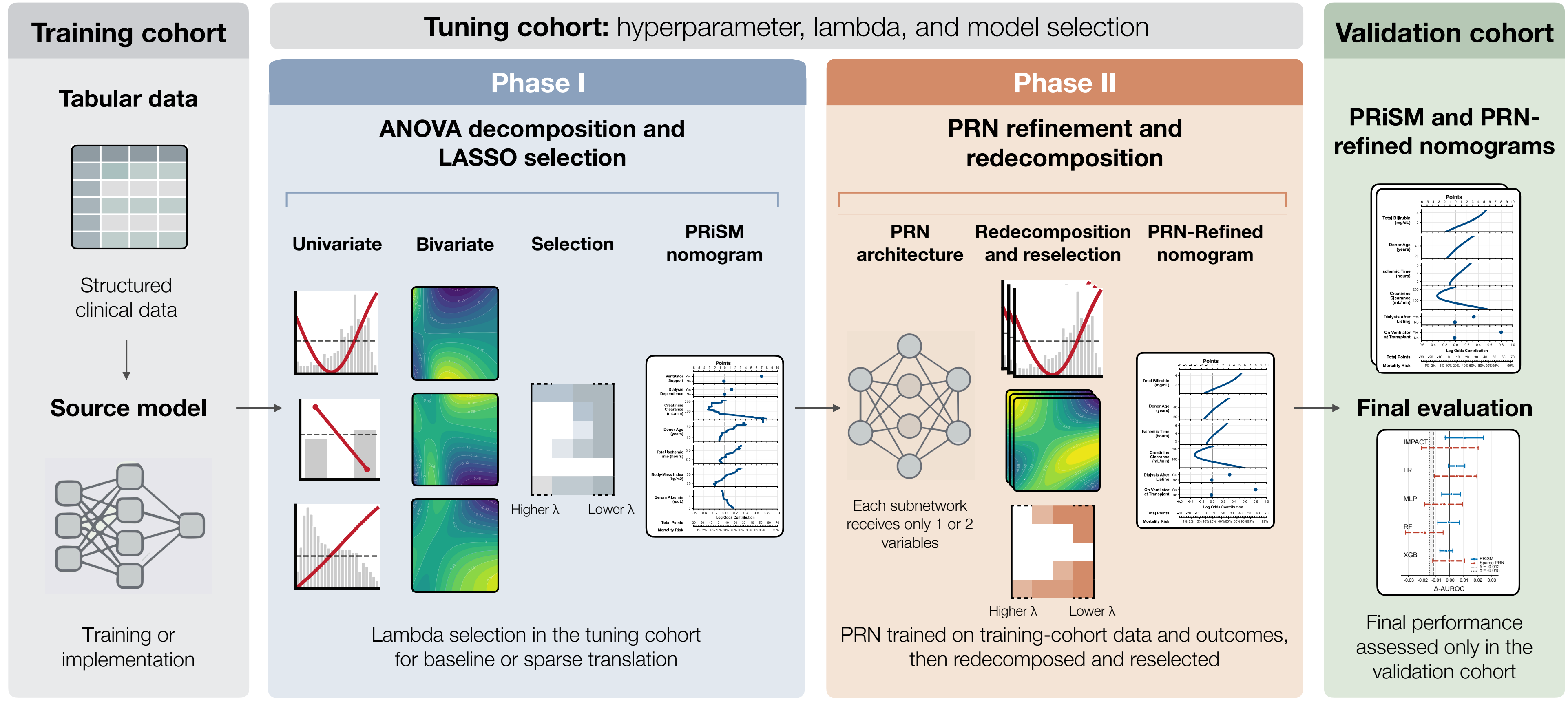

Training cohort
Tuning cohort: hyperparameter, lambda, and model selection
Validation cohort
Tabular data
Structured clinical data
Source model
Training or implementation
Phase I
ANOVA decomposition and LASSO selection
Univariate
Bivariate
Selection
PRiSM nomogram
Higher λ
Lower λ
Lambda selection in the tuning cohort for baseline or sparse translation
Phase II
PRN refinement and redecomposition
PRN architecture
Redecomposition and reselection
PRN-Refined nomogram
Each subnetwork receives only 1 or 2 variables
Higher λ
Lower λ
PRN trained on training-cohort data and outcomes, then redecomposed and reselected
PRiSM and PRN-refined nomograms
Final evaluation
Final performance assessed only in the validation cohort

Figure 2

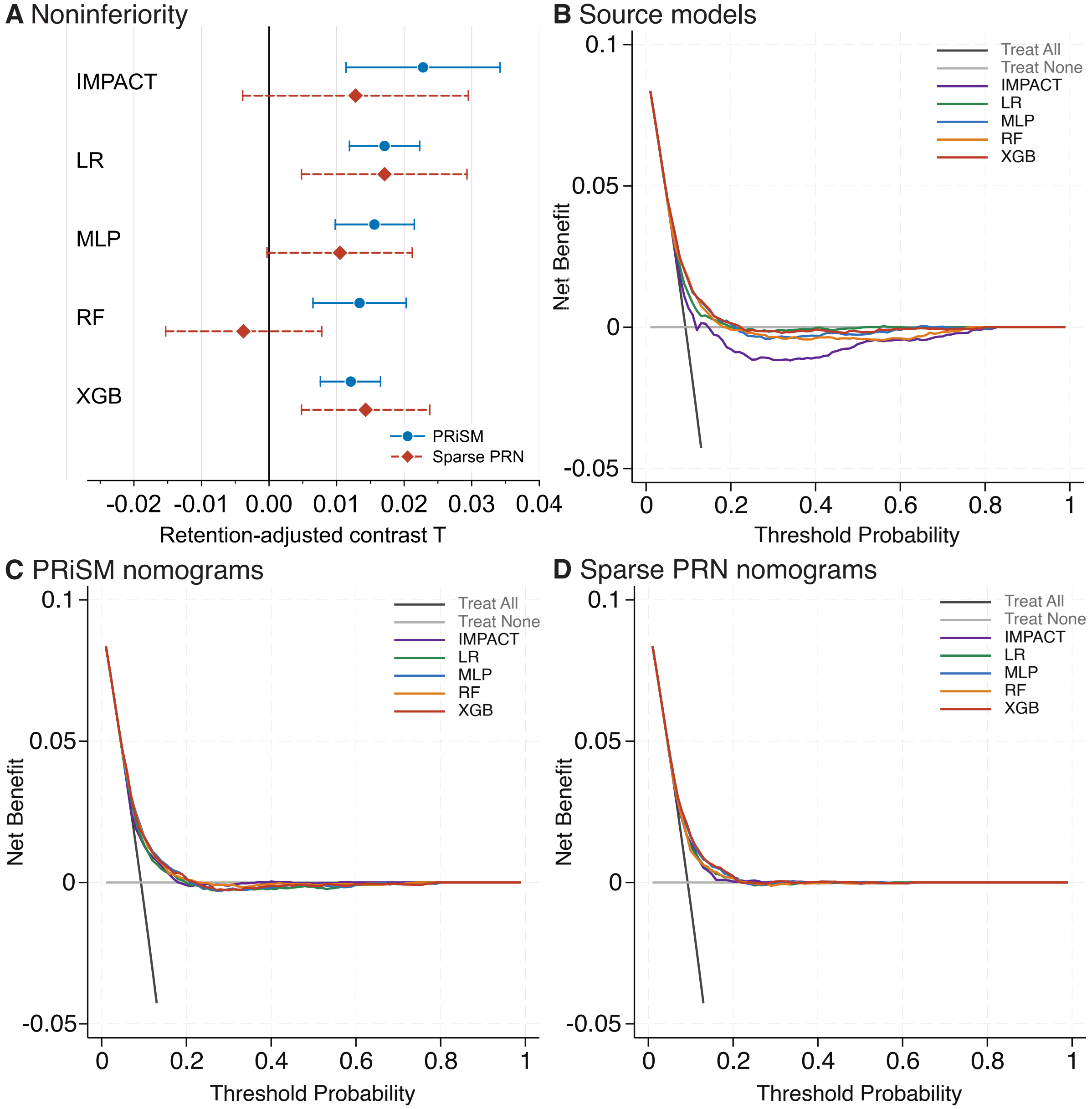

A Noninferiority
IMPACT
LR
MLP
RF
XGB
PRiSM
Sparse PRN
-0.02
-0.01
0.00
0.01
0.02
0.03
0.04
Retention-adjusted contrast T
B Source models
C PRiSM nomograms
D Sparse PRN nomograms
Net Benefit
Threshold Probability
0.1
0.05
0
-0.05
0
0.2
0.4
0.6
0.8
1
Treat All
Treat None
IMPACT
LR
MLP
RF
XGB

Figure 3

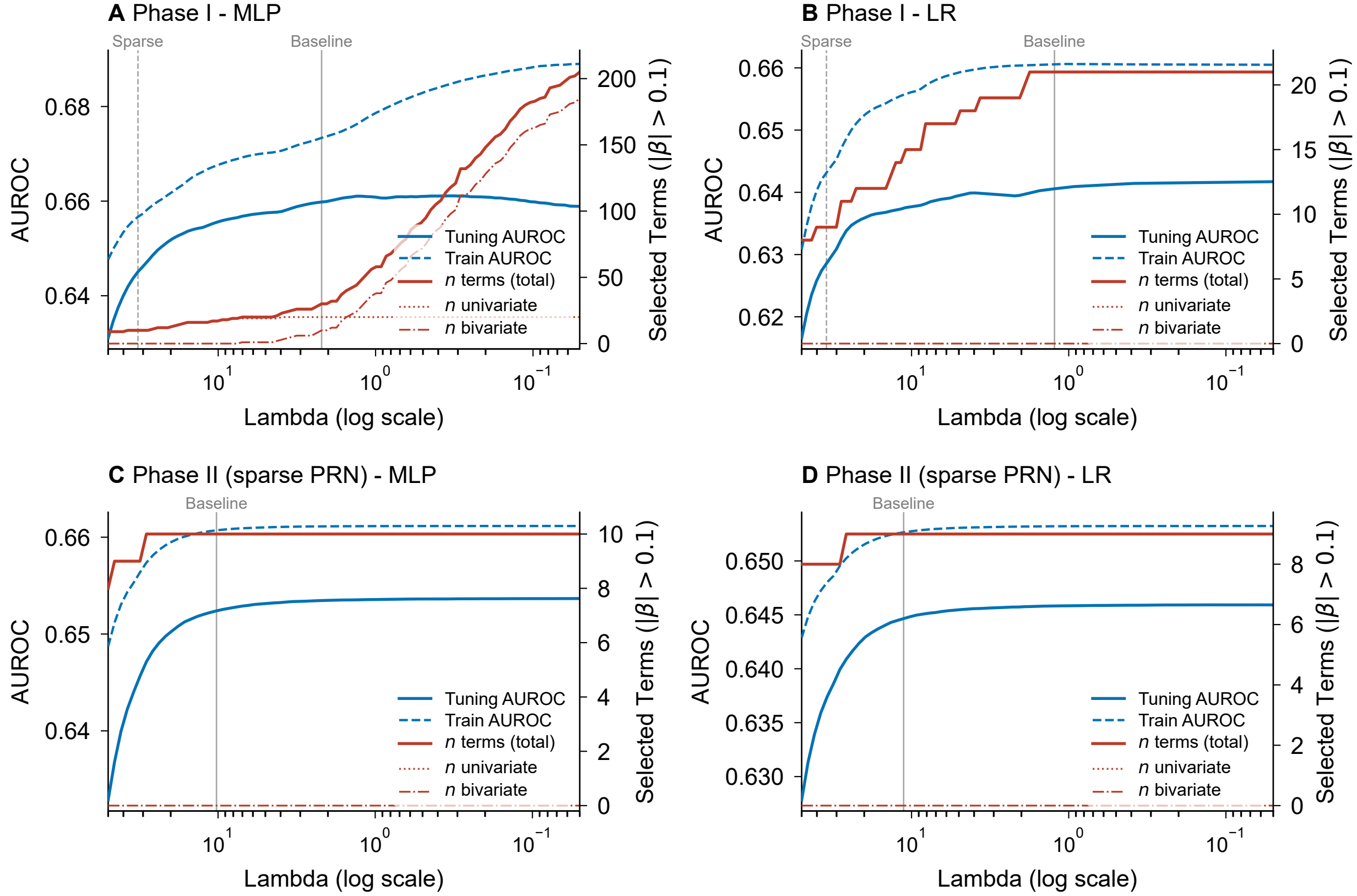

A Phase I - MLP
Sparse
Baseline
AUROC
Selected Terms (|β| > 0.1)
Lambda (log scale)
Tuning AUROC
Train AUROC
n terms (total)
n univariate
n bivariate
B Phase I - LR
Sparse
Baseline
AUROC
Selected Terms (|β| > 0.1)
Lambda (log scale)
Tuning AUROC
Train AUROC
n terms (total)
n univariate
n bivariate
C Phase II (sparse PRN) - MLP
Baseline
AUROC
Selected Terms (|β| > 0.1)
Lambda (log scale)
Tuning AUROC
Train AUROC
n terms (total)
n univariate
n bivariate
D Phase II (sparse PRN) - LR
Baseline
AUROC
Selected Terms (|β| > 0.1)
Lambda (log scale)
Tuning AUROC
Train AUROC
n terms (total)
n univariate
n bivariate

Figure 4

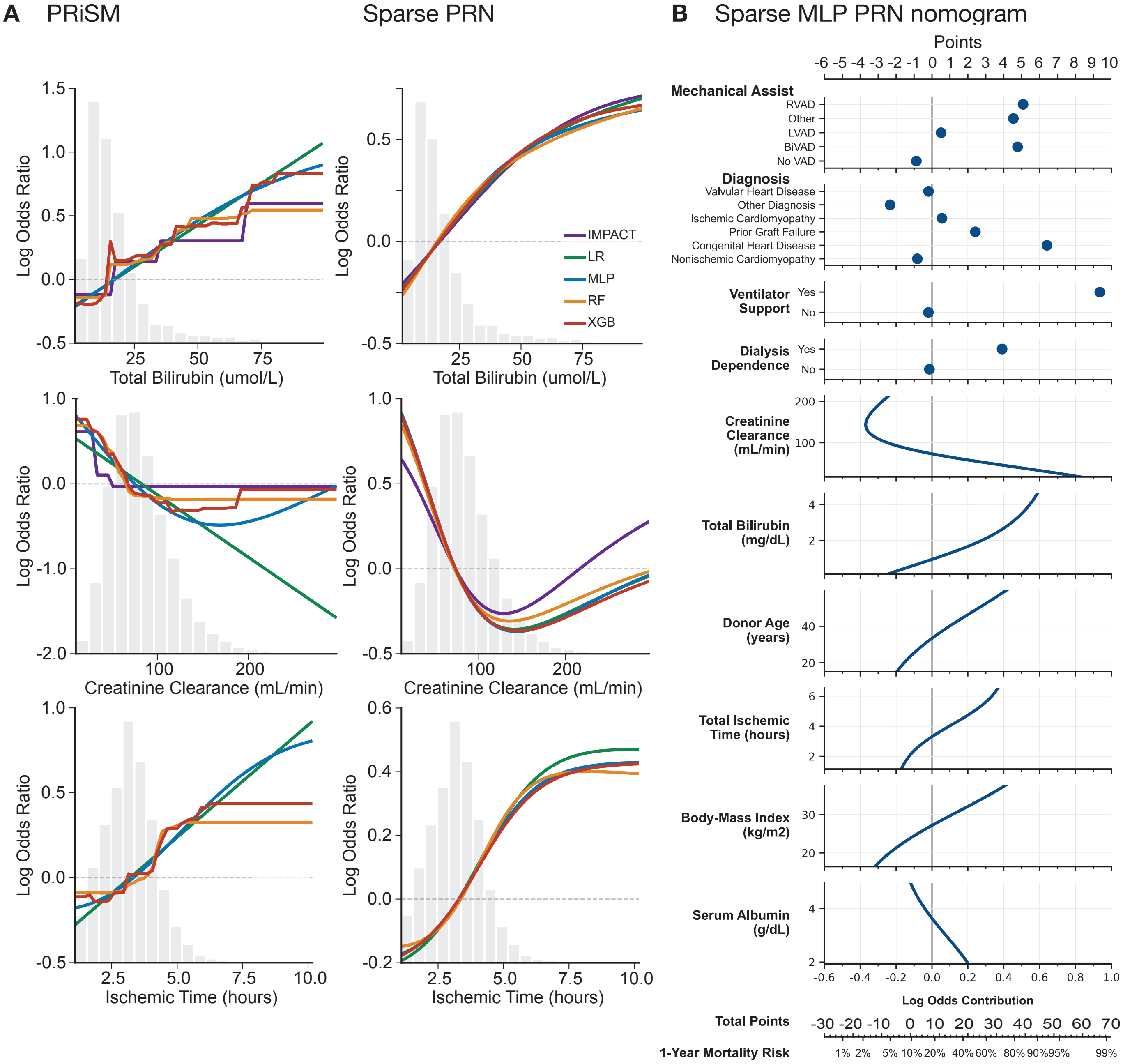

A PRiSM
Sparse PRN
B Sparse MLP PRN nomogram
Log Odds Ratio
Total Bilirubin (umol/L)
Creatinine Clearance (mL/min)
Ischemic Time (hours)
IMPACT
LR
MLP
RF
XGB
Points
Mechanical Assist
RVAD
Other
LVAD
BiVAD
No VAD
Diagnosis
Valvular Heart Disease
Other Diagnosis
Ischemic Cardiomyopathy
Prior Graft Failure
Congenital Heart Disease
Nonischemic Cardiomyopathy
Ventilator Support
Yes
No
Dialysis Dependence
Yes
No
Creatinine Clearance (mL/min)
Total Bilirubin (mg/dL)
Donor Age (years)
Total Ischemic Time (hours)
Body-Mass Index (kg/m2)
Serum Albumin (g/dL)
Log Odds Contribution
Total Points
1-Year Mortality Risk

Supplemental Information

# Translation of Black-Box Clinical Prediction Models into Standalone Transparent Nomograms: Temporal External Validation in Heart Transplantation

Henry Pigot, Paulo J.G. Lisboa, Sandra Ortega-Martorell, Ivan Olier, Joseph Mahon, Johan Nilsson

Supplementary methods, figures and tables referenced from the main text.

# S1 Supplementary Methods

## S1.1 Study Design Details

**Reporting Standards.** The study design, model development, and validation were conducted in accordance with the TRIPOD+AI statement [S1] (Table S1). Adherence to individual TRIPOD+AI items is documented in the checklist.

**Data Source and Ethics.** The data source is the Scientific Registry of Transplant Recipients (SRTR); eligibility criteria, extraction date, and ethics approval are given in main text Methods. Records with missing transplant dates or outcome status were excluded, as were patients censored before 365 days, whose 1-year mortality status is indeterminate for binary modeling. The final adult cohort of 50,356 recipients is 60.5% of adult heart transplant recipients in the source dataset; Table S2 gives the sequential exclusion criteria. Outcome ascertainment is registry-based and objective (vital status at 365 days); no additional outcome blinding was performed. The study was not prospectively registered because it is a secondary analysis of de-identified registry data with no patient contact.

**Cohort Assignment.** Records were ordered by transplant year and allocated to 3 nonoverlapping cohorts—training, internal validation (tuning), and external validation—using clean calendar-year boundaries. The target allocation was approximately 3:1:1, and the two cutoff years nearest to this target were selected: training included transplant years 2000–2016, tuning 2017–2019, and external validation 2020–2022. Because cohort boundaries fall on whole calendar years, the actual sample sizes deviate slightly from the target ratio (Table 1 in main text). No within-year randomization was required, and the split is fully reproducible from the transplant-year variable alone. No patient appears in more than one cohort. The same cohorts were used throughout the study.

**Leakage Audit.** We verified (i) zero intersection of patient identifiers across cohorts, (ii) reproducibility of cohort assignment from the transplant-year cutoffs alone, and (iii) that no post-transplant variables or future information were used as predictors. Scaling factors were derived exclusively from the training cohort and applied unchanged to the tuning and external-validation cohorts. Model hyperparameters, early stopping criteria, and the PRiSM regularization parameter ($\lambda$) were selected using only the tuning cohort performance, with the external-validation cohort remaining entirely held out until final evaluation.

## S1.2 Predictor Selection and Preprocessing

**Predictor Selection.** Predictors were selected through a two-stage process. First, expert clinical review identified 57 candidate variables from 1,293 available UNOS registry variables based on prior literature and clinical experience, spanning recipient demographics, clinical status, comorbidities, donor characteristics, laboratory values, immunology, and transplant factors. Five derived variables were included (e.g., creatinine clearance via Cockcroft–Gault, a six-category diagnosis grouping, and merged calculated panel-reactive antibody [CPRA]). One variable (pulmonary vascular resistance) was subsequently removed for complete missingness in at least one cohort split, leaving 56 candidates for imputation (see Missing Data Imputation below).

Second, variable importance ranking was performed on the training cohort using a separate XGBoost model fitted with grid search hyperparameter optimization (learning rate, tree depth, and number of estimators; 27 combinations, five-fold cross-validation scored by AUROC) and SHAP-based feature importance (mean absolute SHAP values across training samples). This feature-selection XGBoost is distinct from the XGBoost source model later evaluated by PRiSM (coarser grid search on all 56 candidates for ranking only, versus Optuna-tuned on the final 22-predictor set; Table S7) and was applied identically upstream of all 4 trained model families. The top-ranked predictors by SHAP importance were retained, yielding 22 final predictors for the trained models (Table 1 in main text; Table S15). Transplant year was also carried forward to verify the temporal cohort split but was not used as a predictor. Raw serum creatinine was excluded to avoid collinearity with creatinine clearance.

The 57 expert-selected candidates included all variables used by the IMPACT scoring model. Because IMPACT is a pretrained model with fixed published coefficients, it was instead applied using its own variable definitions (12 predictors; Table S8). Nine variables overlap between the two sets; IMPACT additionally uses recipient sex, race or ethnic group, and intraaortic balloon pump (IABP), while the trained models include donor weight, HCV seropositivity, ICD, and other variables not in IMPACT (see Table S15 footnotes).

**Missing Data Imputation.** Missing values were imputed using an iterative imputation procedure following the MissForest algorithm [S2], which handles mixed-type data and preserves relationships between variables. The imputation model was trained on the training cohort and applied unchanged to the tuning and external-validation cohorts to avoid information leakage. Our implementation substituted scikit-learn's `HistGradientBoostingRegressor` for the random forest estimator used in the original MissForest algorithm; this substitution preserves the iterative chained-equations structure of MissForest with faster computation on large datasets. Variables with 100% missingness in any cohort were dropped. A single imputation was used rather than multiple imputation; the rationale for this choice and its limitations—particularly for predictors with high missingness rates—are discussed in Methodological Considerations below.

**Variable Transformations.** Continuous predictors were median-centered and scaled by twice their standard deviation; binary variables remained as 0 (no) and 1 (yes) and were excluded from scaling. Multi-category variables (primary diagnosis, VAD device type, diabetes mellitus, medical condition at transplant, and hepatitis C virus serostatus) were dummy encoded—i.e., each category was represented as a binary indicator with the most common category serving as the reference and dropped from the data to avoid collinearity (reference categories: nonischemic cardiomyopathy, no VAD, no diabetes, not hospitalized, and negative, respectively). Dummy encoding was used uniformly, even for variables with partially ordinal structure (e.g., diabetes type, medical condition at transplant), because the functional ANOVA decomposition estimates a separate effect for each category relative to the reference and does not assume monotonic relationships between them. This approach allows the nomogram to capture category-specific risk contributions without imposing an ordering constraint that may not hold for all categories within a variable (e.g., diabetes "other" has no natural ordinal position relative to type I and type II).

## S1.3 Source Model Development

Five classification models for predicting 1-year mortality were transformed into nomograms using PRiSM: a pretrained clinical scoring model (Index for Mortality Prediction After Cardiac Transplantation, IMPACT) [S3], logistic regression (LR) [S4], neural network (multilayer perceptron, MLP) [S5], random forest (RF) [S6], and XGBoost (XGB) [S7]. The IMPACT model is a published clinical score whose reasoning is explicitly defined by variable thresholds and coefficients. The other 4 were trained on the training cohort. Training hyperparameters were selected using the tuning cohort. Standard training configuration across all models is detailed in Table S3; PRiSM decomposition and LASSO configuration in Table S9.

**Logistic Regression and MLP.** The logistic regression and MLP models were implemented in PyTorch with Adam optimization and early stopping based on tuning cohort loss. The MLP employed a single hidden layer with tanh activation to capture nonlinear relationships while maintaining tractable complexity. Hidden layer size and other hyperparameters were selected via Optuna tuning (Tables S4 and S5).

**Tree-Based Models.** Both tree-based models used the XGBoost library: XGBoost (XGBClassifier) for gradient boosting and the random forest via XGBRFClassifier, which implements bagging with per-node column sampling analogous to sklearn's RandomForestClassifier. Hyperparameters including tree depth, number of estimators, subsampling ratios, and (for XGBoost) learning rate were selected via Optuna tuning (Tables S6 and S7). Model-specific hyperparameters and implementation details are provided in Tables S4 to S8.

**Hyperparameter Tuning.** Hyperparameters for all 4 trained models (LR, MLP, RF, XGB) and the subsequent Partial Response Networks (PRNs; see Section S1.4.5) were selected using Optuna [S8] with the Tree-structured Parzen Estimator (TPE) sampler running 8 parallel jobs, optimizing tuning cohort AUROC. The number of trials was scaled to each model's search-space complexity and per-trial training cost: 256 trials for the tree-based models (XGB and RF), which have larger search spaces including regularization parameters and faster per-trial evaluation; 128 trials for MLP and PRN; and 64 trials for LR, whose three-parameter space converges quickly. Known-good default parameters were enqueued as the first trial to ensure the baseline configuration was always evaluated. Within each trial, the model was trained on the training cohort and evaluated on the tuning cohort; the trial with the highest tuning AUROC was selected. Search spaces for each model type are detailed in Tables S4 to S7 (source models) and Table S10 (PRN).

## S1.4 PRiSM Methodology

### S1.4.1 Mathematical Foundation

The core principle is that any classifier's log-odds predictions can be decomposed into interpretable additive components, i.e., a generalized additive model [S9]:

$$\log \frac{P(C \mid \mathbf{x})}{1 - P(C \mid \mathbf{x})} = \beta_0 + \mathrm{score}(x_1, x_2, \ldots, x_d) \tag{1}$$

where $P(C \mid \mathbf{x})$ is the probability of class $C$ given input variables $\mathbf{x}$, $\beta_0$ is the intercept, and $x_1, x_2, \ldots, x_d$ are the $d$ input variables. Through functional analysis of variance (ANOVA) decomposition applied to the black-box classifier [S10], the score function can be exactly decomposed into:

$$\mathrm{score}(x_1, \ldots, x_d) = \sum_i \phi_i(x_i) + \sum_{i \neq j} \phi_{ij}(x_i, x_j) + \cdots + \phi_{1 \ldots d}(x_1, \ldots, x_d) \tag{2}$$

where $\phi_i(x_i)$ represents the univariate main effect of variable $i$, $\phi_{ij}(x_i, x_j)$ represents the bivariate interaction between variables $i$ and $j$, and higher-order terms capture interactions among multiple variables. The component terms, i.e. the "partial responses," contain non-overlapping contributions and are generally nonlinear. The sum exactly matches the log-odds of the model prediction.

Here, we retain only the univariate main effect and bivariate interaction terms. This is a trade-off between interpretability and fidelity to higher-order interactions that may be present in the source model.

### S1.4.2 Lebesgue ANOVA Decomposition

Two decompositions of the score function are available. The Dirac method isolates each variable's effect by holding all others at their reference value, which for the centered and scaled data used here is the median; it scales linearly with the number of observations, but it evaluates the model at a single fixed point and discards the correlation structure between predictors. The Lebesgue method, used for every result reported in this study, instead integrates over the empirical distribution of the remaining predictors, preserving that correlation structure at the cost of quadratic scaling.

For each individual observation, the variable of interest (or pair of variables) is fixed at observed value $x_i$ (or $(x_i, x_j)$ for bivariate interactions) across all observations in the dataset, and the average model prediction is calculated. To evaluate the effect of variable $X_i$ at the observed value $x_i$, the univariate effect:

$$\phi_i(x_i) = \mathbb{E}[f(\mathbf{X}) \mid X_i = x_i] - \mathbb{E}[f(\mathbf{X})] \tag{3}$$

where $f$ is the black-box model's log-odds prediction function, $\mathbf{X}$ the full set of input variables, $X_i$ the variable at the $i$-th input and $x_i$ one of its observed values, and $\mathbb{E}[f(\mathbf{X})]$ the mean log-odds prediction over the dataset, subtracted as a baseline so that $\phi_i$ isolates the effect of $X_i$ taking the value $x_i$.

Similarly for bivariate effects at specific values $x_i$ and $x_j$:

$$\phi_{ij}(x_i, x_j) = \mathbb{E}[f(\mathbf{X}) \mid X_i = x_i, X_j = x_j] - \phi_i(x_i) - \phi_j(x_j) - \mathbb{E}[f(\mathbf{X})] \tag{4}$$

where the conditional expectation now fixes both variables at their observed values while the rest vary according to their empirical distribution.

The Lebesgue method requires a representative dataset to yield meaningful effects. We used the training cohort.

### S1.4.3 LASSO Partial Response Selection

The inclusion of both univariate and bivariate (interaction) partial responses leads to a considerable increase in dimension, with $d(d-1)/2$ bivariate responses per patient in addition to the $d$ univariate responses. To manage this complexity, we employ the least absolute shrinkage and selection operator (LASSO) to select partial responses for inclusion in the nomogram [S11]. Specifically, we use $L_1$-penalized logistic regression (sklearn `LogisticRegression` with `penalty='l1'`, `solver='saga'`, and warm start from the previous $\lambda$ solution). A logarithmically spaced grid of $\lambda$ values is evaluated from strong to weak regularization. The logistic regression is fit using the partial responses as features and the observed training-cohort outcome (1-year mortality) as the target, so that the beta coefficients reflect each partial response's contribution to predicting the outcome. Only partial responses with $|\beta| > 0.1$ are retained. This threshold is a practical noise filter to exclude near-zero coefficients that LASSO has not fully shrunk to zero due to finite convergence tolerance; it is not intended as a tuned hyperparameter. Because the partial responses are constructed on the log-odds scale from scaled predictors, a coefficient of 0.1 corresponds to a negligible contribution to the predicted log-odds. The primary mechanism controlling model complexity is the $\lambda$ regularization strength, which determines how aggressively the partial responses are pruned, with higher $\lambda$ values resulting in sparser models (main text Figure 3). Full LASSO configuration is detailed in Table S9.

**Categorical Predictor Handling.** Partial responses for categorical predictors were represented as a single feature column in the LASSO selection matrix, with each sample assigned its category-specific partial response value as learned by the model. This preserves all category-level information while maintaining variable-level selection: LASSO determines whether a categorical predictor contributes to the outcome ($\beta \neq 0$), and if selected, all category effects are retained in the final nomogram with uniform scaling by $\beta$.

Grouped handling applies at the decomposition step as well. Each dummy-encoded categorical is treated as one variable throughout: its partial response is evaluated one category at a time, with the dropped reference category serving as the baseline, so the value reported for each category is its effect relative to that reference. Bivariate terms are formed between variables, not dummy columns, and pairs drawn from within a single categorical are excluded by construction, since they would require two categories of the same variable to be present simultaneously. In the counts above, $d$ is therefore the number of predictors, not the larger number of dummy-encoded columns.

**Lambda Selection Variants.** The baseline and sparse $\lambda$ strategies are defined in main text Methods. Both were applied to all 5 models across the full two-phase pipeline, with AUROC results in main text Table 2 and Table S14. The baseline $\lambda$ was used at both LASSO stages; under the sparse strategy, Phase II refinement continued from the sparse Phase I selection but took its own $\lambda$ from the baseline criterion. Each strategy therefore yields a Phase I nomogram and its PRN-refined counterpart, four per source model; the sparse PRN nomogram for MLP is shown in main text Figure 4**B**. The two $\lambda$ positions for MLP and LR are marked in main text Figure 3; notably, the LR decomposition (main text Figure 3**B**) selects only univariate terms because the linear model cannot produce interactions.

#### S1.4.4 Nomogram Prediction Calculation

Each selected partial response contributes a weighted component ($\beta \times \phi(\mathbf{x})$) to the final log-odds prediction. The nomogram prediction equals the sum of all these weighted components plus the intercept:

$$\text{logit} = \beta_0 + \sum \beta_k \, \phi_k(\mathbf{x}) \tag{5}$$

This additive structure allows manual calculation by reading individual contributions from the nomogram and summing them to obtain the total log-odds, which is then converted to probability via logistic transformation. A worked example demonstrating this calculation for an individual patient is provided in Figure S4.

#### S1.4.5 Partial Response Network Refinement

The source model and the observed outcome each contribute to the two phases of the PRiSM pipeline, which differ in what they inherit from the source model and what they learn from the outcome:

- **Phase I**: the functional ANOVA decomposition derives each partial response entirely from the source model's input–output behavior; the outcome labels are not used in the decomposition. The subsequent LASSO selection step, however, fits an $L_1$-penalized logistic regression of the partial responses against the observed training-cohort outcome labels. The partial response *shapes* (nonlinearities and interaction surfaces) therefore remain fixed as derived from the source model, while the outcome labels govern *which terms are selected* and *how they are weighted*.

- **Phase II**: the partial responses selected in Phase I define the architecture of a structured neural network (the Partial Response Network, PRN), in which each subnetwork receives only the one or two variables corresponding to a single selected effect. Higher-order interactions are excluded by construction. The PRN is trained from scratch on the training cohort mortality outcome labels, then redecomposed by functional ANOVA and subjected to a second LASSO selection. The source model's contribution is thus reduced to the *architectural scaffold*—which variables and interactions are permitted—while the PRN re-learns the response shapes from the outcome.

Across both phases, the selected partial responses function as the model's concepts, in the sense of a concept-bottleneck model: the source model supplies a set of candidate concepts and, in Phase II, the scaffold of permitted variables and interactions, while the observed outcome governs which concepts are retained and—in Phase II—their final shapes and weights.

Phase I nomograms accordingly provide the strongest fidelity to the source model's learned relationships, and are the representation in which that structure can be read and audited. Phase II nomograms are not a weaker approximation of the same target: because the decomposition is truncated at second order, the Phase I responses are estimated in the presence of the source model's unmodeled third- and higher-order interactions, and re-estimating them in a network confined to the selected univariate and bivariate effects removes much of that contribution. This was particularly beneficial for sparse nomograms (see Empirical Justification below). The structured architecture used 5 hidden nodes per selected feature with tanh activation, trained using Adam optimization with early stopping based on tuning cohort loss to prevent overfitting (Tables S10 and S11).

**Empirical Justification.** In Phase I the LASSO operates on the full set of decomposed partial responses, including all $d(d-1)/2$ bivariate interaction terms, and the train–tuning AUROC gap widens as the penalty weakens and more of them enter (main text Figure 3**A**). The PRN removes that complexity–overfitting trade-off by construction: higher-order interactions beyond those selected in Phase I cannot enter the model, so the PRN-refined LASSO paths (main text Figure 3**C**, **D**) show tuning-cohort AUROC largely insensitive to $\lambda$, with a fixed term count throughout. The sparse Phase I selection therefore identifies a core predictor set that the constrained architecture preserves without further pruning; refinement in that architecture recovered noninferiority for 2 of the 5 source models (main text Table 2). The sparse PRiSM nomograms (Figure S1) illustrate how PRiSM preserves each model's characteristic response patterns; the corresponding sparse PRN nomograms after Phase II refinement are shown in Figure S2. The complete baseline PRiSM nomogram for the MLP model is shown in Figure S3.

## S1.5 Statistical Analysis

Model training and PRiSM conversion were implemented in Python 3.12 (PyTorch, scikit-learn, XGBoost). All statistical evaluation was performed in R version 4.5.2. The key R packages used were `pROC` (discrimination and paired comparisons) and `stats`, with `rms`, for calibration; bootstrap percentile intervals used ordinary nonparametric case resampling, parallelised with the `parallel` package under a per-replicate seed so that an interval reproduces exactly regardless of worker count and platform.

**AUROC Calculation.** AUROC was calculated using the `pROC` package [S12] (`pROC::roc()`, `pROC::auc()`), with 95% confidence intervals (CI) from `pROC::ci.auc()`, which for a full non-smoothed AUC is the DeLong interval. Paired difference in AUROC (nomogram minus source model) 95% CI also used the DeLong method for correlated ROC curves, to account for the same patients being evaluated by both the source model and its nomogram transformation. The interval is analytic: writing $V_n$ and $V_s$ for the DeLong variances of the two AUROCs and $C$ for their covariance, obtained from `pROC::var()` and `pROC::cov()` under a single row mask so that the pairing holds, the difference $D$ has $\mathrm{Var}(D) = V_n + V_s - 2C$. `pROC::roc.test()` was run separately and supplies only the two-sided $z$ and $p$, as an independent check on the same quantity. Differences against the de novo comparators (main text Table 3, Tables S17 to S19) use the same construction.

**Noninferiority.** The primary endpoint definition is given in main text Methods. Noninferiority was assessed on the retention-adjusted contrast $T = \mathrm{AUROC}_{\mathrm{nomogram}} - 0.90 \times \mathrm{AUROC}_{\mathrm{source}} - 0.05$, which is estimated from the same paired DeLong covariance matrix as $\mathrm{Var}(T) = V_n + 0.81\, V_s - 1.8\, C$, and was met when the lower bound of the two-sided 95% CI for $T$ exceeded zero.

**Calibration Assessment.** For each cohort, model, and transformation stage, we treated model probabilities as fixed and assessed calibration without recalibrating models on validation data. Calibration-in-the-large (CITL) was estimated as the intercept from a logistic regression with outcome $Y$ and an offset equal to the log-odds of predicted probabilities: `glm(Y ~ 1, family=binomial, offset=lp)`, where `lp` represents the linear predictor and 0 indicates perfect level calibration. The calibration slope was estimated as the coefficient of `lp` in `glm(Y ~ lp, family=binomial)`, where 1 indicates ideal spread; values $<$1 suggest prediction compression. Wald 95% confidence intervals were derived from model-based variance–covariance estimates for CITL and calibration slopes (`stats::glm()`; `rms::lrm()` used for cross-validation in early pipeline stages).

**Observed-to-Expected Ratio.** The observed-to-expected (O:E) ratio was calculated as the ratio of observed events to the sum of predicted probabilities (expected events). A ratio of 1.0 indicates perfect calibration-in-the-large. Bootstrap percentile 95% confidence intervals were derived from 1,000 resamples of the cohort with replacement, with the interval taken as the 2.5th and 97.5th percentiles of the replicate statistics (`stats::quantile`).

**Brier Score.** The Brier score quantified overall accuracy as the mean squared difference between predicted probabilities and observed binary outcomes, with bootstrap percentile confidence intervals from the same 1,000-replicate resampling used for the O:E ratio. Lower values indicate better performance.

**Decision Curve Analysis.** Decision curve analysis [S13] quantified net benefit—the difference between the proportion of true positives and a weighted proportion of false positives at a given threshold probability—across clinically relevant thresholds (5–20% for post-transplant mortality interventions) to confirm that noninferior discrimination translated to preserved clinical utility. Decision curves for the source models, baseline PRiSM nomograms, and sparse PRN nomograms in external validation are shown in main text Figure 2, panels B through D.

**Results Tables and Figures.** Discrimination results are reported in Table S12 (training), Table S13 (tuning), and Table S14 (external validation); the external validation results are summarized in main text Table 2. Calibration metrics and Brier scores are reported in Table S16, and decision curves in main text Figure 2, panels B through D.

## S1.6 Software Implementation

PRiSM is released as an open-source package under the BSD 3-Clause licence. Core dependencies include NumPy, PyTorch (with optional CUDA or MPS acceleration), pandas, scikit-learn, and matplotlib for visualization. The package supports cross-platform deployment on Windows, macOS, and Linux. Batching, vectorization, and caching were used to accelerate the computationally intensive Lebesgue ANOVA decomposition. The package, its complete requirements, and demonstration notebooks are at `https://github.com/AIBCTS/PRiSM`; the version used in this study is archived on Zenodo [S14].

## S1.7 De Novo Additive Comparators

PRiSM nomograms were compared with interpretable additive models trained de novo on the same temporal split. Their performance is included in Table S16; paired discrimination comparisons with PRiSM models are in Tables S17 to S19.

Every comparator is fitted over the same 22 predictors as the nomograms, so the term counts are on one scale (Table S16). EBM and the GAM take a multi-category predictor as one variable, as a nominal feature and as a factor, each level carrying its own free value. A NAM's feature networks each take a single scalar input [S15], so NAM was given the dummy-encoded columns of Section S1.2, one network per column: their sum again gives each level its own free value, where a single network over level codes would impose an order the levels do not have. All therefore span the same function class over an additive term, and differ instead in what a feature-ranking step sees: EBM's interaction search ranks 231 pairs of grouped predictors.

Comparators were tuned against the same objective and on the same cohort as the source models, by grid search over each package's principal capacity and regularization settings; the external-validation cohort was never loaded during a search. Each grid below is the cross-product of the key settings that package exposes. The GAM's fitter is deterministic, so its grid was enumerated at one fit per point; EBM and NAM vary across random initialization and bagging, so each of their configurations was fitted at several seeds and compared seed by seed. EBM's larger grid was sampled. A change was adopted only where it separated from the configuration it would replace.

**Explainable Boosting Machine (EBM).** EBM was implemented via InterpretML's `ExplainableBoostingClassifier` (Python, interpret 0.7.6) [S16, S17], which includes main effects and automatically detected pairwise interactions. It was fitted with 128 bins (32 for interaction terms), 20 interaction terms, 8 outer bags, learning rate 0.01, a minimum of 2 samples per leaf, 500 smoothing rounds, and early stopping after 50 rounds against an internal 20% split of the training cohort; settings not named here were left at their package defaults. Sampling 150 configurations from a 4,050-point grid over learning rate, tree depth, interaction count, minimum leaf size, bin count, smoothing rounds and bagging produced nothing that separated from it.

**Generalized Additive Model (GAM).** GAM was implemented via R's `mgcv` package 1.9-3 with a logistic link [S9, S18], using univariate smooth terms for the continuous predictors and parametric terms for the binary and categorical predictors. Smooths used basis dimension $k = 5$ on a cubic regression spline basis, with smoothing parameters estimated by REML and `mgcv`'s remaining defaults unchanged; its additional null-space penalty was fitted as an alternative and was not selected on the tuning cohort. A 72-point enumerated grid over basis dimension, basis type, smoothness-selection criterion, the smoothness inflation factor and that null-space penalty displaced none of these settings.

**Neural Additive Model (NAM).** NAM was implemented via the nam Python package [S15], which models main effects only by architecture. Each feature network is a single 32-unit ExU layer, and dropout at 0.1 was the only regularization applied; optimization used Adam at learning rate $10^{-3}$ with batch size 256, for at most 120 epochs, with tuning-cohort AUROC scored after every epoch and training stopped after 20 without improvement. A 24-point grid over per-feature layer width, additional hidden layers, and dropout put the optimum at the widest network searched; extending the search past that width confirmed a turnover there, with wider networks scoring lower.

## S1.8 Sensitivity and Subgroup Analyses

### S1.8.1 Seed Stability of PRiSM Conversion

To assess the reproducibility of the PRiSM conversion process, we repeated the full pipeline (Phase I decomposition, LASSO selection, and Phase II PRN refinement) with 3 independent random seeds (42, 123, 257) while holding the source models fixed. Seed variation affects the LASSO solver (`saga` stochastic gradient descent), PRN weight initialization, and PRN hyperparameter tuning (Bayesian optimization); the functional ANOVA decomposition itself is deterministic.

**Discrimination.** AUROC point estimates were highly stable across seeds, with a maximum range of 0.003 across all 20 nomograms (Table S20). Seed-to-seed variation was 1–2 orders of magnitude smaller than the width of the DeLong 95% confidence intervals (≈0.040), confirming that stochastic variation in the conversion pipeline is negligible relative to the statistical precision of the estimates. All point estimates from different seeds fell well within each other's confidence intervals. The 3-seed ensemble (patient-level averaged predictions) yielded AUROCs consistent with the individual seeds (Table S23).

**Noninferiority.** All 20 nomograms produced identical noninferiority conclusions across all 3 seeds and the 3-seed ensemble (20/20 consistent; Table S20). The 10 baseline PRiSM and PRN nomograms passed noninferiority uniformly; among the 10 sparse-regularized variants, 8 consistently failed while LR Sparse PRN and XGB Sparse PRN consistently passed.

**Selected Terms.** Phase I PRiSM decomposition selected identical term sets across seeds for all 5 models (Jaccard similarity coefficient = 1.00; higher values indicate greater term-set overlap, 1.00 = identical; Table S20). Phase II PRN refinement introduced modest variation in 3 of the 5 models: MLP PRN selected 19–26 terms across seeds (Jaccard = 0.67), RF PRN 15–18 terms (Jaccard = 0.83), and LR PRN 17–19 terms (Jaccard = 0.89); IMPACT PRN and XGB PRN were identical across seeds. For MLP and RF these differences reflect toggling of borderline interaction terms; the LR selection is univariate at every penalty, so its variation is in which main effects sit at the selection boundary. None materially affected discrimination (AUROC ranges $\leq$0.003). In the sparse nomograms—the deployment candidates where parsimony matters most—term identity was stable: 8 of the 10 sparse nomograms had Jaccard = 1.00, and the remaining two (LR Sparse PRiSM and LR Sparse PRN) differed by a single term (prior blood transfusions, included only in seed 42; Jaccard = 0.90).

**Calibration.** Calibration metrics showed somewhat greater seed sensitivity than discrimination, particularly in PRN stages. Calibration slope ranges across seeds were $\leq$0.06 for PRiSM stages but up to 0.12 for PRN MLP (Tables S21 and S22). This is consistent with the stochastic nature of PRN weight initialization and training. O:E ratios were more stable (ranges $\leq$0.02 across seeds).

**Full Results.** Full performance tables for each seed are reported in Table S21 (seed 42), Table S22 (seed 123), and Table S23 (3-seed ensemble). Seed 257 results appear in main text Table 2.

### S1.8.2 Sensitivity to Transplant Year Inclusion

Transplant year captures trends in surgical technique, donor management, and immunosuppression protocols. Including it as a predictor may improve calibration by adjusting for temporal shifts in baseline mortality, but risks data leakage in temporal validation designs and may reduce the model's transportability to future eras. We assessed the impact of including transplant year by comparing the full pipeline with and without this variable, using seed 257 for both configurations.

**Discrimination.** Including transplant year had negligible effects on discrimination (Table S24). All 20 AUROC confidence intervals overlapped between the two configurations, with point estimates differing by at most 0.019 (LR Sparse PRN: 0.625 vs 0.644)—well within the width of the DeLong 95% CIs ($\approx$0.040). IMPACT models were unaffected (identical AUROCs across all steps), as expected since the IMPACT score formula does not include transplant year.

**Noninferiority.** Noninferiority conclusions agreed for 19 of the 20 nomograms (Table S24). The single disagreement was MLP Sparse PRN, which failed noninferiority without transplant year (lower bound of the 95% CI for $T = -0.0003$) but passed with it (lower bound $= +0.006$). This is a borderline case: without transplant year the bound falls short of zero by less than a thousandth of an AUROC unit. Each configuration is judged against its own source model, so the contrast being tested differs slightly between configurations for models whose source AUROC changed.

**Calibration.** Including transplant year moved O:E ratios and CITL closer to their targets for most non-IMPACT nomograms, with the largest gains in the source models themselves (LR 0.823 to 1.018, MLP 0.808 to 1.052); two sparse PRN nomograms were the exceptions, MLP no closer to 1.0 (0.891 to 1.108) and XGB further from 1.0 (0.902 to 1.140; Table S25). This is expected: transplant year captures secular trends in baseline mortality, allowing models to recalibrate across transplant eras in the external-validation cohort. Calibration slope showed mixed results with no consistent direction of improvement. This improvement in calibration comes at the cost of temporal transportability—models with transplant year require knowledge of the transplant era to generate predictions.

**Model Complexity.** Including transplant year changed model complexity modestly for most nomograms ($-1$ to $+2$ terms), with 3 exceptions: PRiSM XGB gained 9 terms and PRN XGB 6, suggesting transplant year interacts with many features in the XGBoost model, while PRiSM MLP lost 5 (Table S24).

**Full Results.** Full performance results with transplant year included are reported in Table S25.

### S1.8.3 Exploratory Subgroup Analysis

Performance was examined within subgroups of the external-validation cohort defined by recipient sex and by race or ethnic group. Neither variable is among the 22 predictors, so for every model but IMPACT—whose published definition includes both—this is an audit of an attribute the model never saw. Subgroup AUROC, CITL and calibration slope use the estimators of Section S1.5. Discrimination by group is reported in Table S26. No multiplicity adjustment was applied, and the Other category, at 54 deaths, supports no finding of its own.

**Between-Group Differences.** Each level was compared with the reference level of its variable, White or Male, as a difference in AUROC. The two strata are disjoint sets of patients, so the difference takes the unpaired variance $V_\ell + V_{\mathrm{ref}}$ with no covariance term, and a two-sided Wald 95% CI. One-year mortality differed by less than half a percentage point across White, Black and Hispanic recipients, so the difference in discrimination is not an artifact of outcome prevalence.

**Effect of Translation.** Within one group a nomogram and its source model are scored on the same patients, so the change in AUROC there is the paired quantity of Section S1.5 computed within a stratum. The comparison reported is that change in a group minus the same change in the reference group, whose variances add because the groups share no patients. The median 95% CI half-width over the 80 comparisons—4 nomogram versions $\times$ 5 source models $\times$ 4 non-reference groups—is 0.028, so a change large enough to remove a between-group difference of 0.030 to 0.069 AUROC would have separated from zero.

**Calibration Heterogeneity.** Heterogeneity was assessed by an omnibus likelihood-ratio test per model and subgroup variable, adding a `group` term to the CITL and calibration-slope models of Section S1.5, on $k-1$ degrees of freedom for $k$ levels. The slope test nests against a model that already allows the groups separate intercepts, so it isolates a difference in slope from a difference in level. Three of 112 tests reached $p < 0.05$, none on a translated nomogram: IMPACT's CITL by race or ethnic group ($p = 0.002$) and by sex ($p = 0.004$), and NAM's calibration slope by race or ethnic group ($p = 0.017$; 0.40 in Black against 0.83 in White recipients). The only source model flagged is the only one whose definition uses either variable. With 112 unadjusted tests, that count is near what chance alone would produce.

## S1.9 Supplementary Findings

**Calibration Improvement After Translation.** In external validation, the IMPACT source model showed a calibration slope of 0.602 (95% CI, 0.502 to 0.701), indicating substantial prediction compression, whereas its baseline PRiSM nomogram achieved a slope of 0.988 (95% CI, 0.837 to 1.139). This improvement is attributable to the PRiSM decomposition procedure, as opposed to explicit recalibration. IMPACT is a pretrained scoring model with fixed published coefficients that were not updated for the study population; the Lebesgue ANOVA decomposition re-expresses the model's predictions in terms of partial responses averaged over the empirical training-cohort distribution, and the subsequent LASSO reweighting of these responses effectively adjusts the prediction scale. This mechanism is implicit in the decomposition–reweighting pipeline and applies to all source models, but its effect is most visible for IMPACT because the source model was the most poorly calibrated. The recalibration is a byproduct of the translation process, not a separate calibration step, and was not performed on validation data.

**Systematic Overprediction in External Validation.** All O:E ratios in the external-validation cohort were below 1.0 (range 0.642 to 0.917 across source models and translated nomograms), indicating systematic overprediction of 1-year mortality. This pattern is consistent with the temporal design: the external-validation cohort (2020–2022) reflects a later era with lower observed mortality than the training cohort (2000–2016), likely due to advances in immunosuppression, perioperative management, and donor selection. Because models were evaluated without recalibration, this temporal drift is expected to produce overprediction. The O:E ratios improved after PRiSM translation for most model families, consistent with the calibration-slope improvements described above. In any prospective deployment, periodic recalibration to the contemporary patient population would be expected as standard practice and is not specific to the PRiSM framework.

## S1.10 Methodological Considerations

**Single Imputation.** Missing values were handled with a single round of iterative imputation. Multiple imputation would propagate imputation uncertainty into downstream confidence intervals, but the PRiSM pipeline—which includes functional ANOVA decomposition, LASSO selection, and optional PRN training—does not lend itself to straightforward Rubin's-rules pooling across imputed datasets, because each imputed dataset could yield a different set of selected partial responses. Single imputation was therefore used for tractability. The potential impact is greatest for predictors with high missingness rates; serum albumin, which had 57.0% missingness in the tuning cohort and 25.9% in the training cohort, is the most affected variable. Its imputed values in the tuning cohort are largely model-generated, which may affect the reliability of lambda-selection decisions that depend on tuning-cohort AUROC. The seed-stability analysis reported in Section S1.8.1 above (Table S20) demonstrated that LASSO term selection was highly reproducible across random seeds, but this addresses stochastic variation in the post-imputation pipeline (LASSO solver, PRN initialization, and hyperparameter tuning), not the imputation uncertainty itself, since the imputed data were held fixed across seeds. The impact of alternative imputation draws on term selection was not evaluated and remains a limitation.

**Outcome Supervision in the PRiSM Pipeline.** Outcome labels supervise both LASSO stages and the PRN (main text Methods; Section S1.4.5), which lets the LASSO reweighting implicitly recalibrate the source model (Section S1.9). The trade-off is that the nomogram's coefficients are not guaranteed to reproduce the source model's exact risk estimates for individual patients, even when aggregate discrimination is preserved.

**Outcome-Independent Translation.** An alternative design would fit the LASSO against the source model's predicted log-odds instead of the observed outcome, giving a fully outcome-independent translation that requires only the source model and input data. It would preserve the source model's individual-level predictions, and propagate any miscalibration along with them. This may be desireable when a regulatory-approved model must be reproduced rather than replaced, or when a model is audited in a population whose outcomes are not yet observed. Comparison of the two approaches is a direction for future work.

**Marginal Averaging and Extrapolation.** The Lebesgue decomposition used throughout this study averages the source model over the empirical distribution of the remaining predictors. When predictors are correlated, this evaluates the model at combinations of predictor values that are rare or absent in the data. This is the extrapolation problem that marginal functional ANOVA raises for functions of dependent predictors: partial responses in sparsely observed regions are determined by model behavior that the data do not constrain, and in principle this could distort which terms the LASSO selects. Weighted formulations that restrict the decomposition to high-density regions address the problem, at the cost of a decomposition that is no longer defined over the full product space and that requires an estimate of the joint predictor density; Hooker (2007) develops this approach. Neither weighting nor density estimation was implemented here.

The Dirac decomposition is not a remedy: fixing the remaining predictors at their reference values evaluates the model at a single point that may itself be unobserved.

Two observations bound the practical concern for the models reported here. First, all 5 baseline PRiSM nomograms retained their source models' discrimination under temporal external validation, in a cohort whose case mix had shifted substantially from the training era; a decomposition materially distorted by extrapolation would not be expected to transfer across that shift. Second, term selection was reproducible across seeds (Section S1.8.1), so selection is not driven by unstable regions of the response surface, although that analysis holds the imputed data fixed and so does not address extrapolation directly. The magnitude of any residual effect was not quantified and remains a limitation.

## S2 Supplementary Figures

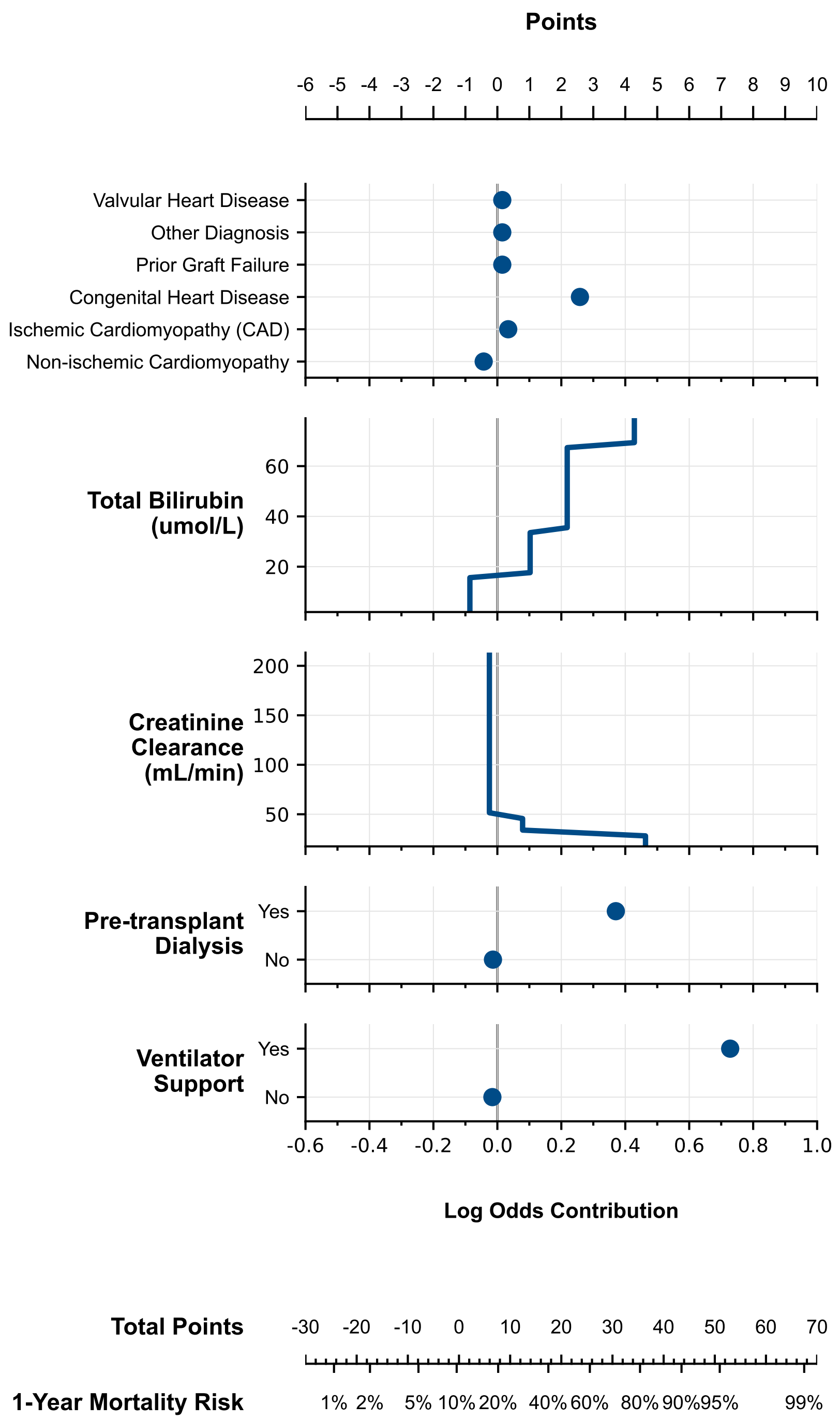

Figure S1. **Complete sparse PRiSM nomograms for all 5 model families, related to Figure 4.** Each page shows the nomogram for one model after Phase I PRiSM decomposition under the sparse $\lambda$. Rows show each predictor's points contribution (top scale) as a function of the predictor value; the bottom scale converts the summed total points to predicted 1-year mortality risk. Nomograms were derived from the source models fitted in the training cohort ($n = 32{,}543$), with $\lambda$ selected in the tuning cohort ($n = 8{,}558$); points are point estimates and carry no confidence intervals. **A,** IMPACT. IMPACT denotes Index for Mortality Prediction After Cardiac Transplantation; and PRiSM, Partial Responses in Structured Models.

**Partial Response Nomogram – LR**

**Points**

-6 -5 -4 -3 -2 -1 0 1 2 3 4 5 6 7 8 9 10

RVAD
Other
LVAD
BiVAD
No VAD

**Ventilator Support**
Yes
No

**Dialysis Dependence**
Yes
No

**Creatinine Clearance (mL/min)**
200
100

**Total Bilirubin (mg/dL)**
4
2

**Donor Age (years)**
40
20

**Total Ischemic Time (hours)**
6
4
2

**Body-Mass Index (kg/m2)**
30
20

**Serum Albumin (g/dL)**
4
2

-0.6 -0.4 -0.2 0.0 0.2 0.4 0.6 0.8 1.0

**Log Odds Contribution**

**Total Points**

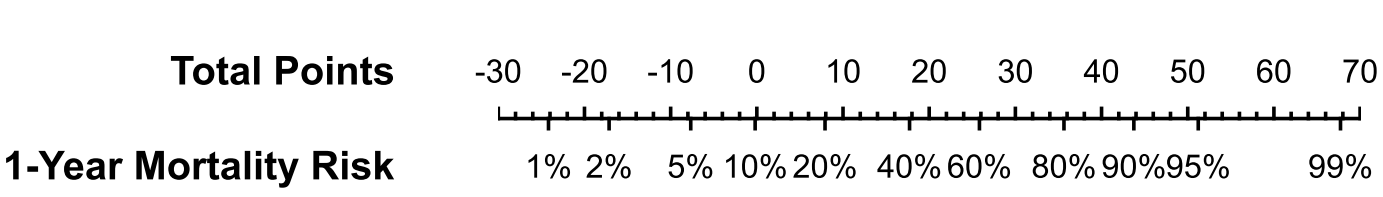

**1-Year Mortality Risk**

Figure S1. **B,** Logistic Regression (LR).

Partial Response Nomogram – MLP

Points

RVAD
Other
LVAD
BiVAD
No VAD

Valvular Heart Disease
Other Diagnosis
Ischemic Cardiomyopathy
Prior Graft Failure
Congenital Heart Disease
Nonischemic Cardiomyopathy

Ventilator Support
Yes
No

Dialysis Dependence
Yes
No

Creatinine Clearance (mL/min)

Total Bilirubin (mg/dL)

Donor Age (years)

Total Ischemic Time (hours)

Body-Mass Index (kg/m2)

Serum Albumin (g/dL)

Log Odds Contribution

Total Points

1-Year Mortality Risk

Figure S1. **C,** Multilayer perceptron (MLP).

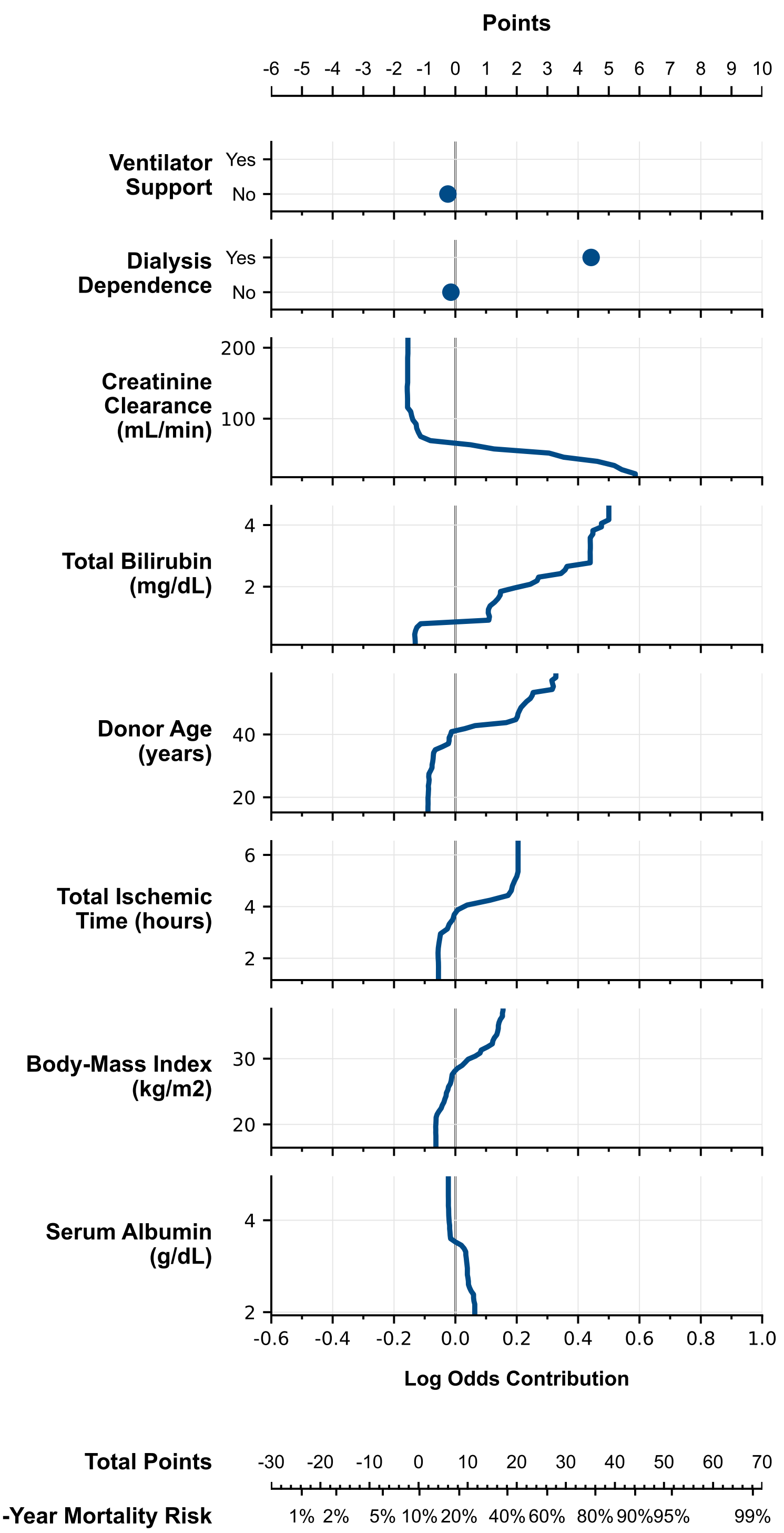

Figure S1. **D,** Random Forest (RF).

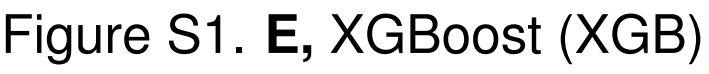

Figure S1. **E,** XGBoost (XGB).

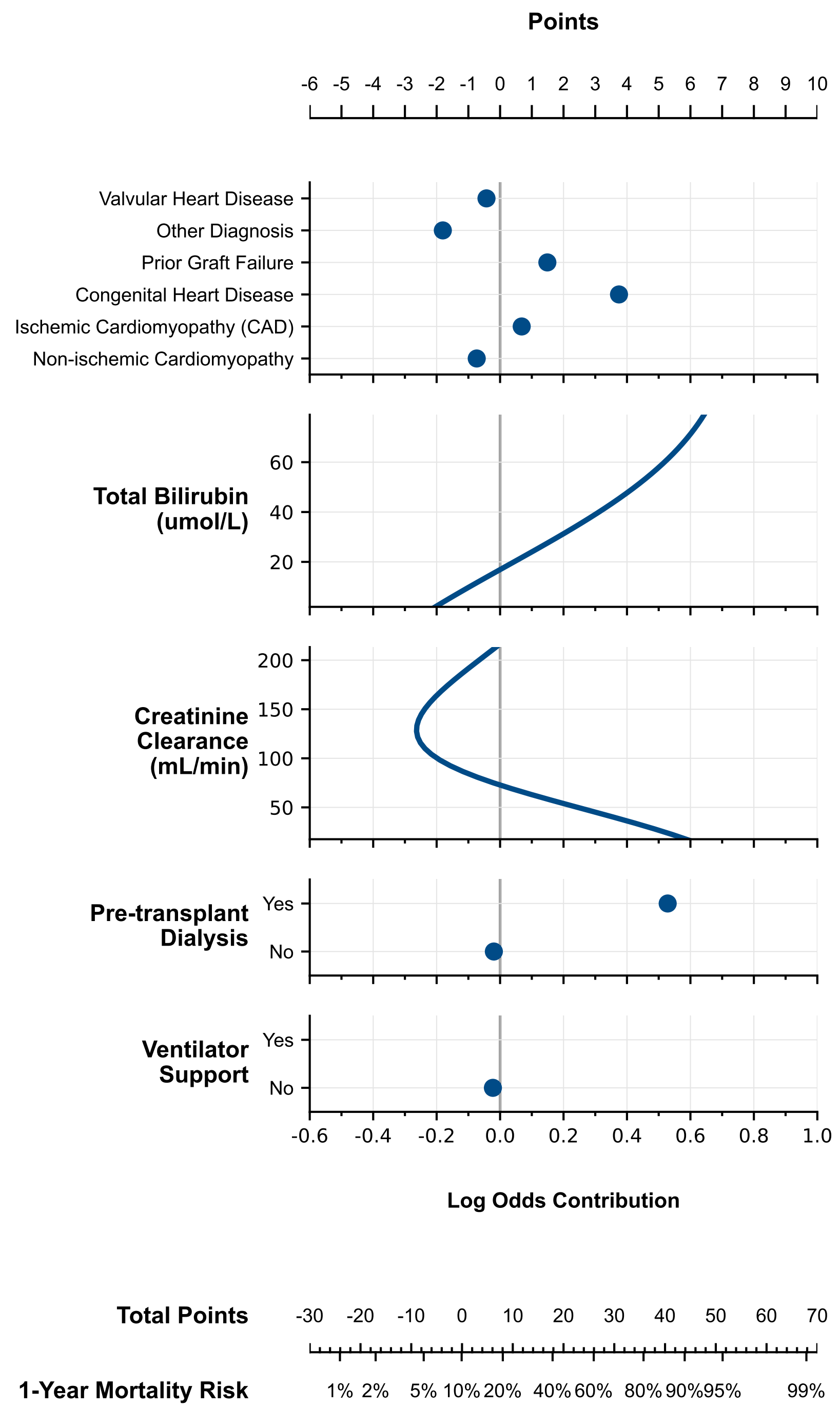

Figure S2. **Complete sparse PRN nomograms for all 5 model families, related to Figure 4.** Each page shows the nomogram for one model after Phase II PRN refinement under the sparse $\lambda$. Rows show each predictor's points contribution (top scale) as a function of the predictor value; the bottom scale converts the summed total points to predicted 1-year mortality risk. PRNs were trained in the training cohort ($n = 32{,}543$), with $\lambda$ selected in the tuning cohort ($n = 8{,}558$); points are point estimates and carry no confidence intervals. **A,** IMPACT. IMPACT denotes Index for Mortality Prediction After Cardiac Transplantation; and PRN, Partial Response Network.

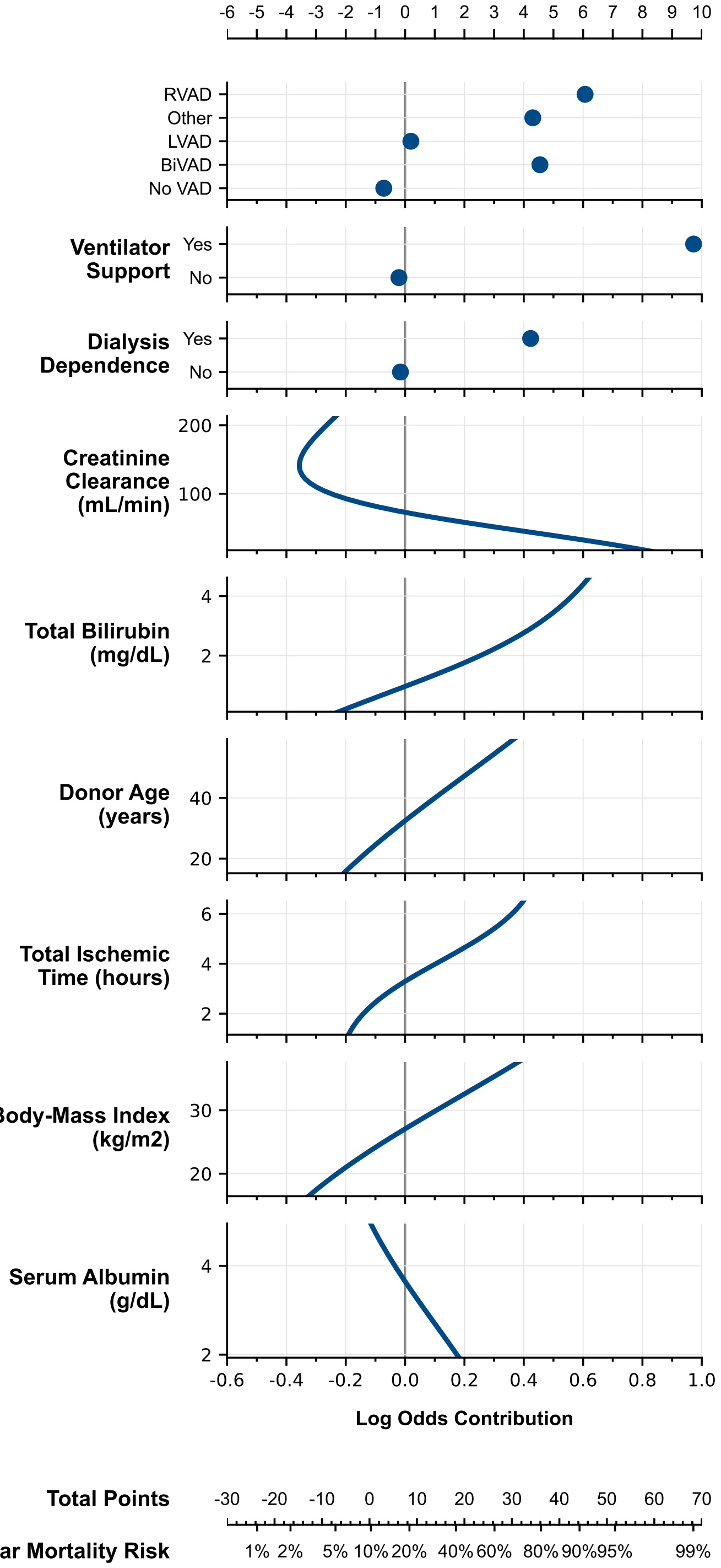

Figure S2. **B,** Logistic Regression (LR).

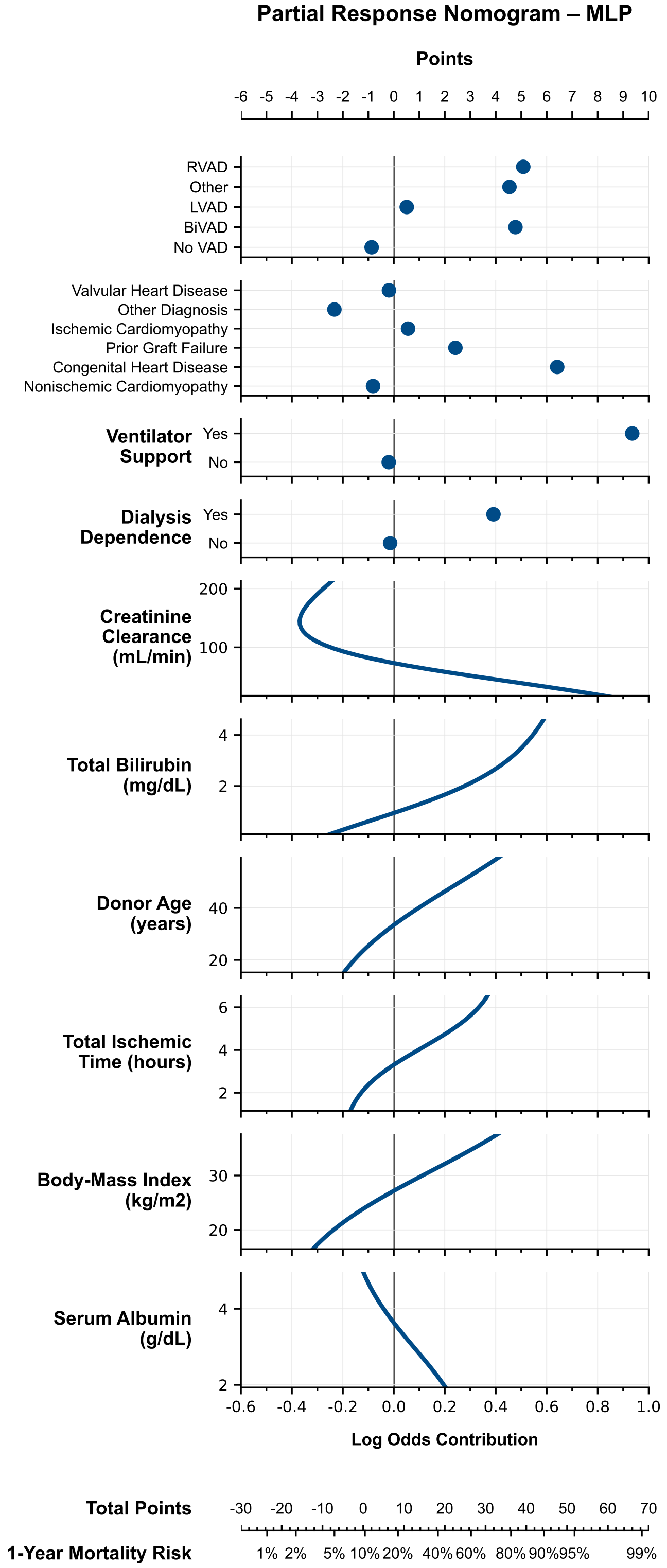

Figure S2. **C,** Multilayer Perceptron (MLP) (10 predictors; see main text Figure 4**B** and Table 2).

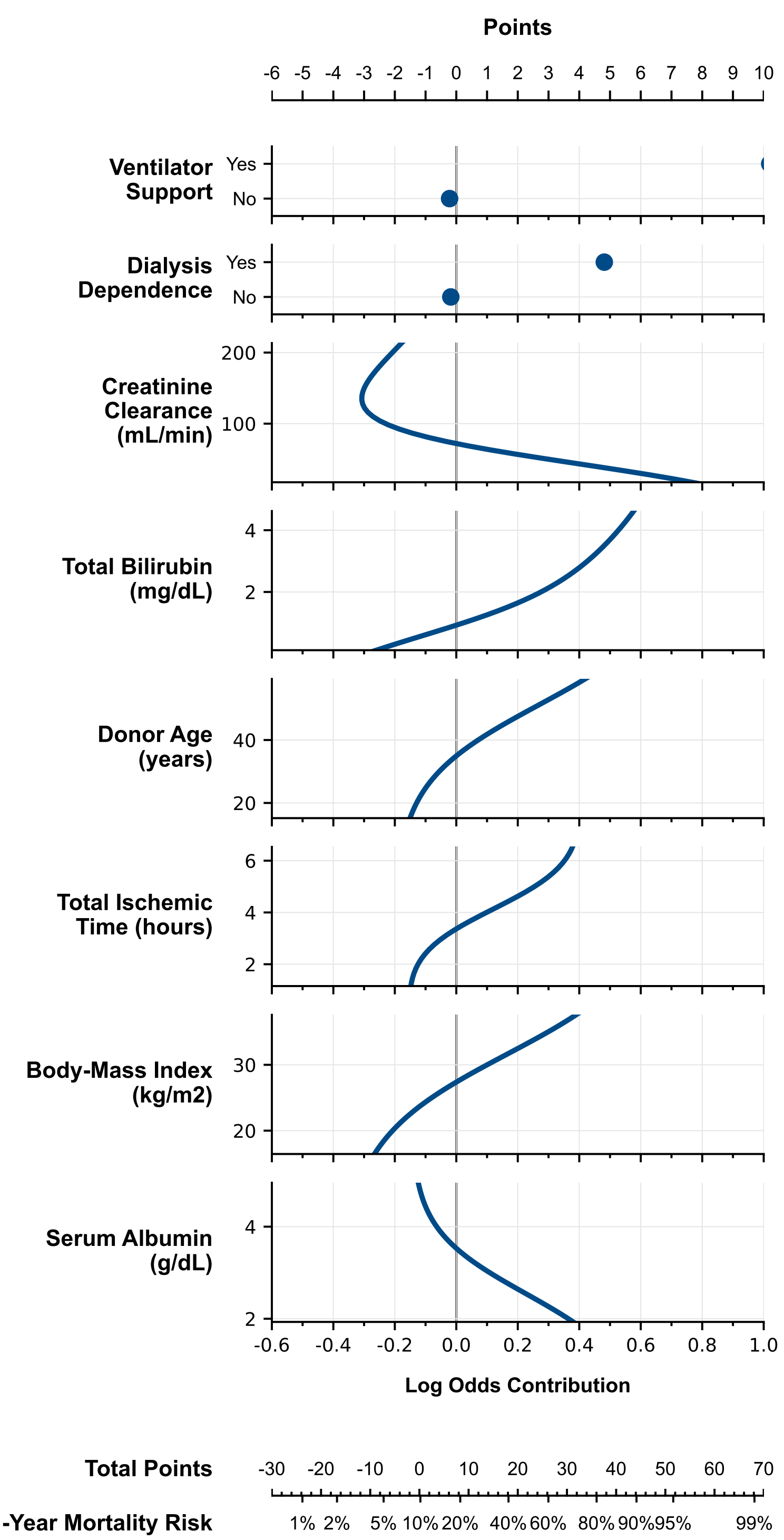

Figure S2. **D,** Random Forest (RF).

**Partial Response Nomogram – XGB**

**Points**

-6 -5 -4 -3 -2 -1 0 1 2 3 4 5 6 7 8 9 10

RVAD
Other
LVAD
BiVAD
No VAD

Valvular Heart Disease
Other Diagnosis
Ischemic Cardiomyopathy
Prior Graft Failure
Congenital Heart Disease
Nonischemic Cardiomyopathy

**Ventilator Support** Yes No

**Dialysis Dependence** Yes No

**Creatinine Clearance (mL/min)** 200 100

**Total Bilirubin (mg/dL)** 4 2

**Donor Age (years)** 40 20

**Total Ischemic Time (hours)** 6 4 2

**Body-Mass Index (kg/m2)** 30 20

**Serum Albumin (g/dL)** 4 2

**Duration on Waiting List (days)** 2000 1000

-0.6 -0.4 -0.2 0.0 0.2 0.4 0.6 0.8 1.0

**Log Odds Contribution**

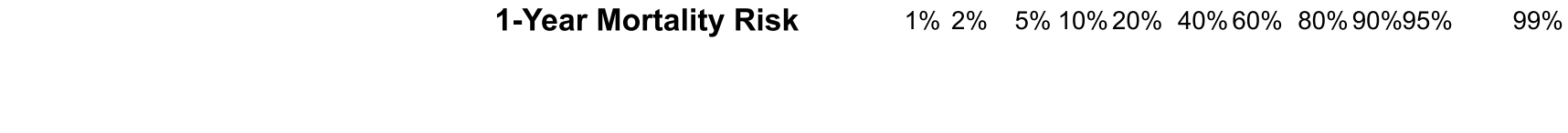

Figure S2. **E,** XGBoost (XGB).

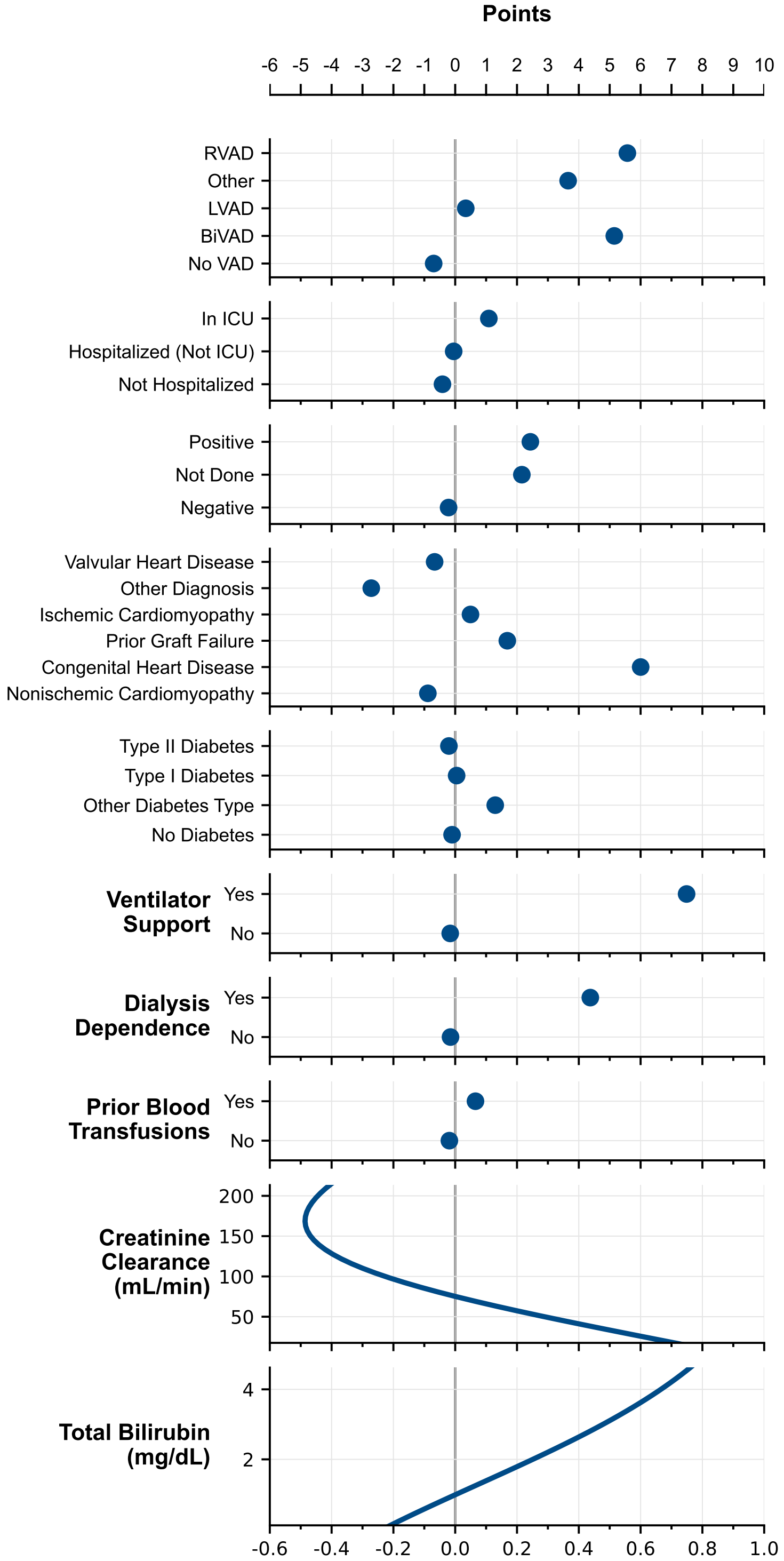

Figure S3. **Complete baseline PRiSM nomogram for the MLP model, related to Figure 4.** This nomogram includes all univariate and bivariate terms selected at the baseline regularization strength ($\geq$99.8% of maximum tuning cohort AUROC) after Phase I decomposition, illustrating the full set of main effects and interactions captured by the MLP before sparse selection. The nomogram was derived from the MLP fitted in the training cohort ($n = 32{,}543$), with $\lambda$ selected in the tuning cohort ($n = 8{,}558$); points are point estimates and carry no confidence intervals. **A,** Page 1. AUROC denotes area under the receiver operating characteristic curve; MLP, multilayer perceptron; and PRiSM, Partial Responses in Structured Models.

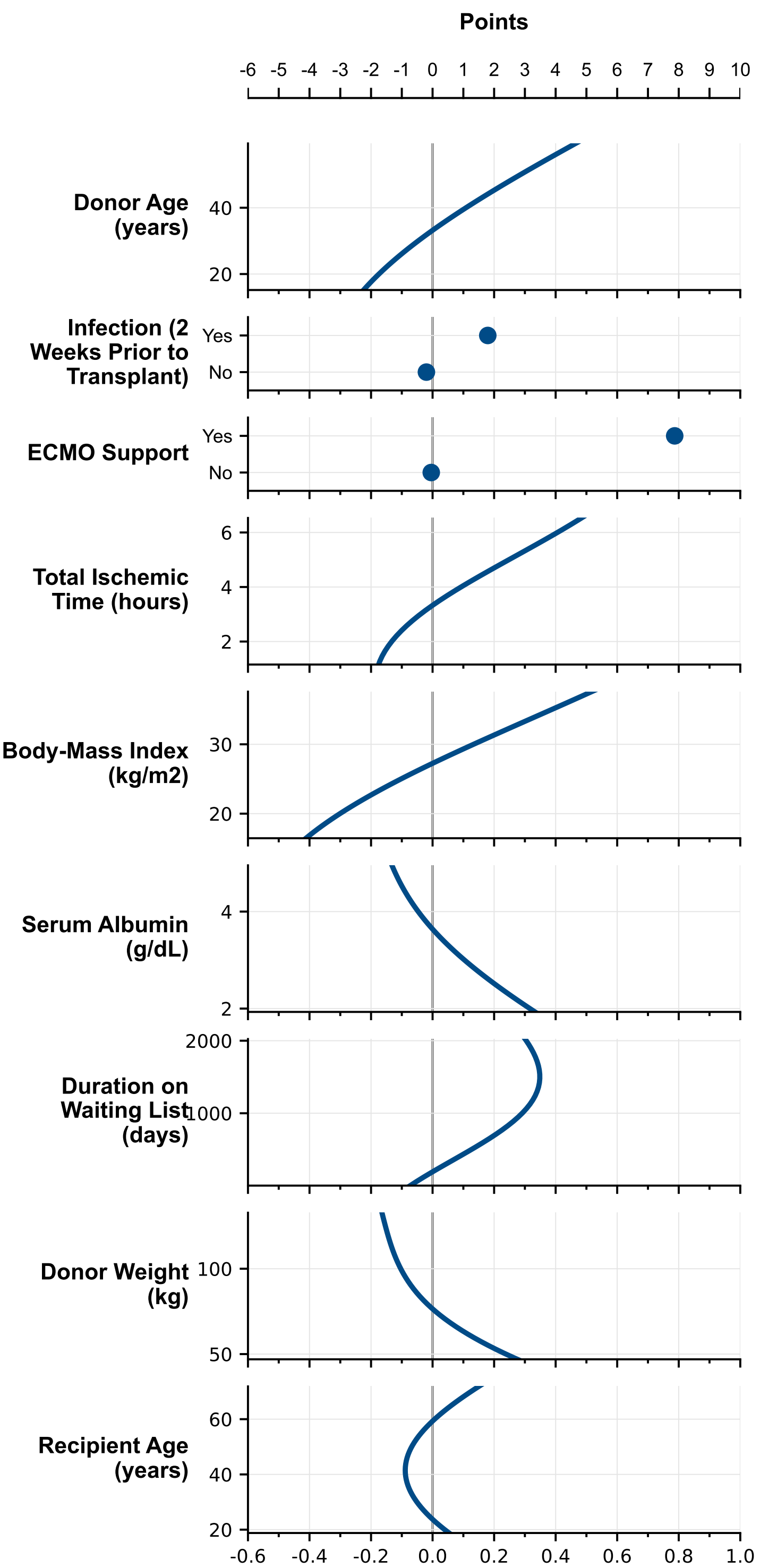

Figure S3. **B,** Page 2.

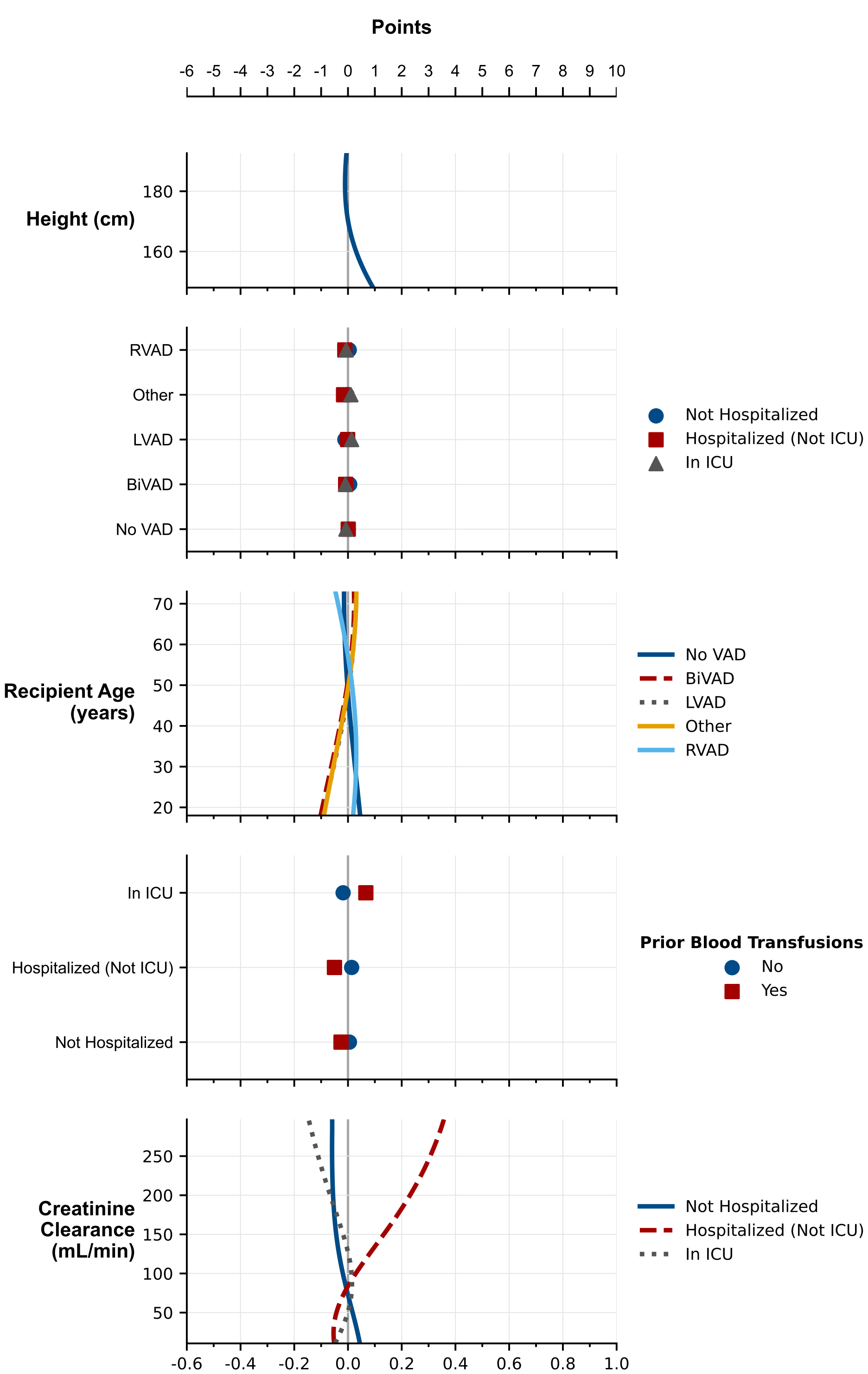

Figure S3. **C,** Page 3.

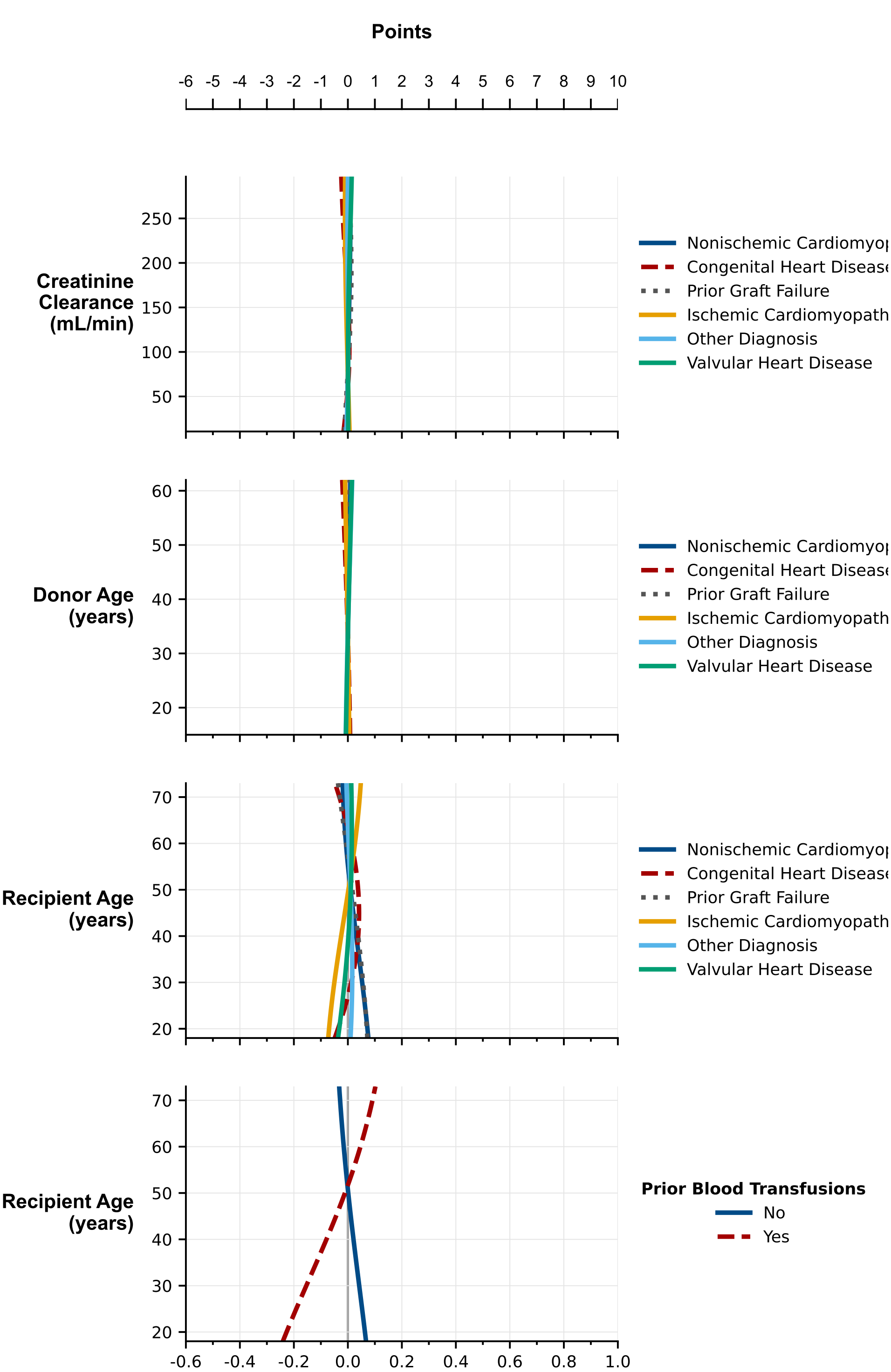

Figure S3. **D,** Page 4.

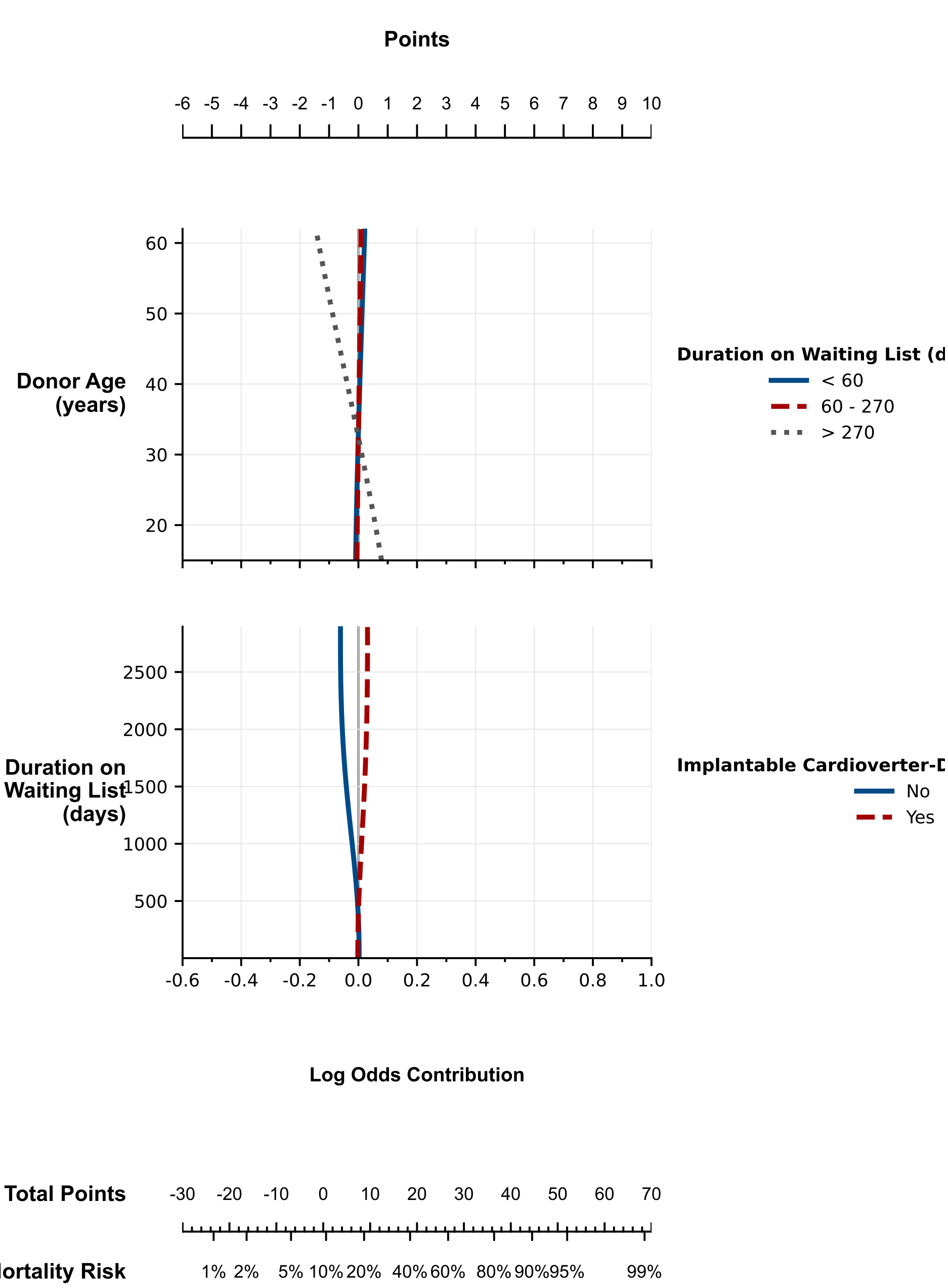

Figure S3. **E,** Page 5; bottom scale converts summed total points to predicted 1-year mortality risk.

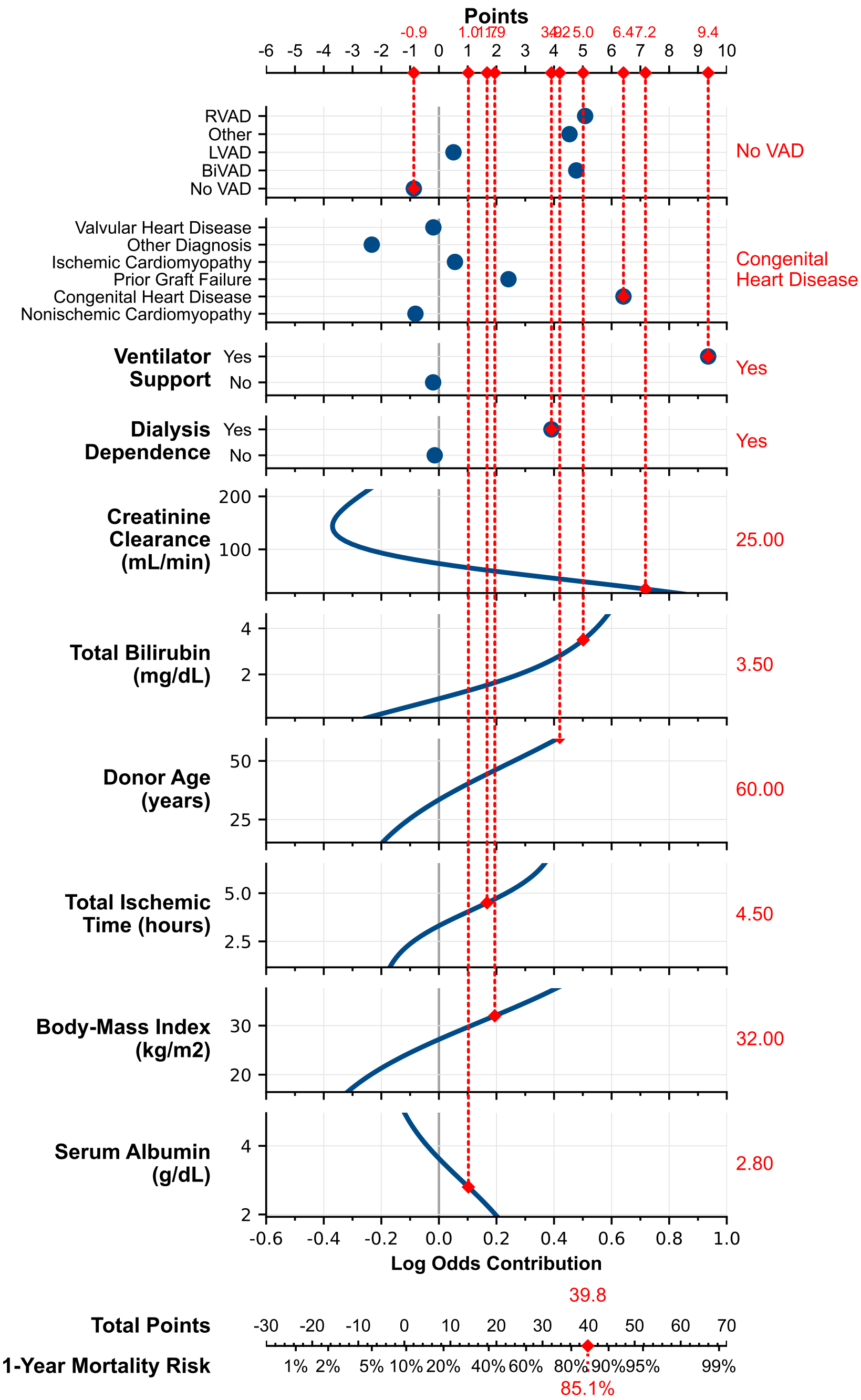

Figure S4. **Worked example of a bedside risk calculation using the sparse PRN MLP nomogram, related to Figure 4.** Red diamonds and dotted lines trace the calculation for a single synthetic high-risk patient; no cohort data are shown, and the traced values carry no confidence intervals. Points are read from the top scale at the patient's value: directly from the category markers for categorical predictors, and by projection along the response curve for continuous ones. *Patient profile:* No VAD ($-0.9$ pt); congenital heart disease ($+6.4$ pt); ventilator = yes ($+9.4$ pt); dialysis = yes ($+3.9$ pt); creatinine clearance 25 mL/min ($+7.2$ pt); total bilirubin 3.5 mg/dL ($+5.0$ pt); donor age 60 yr ($+4.2$ pt); ischemic time 4.5 hr ($+1.7$ pt); body mass index 32 kg/m$^2$ ($+1.9$ pt); serum albumin 2.8 g/dL ($+1.0$ pt). Summing gives $+39.8$ total points $\rightarrow$ 85.1% 1-year mortality (1 point $=$ 0.1 log-odds; log-odds $= 3.98 + \beta_0 = 3.98 - 2.24 = +1.74$). MLP denotes multilayer perceptron; PRN, Partial Response Network; and VAD, ventricular assist device.

# S3 Supplementary Tables

Table S1. TRIPOD+AI compliance checklist

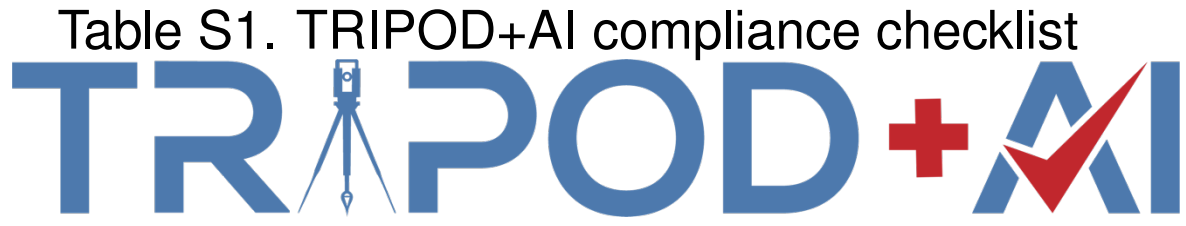

Version: 11-January-2024

Named sections are in the main text.
S-numbered items (§S1.2, Tab. S3, Fig. S1) are in Document S1.

| Section/Topic | Item | Development / evaluation[1] | Checklist item | Reported on page |
|---|---|---|---|---|
| **TITLE** | | | | |
| *Title* | 1 | D;E | Identify the study as developing or evaluating the performance of a multivariable prediction model, the target population, and the outcome to be predicted | Title |
| **ABSTRACT** | | | | |
| *Abstract* | 2 | D;E | See TRIPOD+AI for Abstracts checklist | Summary |
| **INTRODUCTION** | | | | |
| *Background* | 3a | D;E | Explain the healthcare context (including whether diagnostic or prognostic) and rationale for developing or evaluating the prediction model, including references to existing models | Intro |
| | 3b | D;E | Describe the target population and the intended purpose of the prediction model in the context of the care pathway, including its intended users (e.g., healthcare professionals, patients, public) | Intro, Disc. |
| | 3c | D;E | Describe any known health inequalities between sociodemographic groups | None |
| *Objectives* | 4 | D;E | Specify the study objectives, including whether the study describes the development or validation of a prediction model (or both) | Intro |
| **METHODS** | | | | |
| *Data* | 5a | D;E | Describe the sources of data separately for the development and evaluation datasets (e.g., randomised trial, cohort, routine care or registry data), the rationale for using these data, and representativeness of the data | Methods; §S1.1 |
| | 5b | D;E | Specify the dates of the collected participant data, including start and end of participant accrual; and, if applicable, end of follow-up | Methods; §S1.1 |
| *Participants* | 6a | D;E | Specify key elements of the study setting (e.g., primary care, secondary care, general population) including the number and location of centres | Methods; §S1.1 |
| | 6b | D;E | Describe the eligibility criteria for study participants | Tab. S2 |
| | 6c | D;E | Give details of any treatments received, and how they were handled during model development or evaluation, if relevant | Methods; Tab. 1 |
| *Data preparation* | 7 | D;E | Describe any data pre-processing and quality checking, including whether this was similar across relevant sociodemographic groups | Methods; §S1.2 |
| *Outcome* | 8a | D;E | Clearly define the outcome that is being predicted and the time horizon, including how and when assessed, the rationale for choosing this outcome, and whether the method of outcome assessment is consistent across sociodemographic groups | Methods; §S1.1 |
| | 8b | D;E | If outcome assessment requires subjective interpretation, describe the qualifications and demographic characteristics of the outcome assessors | N/A |
| | 8c | D;E | Report any actions to blind assessment of the outcome to be predicted | §S1.1 |
| *Predictors* | 9a | D | Describe the choice of initial predictors (e.g., literature, previous models, all available predictors) and any pre-selection of predictors before model building | Methods; §S1.2 |
| | 9b | D;E | Clearly define all predictors, including how and when they were measured (and any actions to blind assessment of predictors for the outcome and other predictors) | Tab. 1; §S1.2, Tab. S15 |
| | 9c | D;E | If predictor measurement requires subjective interpretation, describe the qualifications and demographic characteristics of the predictor assessors | N/A |
| *Sample size* | 10 | D;E | Explain how the study size was arrived at (separately for development and evaluation), and justify that the study size was sufficient to answer the research question. Include details of any sample size calculation | Results; Tab. S3 |
| *Missing data* | 11 | D;E | Describe how missing data were handled. Provide reasons for omitting any data | Methods; §S1.2 |
| *Analytical methods* | 12a | D | Describe how the data were used (e.g., for development and evaluation of model performance) in the analysis, including whether the data were partitioned, considering any sample size requirements | Methods; Tab. S3 |
| | 12b | D | Depending on the type of model, describe how predictors were handled in the analyses (functional form, rescaling, transformation, or any standardisation). | Methods; §S1.2 |
| | 12c | D | Specify the type of model, rationale[2], all model-building steps, including any hyperparameter tuning, and method for internal validation | Methods; §S1.3 |
| | 12d | D;E | Describe if and how any heterogeneity in estimates of model parameter values and model performance was handled and quantified across clusters (e.g., hospitals, countries). See TRIPOD-Cluster for additional considerations[3] | N/A |
| | 12e | D;E | Specify all measures and plots used (and their rationale) to evaluate model performance (e.g., discrimination, calibration, clinical utility) and, if relevant, to compare multiple models | Methods; §S1.5 |
| | 12f | E | Describe any model updating (e.g., recalibration) arising from the model evaluation, either overall or for particular sociodemographic groups or settings | Methods; §S1.9 |
| | 12g | E | For model evaluation, describe how the model predictions were calculated (e.g., formula, code, object, application programming interface) | Methods; §S1.4.4 |
| *Class imbalance* | 13 | D;E | If class imbalance methods were used, state why and how this was done, and any subsequent methods to recalibrate the model or the model predictions | N/A |
| *Fairness* | 14 | D;E | Describe any approaches that were used to address model fairness and their rationale | §S1.8.3 |
| *Model output* | 15 | D | Specify the output of the prediction model (e.g., probabilities, classification). Provide details and rationale for any classification and how the thresholds were identified | Methods; §S1.4.4 |

[1] D=items relevant only to the development of a prediction model; E=items relating solely to the evaluation of a prediction model; D;E=items applicable to both the development and evaluation of a prediction model
[2] Separately for all model building approaches.
[3] TRIPOD-Cluster is a checklist of reporting recommendations for studies developing or validating models that explicitly account for clustering or explore heterogeneity in model performance (eg, at different hospitals or centres). Debray et al, BMJ 2023; 380: e071018 [DOI: 10.1136/bmj-2022-071018]

Table S1. TRIPOD+AI compliance checklist (continued)

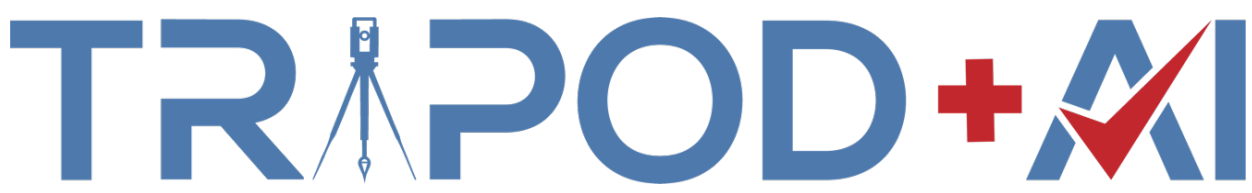

Named sections: main text. S-numbered items: Document S1.

Version: 11-January-2024

| | | | | |
|---|---|---|---|---|
| *Training versus evaluation* | 16 | D;E | Identify any differences between the development and evaluation data in healthcare setting, eligibility criteria, outcome, and predictors | Results; Tab. 1, Tab. S15 |
| *Ethical approval* | 17 | D;E | Name the institutional research board or ethics committee that approved the study and describe the participant-informed consent or the ethics committee waiver of informed consent | Methods |
| **OPEN SCIENCE** | | | | |
| *Funding* | 18a | D;E | Give the source of funding and the role of the funders for the present study | Acknowl. |
| *Conflicts of interest* | 18b | D;E | Declare any conflicts of interest and financial disclosures for all authors | Decl. int. |
| *Protocol* | 18c | D;E | Indicate where the study protocol can be accessed or state that a protocol was not prepared | None |
| *Registration* | 18d | D;E | Provide registration information for the study, including register name and registration number, or state that the study was not registered | §S1.1 |
| *Data sharing* | 18e | D;E | Provide details of the availability of the study data | Res. avail. |
| *Code sharing* | 18f | D;E | Provide details of the availability of the analytical code[4] | Res. avail. |
| **PATIENT & PUBLIC INVOLVEMENT** | | | | |
| *Patient & Public Involvement* | 19 | D;E | Provide details of any patient and public involvement during the design, conduct, reporting, interpretation, or dissemination of the study or state no involvement. | None |
| **RESULTS** | | | | |
| *Participants* | 20a | D;E | Describe the flow of participants through the study, including the number of participants with and without the outcome and, if applicable, a summary of the follow-up time. A diagram may be helpful. | Results; Tab. S2 |
| | 20b | D;E | Report the characteristics overall and, where applicable, for each data source or setting, including the key dates, key predictors (including demographics), treatments received, sample size, number of outcome events, follow-up time, and amount of missing data. A table may be helpful. Report any differences across key demographic groups. | Results; Tab. 1, Tab. S15 |
| | 20c | E | For model evaluation, show a comparison with the development data of the distribution of important predictors (demographics, predictors, and outcome). | Tab. 1, Tab. S15 |
| *Model development* | 21 | D;E | Specify the number of participants and outcome events in each analysis (e.g., for model development, hyperparameter tuning, model evaluation) | Results; Tab. S3 |
| *Model specification* | 22 | D | Provide details of the full prediction model (e.g., formula, code, object, application programming interface) to allow predictions in new individuals and to enable third-party evaluation and implementation, including any restrictions to access or re-use (e.g., freely available, proprietary)[5] | Res. avail.; Fig. S1-S3 |
| *Model performance* | 23a | D;E | Report model performance estimates with confidence intervals, including for any key subgroups (e.g., sociodemographic). Consider plots to aid presentation. | Results, Tab. 2; Tab. S26 |
| | 23b | D;E | If examined, report results of any heterogeneity in model performance across clusters. See TRIPOD Cluster for additional details[3]. | N/A |
| *Model updating* | 24 | E | Report the results from any model updating, including the updated model and subsequent performance | N/A |
| **DISCUSSION** | | | | |
| *Interpretation* | 25 | D;E | Give an overall interpretation of the main results, including issues of fairness in the context of the objectives and previous studies | Disc. |
| *Limitations* | 26 | D;E | Discuss any limitations of the study (such as a non-representative sample, sample size, overfitting, missing data) and their effects on any biases, statistical uncertainty, and generalizability | Disc. |
| *Usability of the model in the context of current care* | 27a | D | Describe how poor quality or unavailable input data (e.g., predictor values) should be assessed and handled when implementing the prediction model | None |
| | 27b | D | Specify whether users will be required to interact in the handling of the input data or use of the model, and what level of expertise is required of users | Disc. |
| | 27c | D;E | Discuss any next steps for future research, with a specific view to applicability and generalizability of the model | Disc. |

From: Collins GS, Moons KGM, Dhiman P, et al. *BMJ* 2024;385:e078378. doi:10.1136/bmj-2023-078378

[4] This relates to the analysis code, for example, any data cleaning, feature engineering, model building, evaluation.
[5] This relates to the code to implement the model to get estimates of risk for a new individual.

Table S2. Sequential exclusion criteria, related to Table 1

| Criterion | Patients Excluded | Patients Remaining |
|---|---|---|
| Source dataset (SRTR thoracic) | – | 224,983 |
| Non-heart transplant types | 129,115 | 95,868 |
| Recipient age $<$ 18 years | 12,697 | 83,171 |
| Donor age $<$ 15 years | 2,466 | 80,705 |
| Transplant year outside 2000–2022 | 29,164 | 51,541 |
| Non-standard removal reason codes | 104 | 51,437 |
| Censored before 365 days | 1,081 | 50,356 |
| **Final cohort** | – | **50,356** |

SRTR denotes Scientific Registry of Transplant Recipients.

Table S3. Standard training configuration

| Parameter | Value |
|---|---|
| Outcome ($y$) | 1-year mortality |
| Training cohort | 32,543 patients (2000–2016) |
| Tuning cohort | 8,558 patients (2017–2019) |
| External validation cohort | 9,255 patients (2020–2022) |
| Cohort assignment | Predefined by transplant year; no within-year randomization |
| Missing data handling | MissForest imputation model trained on training cohort, applied unchanged to tuning and validation cohorts |
| Continuous variable scaling | Median-centered, scaled by $2 \times \mathrm{SD}$ (PRiSMScaler with `sd_scale=2.0`) |
| Binary variable handling | Excluded from scaling (auto-detected; remain 0/1) |
| Categorical encoding | Dummy encoded (most-common category as reference, dropped) |
| Hyperparameter optimization | Optuna (TPE sampler, 64–256 trials per model, 8 parallel jobs) |
| Optimization metric | Tuning cohort AUROC |
| Random seed | 257 |

AUROC denotes area under the receiver operating characteristic curve; SD, standard deviation; and TPE, Tree-structured Parzen Estimator.

Table S4. Logistic regression hyperparameters

| Parameter | Tuning Range | Tuned Value |
|---|---|---|
| Learning rate | $3 \times 10^{-3}$ to $3 \times 10^{-2}$ (log) | 0.0200 |
| Weight decay (L2) | $10^{-5}$ to $2 \times 10^{-2}$ (log) | $1.00 \times 10^{-3}$ |
| Early stopping patience | 3 to 15 | 12 |
| Maximum epochs | – | 2000 |
| Early stopping tolerance | – | $10^{-4}$ |

Implementation: PyTorch with BCEWithLogitsLoss and Adam optimizer. Early stopping on tuning cohort loss. L2 denotes L2 (ridge) weight-decay regularization.

Table S5. MLP hyperparameters

| Parameter | Tuning Range | Tuned Value |
|---|---|---|
| Hidden units | 20 to 40 | 37 |
| Learning rate | $3 \times 10^{-4}$ to $3 \times 10^{-3}$ (log) | $1.15 \times 10^{-3}$ |
| Weight decay (L2) | $10^{-5}$ to $10^{-4}$ (log) | $7.47 \times 10^{-5}$ |
| Early stopping patience | 8 to 20 | 18 |
| Batch size | {512, 1024, 2048} | 512 |
| Maximum epochs | – | 4000 |
| Early stopping tolerance | – | $10^{-4}$ |

Implementation: PyTorch single hidden layer MLP with tanh activation and sigmoid output. Adam optimizer with BCELoss. Xavier uniform weight initialization. MLP denotes multilayer perceptron; and L2, L2 (ridge) weight-decay regularization.

## Table S6. Random forest hyperparameters

| Parameter | Tuning Range | Tuned Value |
|---|---|---|
| Number of estimators | 100 to 300 | 242 |
| Maximum depth | 4 to 8 | 8 |
| Minimum child weight | 8 to 20 | 8 |
| Row subsampling | 0.7 to 0.9 | 0.704 |
| Per-node column sampling | 0.6 to 0.8 | 0.603 |
| Learning rate | – | 1.0 |

Implementation: XGBRFClassifier (XGBoost library bagging mode with per-node column sampling, analogous to sklearn RandomForestClassifier). Objective: binary:logistic. Evaluation metric: logloss. The XGBoost shrinkage parameter (learning rate) was fixed at 1.0 to disable boosting, yielding pure bagging. RF denotes random forest.

## Table S7. XGBoost hyperparameters

| Parameter | Tuning Range | Tuned Value |
|---|---|---|
| Maximum depth | 2 to 6 | 3 |
| Learning rate | $2 \times 10^{-2}$ to $10^{-1}$ (log) | 0.0356 |
| Row subsampling | 0.7 to 0.9 | 0.743 |
| Per-tree column sampling | 0.6 to 0.8 | 0.670 |
| Minimum child weight | 5 to 15 | 10 |
| Number of estimators | 100 to 300 | 288 |
| Gamma ($\gamma$) | 0 to 3.0 | 2.775 |
| L1 regularization ($\alpha$) | $10^{-6}$ to $5.0$ (log) | $9.17 \times 10^{-4}$ |
| L2 regularization ($\lambda$) | $10^{-4}$ to $10.0$ (log) | 0.0403 |
| Early stopping rounds | – | 10 |

Implementation: XGBClassifier (XGBoost library). Objective: binary:logistic. Evaluation metric: logloss. L1 denotes L1 (lasso) weight regularization; L2, L2 (ridge) weight-decay regularization; and XGB, XGBoost.

## Table S8. IMPACT model configuration

| Parameter | Value |
|---|---|
| Model type | Pretrained scoring model (logistic regression with fixed coefficients) |
| Implementation | PyTorch |
| Calibration | Original published coefficients; no recalibration performed |
| VAD simplification | All VAD types assigned 3 points (original paper distinguishes pulsatile, continuous, and HeartMate II) |

The IMPACT model (Weiss et al. 2011) is a pretrained clinical scoring model with published coefficients. No tuning or training was performed; the model was applied directly using its original variable thresholds and logit coefficients. The original IMPACT score assigns different point values to pulsatile, continuous-flow, and HeartMate II ventricular assist devices; these subtypes are not recorded as distinct categories in the current SRTR database, so a single pooled VAD assignment of 3 points was used for all device types. IMPACT denotes Index for Mortality Prediction After Cardiac Transplantation; and VAD, ventricular assist device.

## Table S9. PRiSM configuration, related to Figure 1

| Parameter | Value |
|---|---|
| Decomposition method | Lebesgue (averages over empirical training distribution) |
| Training data used | Training cohort (*n*=32,543), mortality outcome labels |
| LASSO implementation | $L_1$-penalized logistic regression (sklearn LogisticRegression, solver=`'saga'`, warm start) |
| LASSO $\lambda$ grid (black-box) | 150 values, logspace from 50 to 0.001 |
| LASSO $\lambda$ grid (PRN) | 75 values, logspace from 50 to 0.001 |
| LASSO maximum iterations | 10,000 per $\lambda$ |
| LASSO convergence tolerance | $10^{-4}$ |
| Feature selection threshold | 0.1 (features with $\lvert\beta\rvert > 0.1$ are selected) |
| Baseline $\lambda$ selection | `max_test_auc` with `target_ratio=0.998` (first $\lambda$ achieving $\geq$99.8% of max tuning AUROC) |
| Random seed | 257 |

AUROC denotes area under the receiver operating characteristic curve; LASSO, least absolute shrinkage and selection operator; PRiSM, Partial Responses in Structured Models; and PRN, Partial Response Network.

Table S10. PRN refinement configuration, related to Figure 1

| Parameter | Value | Tuning Range |
|---|---|---|
| Training data used | Training cohort (*n*=32,543), mortality outcome labels | – |
| Architecture | Single hidden layer (tanh activation, sigmoid output) | – |
| Hidden nodes per selected feature | 5 | – |
| Weight initialization | Xavier uniform (tanh/sigmoid gains) | – |
| Optimizer | Adam | – |
| Loss function | BCELoss | – |
| Learning rate | – | $3 \times 10^{-4}$ to $3 \times 10^{-3}$ (log) |
| Weight decay (L2) | – | $10^{-5}$ to $10^{-4}$ (log) |
| Early stopping patience | – | 30 to 100 |
| Early stopping tolerance | 0.001 | – |
| Maximum epochs | 4000 | – |
| Batch size | – | {128, 256, 512} |
| Mask enforcement | Applied after each optimizer step | – |
| Tuning trials | 128 (Optuna TPE) | – |

The Partial Response Network (PRN) is a structured MLP constrained to LASSO-selected effects via an input mask. Hyperparameters were tuned separately for each source model via Optuna (128 trials, 8 parallel jobs, TPE sampler, maximizing tuning cohort AUROC). AUROC denotes area under the receiver operating characteristic curve; L2, L2 (ridge) weight-decay regularization; LASSO, least absolute shrinkage and selection operator; MLP, multilayer perceptron; PRN, Partial Response Network; and TPE, Tree-structured Parzen Estimator.

Table S11. PRN tuned values by source model, sparse lambda, related to Figure 1

| Source Model | Learning Rate | Weight Decay | Patience | Batch Size |
|---|---|---|---|---|
| IMPACT | $8.59 \times 10^{-4}$ | $9.14 \times 10^{-5}$ | 62 | 512 |
| LR | $1.82 \times 10^{-3}$ | $3.23 \times 10^{-5}$ | 50 | 256 |
| MLP | $2.29 \times 10^{-3}$ | $5.10 \times 10^{-5}$ | 93 | 256 |
| RF | $2.43 \times 10^{-3}$ | $1.27 \times 10^{-5}$ | 68 | 128 |
| XGB | $2.34 \times 10^{-3}$ | $5.63 \times 10^{-5}$ | 60 | 128 |

IMPACT denotes Index for Mortality Prediction After Cardiac Transplantation; LR, logistic regression; MLP, multilayer perceptron; PRN, Partial Response Network; RF, random forest; and XGB, XGBoost.

Table S12. Discrimination results: training cohort, *n*=32,543, related to Table 2

| Model | Version | No. of Variables or Terms | AUROC (95% CI) | Difference in AUROC (95% CI) | Retention Contrast *T* (95% CI) |
|---|---|---|---|---|---|
| IMPACT | Source | 12[v] | 0.620 (0.610, 0.631) | — | — |
| | PRiSM | 9 | 0.621 (0.611, 0.631) | +0.001 (−0.006, +0.006) | +0.013 (+0.007, +0.018) |
| | Sparse PRiSM | 5 | 0.604 (0.594, 0.614) | −0.017 (−0.025, −0.008) | −0.004 (−0.013, +0.004) |
| | PRN | 9 | 0.635 (0.625, 0.646) | +0.015 (+0.008, +0.022) | +0.027 (+0.020, +0.034) |
| | Sparse PRN | 5 | 0.623 (0.612, 0.633) | +0.002 (−0.007, +0.011) | +0.014 (+0.005, +0.023) |
| LR | Source | 22[v] | 0.657 (0.647, 0.667) | — | — |
| | PRiSM | 21 | 0.661 (0.651, 0.670) | +0.004 (+0.001, +0.006) | +0.019 (+0.017, +0.022) |
| | Sparse PRiSM | 9 | 0.644 (0.634, 0.654) | −0.013 (−0.018, −0.007) | +0.003 (−0.002, +0.008) |
| | PRN | 19 | 0.669 (0.659, 0.679) | +0.012 (+0.008, +0.017) | +0.028 (+0.023, +0.032) |
| | Sparse PRN | 9 | 0.653 (0.643, 0.663) | −0.004 (−0.010, +0.002) | +0.011 (+0.006, +0.017) |
| MLP | Source | 22[v] | 0.688 (0.678, 0.697) | — | — |
| | PRiSM | 30 | 0.673 (0.664, 0.683) | −0.014 (−0.017, −0.012) | +0.004 (+0.002, +0.007) |
| | Sparse PRiSM | 10 | 0.656 (0.646, 0.666) | −0.032 (−0.037, −0.026) | −0.013 (−0.019, −0.008) |
| | PRN | 22 | 0.671 (0.661, 0.681) | −0.017 (−0.020, −0.013) | +0.002 (−0.002, +0.005) |
| | Sparse PRN | 10 | 0.661 (0.651, 0.671) | −0.027 (−0.032, −0.022) | −0.008 (−0.013, −0.003) |
| RF | Source | 22[v] | 0.713 (0.704, 0.723) | — | — |
| | PRiSM | 22 | 0.671 (0.661, 0.681) | −0.042 (−0.045, −0.039) | −0.021 (−0.024, −0.017) |
| | Sparse PRiSM | 8 | 0.645 (0.635, 0.655) | −0.069 (−0.073, −0.064) | −0.047 (−0.052, −0.042) |
| | PRN | 15 | 0.667 (0.657, 0.676) | −0.047 (−0.051, −0.042) | −0.025 (−0.030, −0.021) |
| | Sparse PRN | 8 | 0.648 (0.638, 0.658) | −0.066 (−0.071, −0.060) | −0.044 (−0.049, −0.039) |
| XGB | Source | 22[v] | 0.710 (0.700, 0.719) | — | — |
| | PRiSM | 26 | 0.684 (0.675, 0.694) | −0.025 (−0.028, −0.023) | −0.004 (−0.007, −0.002) |
| | Sparse PRiSM | 11 | 0.668 (0.658, 0.678) | −0.041 (−0.046, −0.037) | −0.021 (−0.025, −0.016) |
| | PRN | 17 | 0.669 (0.659, 0.679) | −0.041 (−0.044, −0.037) | −0.020 (−0.023, −0.016) |
| | Sparse PRN | 11 | 0.663 (0.653, 0.672) | −0.047 (−0.052, −0.043) | −0.026 (−0.031, −0.022) |

Difference in AUROC = Nomogram − Source, a descriptive quantity. The retention contrast is $T = \mathrm{AUROC}_{\mathrm{nomogram}} - 0.90 \times \mathrm{AUROC}_{\mathrm{source}} - 0.05$, which exceeds zero exactly when the nomogram retains more than 90% of its source model's discrimination above chance. It is the primary endpoint, and is the quantity main text Table 2 reports. $T$ is shown here for completeness. The noninferiority criterion was prespecified for the external-validation cohort only, and no verdict is marked in this table. 95% CIs for AUROC by the DeLong method, and for the difference in AUROC and for $T$ from the same paired DeLong covariance matrix of the two AUROCs. [v] No. of variables or terms denotes input variables for source models and nomogram terms for all other rows. PRiSM = Phase I only (ANOVA decomposition with LASSO selection); PRN = Phase I + Phase II (PRN refinement). Baseline: lambda preserving $\geq$99.8% of maximum tuning AUROC. Sparse: the sparse lambda, minimizing $n$-terms while targeting noninferiority on tuning AUROC; Sparse PRN applies PRN refinement to the sparse selection. AUROC denotes area under the receiver operating characteristic curve; IMPACT, Index for Mortality Prediction After Cardiac Transplantation; LR, logistic regression; MLP, multilayer perceptron; PRiSM, Partial Responses in Structured Models; PRN, Partial Response Network; RF, random forest; and XGB, XGBoost.

Table S13. Discrimination results: tuning cohort, *n*=8,558, related to Table 2

| Model | Version | No. of Variables or Terms | AUROC (95% CI) | Difference in AUROC (95% CI) | Retention Contrast *T* (95% CI) |
|---|---|---|---|---|---|
| IMPACT | Source | 12[v] | 0.593 (0.571, 0.614) | — | — |
| | PRiSM | 9 | 0.603 (0.582, 0.625) | +0.011 (−0.004, +0.025) | +0.020 (+0.006, +0.034) |
| | Sparse PRiSM | 5 | 0.585 (0.563, 0.607) | −0.008 (−0.027, +0.011) | +0.001 (−0.017, +0.020) |
| | PRN | 9 | 0.617 (0.596, 0.639) | +0.025 (+0.008, +0.041) | +0.034 (+0.018, +0.050) |
| | Sparse PRN | 5 | 0.612 (0.591, 0.633) | +0.019 (−0.001, +0.040) | +0.029 (+0.009, +0.048) |
| LR | Source | 22[v] | 0.643 (0.622, 0.664) | — | — |
| | PRiSM | 21 | 0.641 (0.620, 0.662) | −0.002 (−0.008, +0.003) | +0.012 (+0.006, +0.018) |
| | Sparse PRiSM | 9 | 0.630 (0.608, 0.651) | −0.013 (−0.025, −0.001) | +0.001 (−0.011, +0.013) |
| | PRN | 19 | 0.656 (0.636, 0.677) | +0.013 (+0.003, +0.024) | +0.028 (+0.018, +0.038) |
| | Sparse PRN | 9 | 0.645 (0.624, 0.665) | +0.002 (−0.011, +0.015) | +0.016 (+0.003, +0.029) |
| MLP | Source | 22[v] | 0.659 (0.638, 0.679) | — | — |
| | PRiSM | 30 | 0.660 (0.640, 0.680) | +0.001 (−0.005, +0.007) | +0.017 (+0.011, +0.023) |
| | Sparse PRiSM | 10 | 0.644 (0.623, 0.664) | −0.015 (−0.029, −0.002) | +0.001 (−0.012, +0.014) |
| | PRN | 22 | 0.658 (0.638, 0.678) | −0.001 (−0.009, +0.007) | +0.015 (+0.007, +0.023) |
| | Sparse PRN | 10 | 0.652 (0.632, 0.673) | −0.006 (−0.018, +0.005) | +0.009 (−0.002, +0.021) |
| RF | Source | 22[v] | 0.644 (0.624, 0.665) | — | — |
| | PRiSM | 22 | 0.643 (0.623, 0.664) | −0.001 (−0.008, +0.007) | +0.014 (+0.006, +0.021) |
| | Sparse PRiSM | 8 | 0.630 (0.609, 0.651) | −0.014 (−0.025, −0.003) | 0.000 (−0.010, +0.011) |
| | PRN | 15 | 0.656 (0.636, 0.677) | +0.012 (+0.002, +0.023) | +0.027 (+0.016, +0.037) |
| | Sparse PRN | 8 | 0.644 (0.624, 0.665) | 0.000 (−0.012, +0.013) | +0.015 (+0.003, +0.027) |
| XGB | Source | 22[v] | 0.662 (0.641, 0.682) | — | — |
| | PRiSM | 26 | 0.653 (0.633, 0.674) | −0.008 (−0.013, −0.004) | +0.008 (+0.003, +0.013) |
| | Sparse PRiSM | 11 | 0.646 (0.625, 0.666) | −0.016 (−0.025, −0.007) | 0.000 (−0.009, +0.009) |
| | PRN | 17 | 0.656 (0.636, 0.676) | −0.005 (−0.013, +0.002) | +0.011 (+0.003, +0.018) |
| | Sparse PRN | 11 | 0.654 (0.634, 0.674) | −0.008 (−0.018, +0.003) | +0.009 (−0.001, +0.018) |

Difference in AUROC = Nomogram − Source, a descriptive quantity. The retention contrast is $T = \mathrm{AUROC}_{\mathrm{nomogram}} - 0.90 \times \mathrm{AUROC}_{\mathrm{source}} - 0.05$, which exceeds zero exactly when the nomogram retains more than 90% of its source model's discrimination above chance. It is the primary endpoint, and is the quantity main text Table 2 reports. $T$ is shown here for completeness. The noninferiority criterion was prespecified for the external-validation cohort only, and no verdict is marked in this table. 95% CIs for AUROC by the DeLong method, and for the difference in AUROC and for $T$ from the same paired DeLong covariance matrix of the two AUROCs. [v] No. of variables or terms denotes input variables for source models and nomogram terms for all other rows. PRiSM = Phase I only (ANOVA decomposition with LASSO selection); PRN = Phase I + Phase II (PRN refinement). Baseline: lambda preserving $\geq$99.8% of maximum tuning AUROC. Sparse: the sparse lambda, minimizing $n$-terms while targeting noninferiority on tuning AUROC; Sparse PRN applies PRN refinement to the sparse selection. AUROC denotes area under the receiver operating characteristic curve; IMPACT, Index for Mortality Prediction After Cardiac Transplantation; LR, logistic regression; MLP, multilayer perceptron; PRiSM, Partial Responses in Structured Models; PRN, Partial Response Network; RF, random forest; and XGB, XGBoost. Note: Lambda selection for sparse nomograms was optimized on tuning cohort AUROC; tuning results therefore represent in-sample performance for the lambda selection step.

Table S14. Discrimination results: external validation cohort, *n*=9,255, related to Table 2

| Model | Version | No. of Variables or Terms | AUROC (95% CI) | Difference in AUROC (95% CI) | Retention Contrast *T* (95% CI) |
|---|---|---|---|---|---|
| IMPACT | Source | 12[v] | 0.619 (0.599, 0.639) | — | — |
| | PRiSM | 9 | 0.630 (0.610, 0.649) | +0.011 (−0.001, +0.023) | +0.023[†] (+0.011, +0.034) |
| | Sparse PRiSM | 5 | 0.601 (0.581, 0.621) | −0.018 (−0.034, −0.001) | −0.006 (−0.021, +0.010) |
| | PRN | 9 | 0.635 (0.616, 0.655) | +0.017 (+0.003, +0.030) | +0.028[†] (+0.015, +0.042) |
| | Sparse PRN | 5 | 0.620 (0.600, 0.639) | +0.001 (−0.017, +0.018) | +0.013 (−0.004, +0.029) |
| LR | Source | 22[v] | 0.620 (0.600, 0.640) | — | — |
| | PRiSM | 21 | 0.625 (0.605, 0.645) | +0.005 (0.000, +0.010) | +0.017[†] (+0.012, +0.022) |
| | Sparse PRiSM | 9 | 0.611 (0.591, 0.631) | −0.008 (−0.020, +0.003) | +0.004 (−0.007, +0.015) |
| | PRN | 19 | 0.645 (0.625, 0.665) | +0.025 (+0.016, +0.034) | +0.037[†] (+0.028, +0.046) |
| | Sparse PRN | 9 | 0.625 (0.605, 0.645) | +0.005 (−0.008, +0.018) | +0.017[†] (+0.005, +0.029) |
| MLP | Source | 22[v] | 0.645 (0.625, 0.665) | — | — |
| | PRiSM | 30 | 0.646 (0.626, 0.666) | +0.001 (−0.005, +0.007) | +0.016[†] (+0.010, +0.021) |
| | Sparse PRiSM | 10 | 0.636 (0.616, 0.655) | −0.009 (−0.022, +0.004) | +0.005 (−0.007, +0.017) |
| | PRN | 22 | 0.646 (0.626, 0.665) | +0.001 (−0.007, +0.008) | +0.015[†] (+0.008, +0.022) |
| | Sparse PRN | 10 | 0.641 (0.621, 0.661) | −0.004 (−0.015, +0.007) | +0.011 (0.000, +0.021) |
| RF | Source | 22[v] | 0.642 (0.622, 0.661) | — | — |
| | PRiSM | 22 | 0.641 (0.621, 0.660) | −0.001 (−0.008, +0.006) | +0.013[†] (+0.006, +0.020) |
| | Sparse PRiSM | 8 | 0.621 (0.601, 0.641) | −0.020 (−0.030, −0.010) | −0.006 (−0.016, +0.004) |
| | PRN | 15 | 0.649 (0.629, 0.668) | +0.007 (−0.003, +0.017) | +0.021[†] (+0.012, +0.030) |
| | Sparse PRN | 8 | 0.624 (0.604, 0.644) | −0.018 (−0.030, −0.006) | −0.004 (−0.015, +0.008) |
| XGB | Source | 22[v] | 0.646 (0.627, 0.666) | — | — |
| | PRiSM | 26 | 0.644 (0.624, 0.663) | −0.003 (−0.007, +0.002) | +0.012[†] (+0.008, +0.017) |
| | Sparse PRiSM | 11 | 0.635 (0.615, 0.654) | −0.012 (−0.020, −0.003) | +0.003 (−0.005, +0.012) |
| | PRN | 17 | 0.645 (0.626, 0.665) | −0.001 (−0.008, +0.006) | +0.013[†] (+0.006, +0.021) |
| | Sparse PRN | 11 | 0.646 (0.627, 0.666) | 0.000 (−0.010, +0.009) | +0.014[†] (+0.005, +0.024) |

Difference in AUROC = Nomogram − Source, a descriptive quantity. The retention contrast is $T = \mathrm{AUROC}_{\mathrm{nomogram}} - 0.90 \times \mathrm{AUROC}_{\mathrm{source}} - 0.05$, which exceeds zero exactly when the nomogram retains more than 90% of its source model's discrimination above chance. It is the primary endpoint, and is the quantity main text Table 2 reports. [†]Noninferiority met (lower bound of the 95% CI for $T$ exceeding 0). 95% CIs for AUROC by the DeLong method, and for the difference in AUROC and for $T$ from the same paired DeLong covariance matrix of the two AUROCs. [v] No. of variables or terms denotes input variables for source models and nomogram terms for all other rows. PRiSM = Phase I only (ANOVA decomposition with LASSO selection); PRN = Phase I + Phase II (PRN refinement). Baseline: lambda preserving $\geq$99.8% of maximum tuning AUROC. Sparse: the sparse lambda, minimizing $n$-terms while targeting noninferiority on tuning AUROC; Sparse PRN applies PRN refinement to the sparse selection. AUROC denotes area under the receiver operating characteristic curve; IMPACT, Index for Mortality Prediction After Cardiac Transplantation; LR, logistic regression; MLP, multilayer perceptron; PRiSM, Partial Responses in Structured Models; PRN, Partial Response Network; RF, random forest; and XGB, XGBoost.

Table S15. Baseline characteristics of the study population, related to Table 1

| Characteristic | Training Cohort (*n*=32,543) | Tuning Cohort (*n*=8,558) | Temporal Validation Cohort (*n*=9,255) |
|---|---|---|---|
| **Recipient Demographics** | | | |
| Recipient Age[a,b] — yr | 55.0 [46.0, 62.0] | 57.0 [46.2, 63.0] | 57.0 [46.0, 63.0] |
| Female sex[b] | 7,913 (24.3) | 2,310 (27.0) | 2,435 (26.3) |
| Race or ethnic group[b] | | | |
| White | 22,760 (69.9) | 5,412 (63.2) | 5,414 (58.5) |
| Black | 6,073 (18.7) | 1,980 (23.1) | 2,409 (26.0) |
| Hispanic/Latino | 2,507 (7.7) | 772 (9.0) | 968 (10.5) |
| Asian | 888 (2.7) | 304 (3.6) | 349 (3.8) |
| Native American | 103 (0.3) | 23 (0.3) | 32 (0.3) |
| Pacific Islander | 103 (0.3) | 32 (0.4) | 35 (0.4) |
| Multi-Racial | 109 (0.3) | 35 (0.4) | 48 (0.5) |
| Body-Mass Index[a] — kg/m$^2$ | 26.6 [23.4, 30.0] | 27.4 [24.0, 31.1] | 27.3 [24.1, 31.1] |
| *Missing* | *579 (1.8)* | *160 (1.9)* | *218 (2.4)* |
| Weight — kg | 80.7 [69.4, 92.5] | 82.6 [70.8, 95.7] | 82.8 [71.2, 95.3] |
| *Missing* | *552 (1.7)* | *186 (2.2)* | *261 (2.8)* |
| Height[a] — cm | 175.3 [167.6, 180.3] | 175.0 [167.6, 180.3] | 175.0 [167.6, 180.3] |
| *Missing* | *335 (1.0)* | *106 (1.2)* | *126 (1.4)* |
| **Recipient Clinical Characteristics** | | | |
| Primary cause of heart failure[a,b] | | | |
| Nonischemic cardiomyopathy | 16,282 (50.0) | 5,201 (60.8) | 5,786 (62.5) |
| Ischemic cardiomyopathy | 13,125 (40.3) | 2,501 (29.2) | 2,510 (27.1) |
| Congenital | 836 (2.6) | 290 (3.4) | 356 (3.8) |
| Graft Failure | 978 (3.0) | 234 (2.7) | 254 (2.7) |
| Valvular Heart Disease | 608 (1.9) | 98 (1.1) | 88 (1.0) |
| Other | 714 (2.2) | 234 (2.7) | 261 (2.8) |
| Year of Transplant | 2009.0 [2004.0, 2013.0] | 2018.0 [2017.0, 2019.0] | 2021.0 [2020.0, 2022.0] |
| Duration on waiting list[a] — days | 91.0 [27.0, 256.0] | 73.0 [18.0, 257.0] | 31.0 [9.0, 150.0] |
| ABO blood group | | | |
| A | 13,459 (41.4) | 3,432 (40.1) | 3,564 (38.5) |
| B | 4,716 (14.5) | 1,314 (15.4) | 1,423 (15.4) |
| AB | 1,773 (5.4) | 453 (5.3) | 456 (4.9) |
| O | 12,595 (38.7) | 3,359 (39.2) | 3,812 (41.2) |
| History of Previous Transplant | 1,086 (3.3) | 268 (3.1) | 282 (3.0) |
| History of Prior Cardiac Surgery | 6,462 (25.6) | 1,841 (21.5) | 1,823 (19.7) |
| *Missing* | *7,337 (22.5)* | | *7 (0.1)* |
| **Recipient Medical Status at Transplant — no. (%)** | | | |
| Status at Transplant[a] | | | |
| ICU | 9,603 (29.5) | 3,370 (39.4) | 5,099 (55.1) |
| Hospitalized (not ICU) | 5,723 (17.6) | 1,299 (15.2) | 1,306 (14.1) |
| Not Hospitalized | 17,217 (52.9) | 3,889 (45.4) | 2,850 (30.8) |
| Mechanical circulatory support | 23,873 (73.4) | 7,184 (83.9) | 7,339 (79.3) |
| Ventilator Support[a,b] | 681 (2.1) | 119 (1.4) | 201 (2.2) |
| Ventricular assist device (VAD)[a,b] | | | |
| None | 20,066 (64.7) | 4,807 (56.2) | 6,051 (65.4) |
| LVAD | 8,274 (26.7) | 3,512 (41.0) | 2,986 (32.3) |
| RVAD | 64 (0.2) | 26 (0.3) | 40 (0.4) |
| BiVAD | 1,060 (3.4) | 213 (2.5) | 178 (1.9) |
| Unspecified | 1,569 (5.1) | 0 (0.0) | 0 (0.0) |
| *Missing* | *1,510 (4.6)* | — | — |
| Extracorporeal membrane oxygenation (ECMO)[a,b] | 184 (0.6) | 239 (2.8) | 560 (6.1) |
| Intra-aortic balloon pump[b] | 1,866 (5.7) | 1,419 (16.6) | 2,580 (27.9) |
| Intravenous inotropic support | 13,581 (41.7) | 3,151 (36.8) | 3,681 (39.8) |
| Dialysis dependence[a,b] | 1,165 (3.6) | 419 (4.9) | 528 (5.7) |
| **Recipient Comorbidities & Labs** | | | |
| Diabetes mellitus[a] | | | |
| No | 23,984 (74.2) | 6,083 (71.1) | 6,442 (69.6) |
| Type I | 616 (1.9) | 89 (1.0) | 106 (1.1) |
| Type II | 5,557 (17.2) | 2,285 (26.7) | 2,599 (28.1) |
| Other | 2,159 (6.7) | 97 (1.1) | 107 (1.2) |

*Continued on next page*

*Table S15 continued*

| Characteristic | Training Cohort (*n*=32,543) | Tuning Cohort (*n*=8,558) | Temporal Validation Cohort (*n*=9,255) |
|---|---|---|---|
| History of cerebrovascular disease | 1,561 (4.9) | 535 (6.3) | 716 (7.8) |
| *Missing* | *426 (1.3)* | *57 (0.7)* | *63 (0.7)* |
| History of smoking | 11,467 (47.5) | 3,710 (43.4) | 3,740 (40.4) |
| *Missing* | *8,423 (25.9)* | | |
| History of malignancy | 2,103 (6.5) | 766 (9.0) | 873 (9.4) |
| Implantable Cardioverter–Defibrillator (ICD)[a] | 21,263 (66.4) | 6,318 (74.6) | 6,246 (68.3) |
| *Missing* | *509 (1.6)* | *89 (1.0)* | *116 (1.3)* |
| Infection (within 2 weeks prior to transplant)[a,b] | 3,551 (11.3) | 818 (9.6) | 1,042 (11.3) |
| *Missing* | *1,046 (3.2)* | *20 (0.2)* | *39 (0.4)* |
| History of prior blood transfusions[a] | 7,189 (23.3) | 1,657 (19.4) | 1,452 (15.8) |
| *Missing* | *1,679 (5.2)* | *30 (0.4)* | *50 (0.5)* |
| Peak exercise oxygen consumption — mL/kg/min | 12.0 [9.0, 14.0] | 12.0 [9.5, 14.3] | 12.2 [9.9, 14.9] |
| *Missing* | *21,630 (66.5)* | *5,526 (64.6)* | *6,293 (68.0)* |
| Creatinine clearance[a,b] — mL/min | 76.6 [57.8, 99.8] | 78.6 [59.0, 102.3] | 78.5 [59.0, 103.6] |
| *Missing* | *937 (2.9)* | *261 (3.0)* | *361 (3.9)* |
| Total bilirubin[a,b] — mg/dL | 0.8 [0.5, 1.2] | 0.7 [0.4, 1.0] | 0.7 [0.5, 1.1] |
| *Missing* | *1,204 (3.7)* | *82 (1.0)* | *96 (1.0)* |
| Serum albumin[a] — g/dL | 3.7 [3.3, 4.1] | 3.8 [3.4, 4.2] | 3.8 [3.3, 4.1] |
| *Missing* | *8,435 (25.9)* | *4,880 (57.0)* | *202 (2.2)* |
| **Recipient Immunology** | | | |
| Calculated panel-reactive antibody (CPRA) — % | 0.0 [0.0, 5.0] | 0.0 [0.0, 7.0] | 0.0 [0.0, 7.0] |
| *Missing* | *28,722 (88.3)* | *1,883 (22.0)* | *2,355 (25.4)* |
| PRA category | | | |
| 0–10% | 2,998 (78.5) | 5,161 (77.3) | 5,373 (77.9) |
| >10–80% | 730 (19.1) | 1,322 (19.8) | 1,363 (19.8) |
| >80% | 93 (2.4) | 192 (2.9) | 164 (2.4) |
| *Missing* | *28,722 (88.3)* | *1,883 (22.0)* | *2,355 (25.4)* |
| Cytomegalovirus (CMV) seropositivity | | | |
| Negative | 11,710 (38.4) | 3,722 (43.6) | 4,078 (44.1) |
| Not Done | 41 (0.1) | 129 (1.5) | 88 (1.0) |
| Positive | 18,761 (61.5) | 4,693 (54.9) | 5,073 (54.9) |
| *Missing* | *2,031 (6.2)* | *14 (0.2)* | *16 (0.2)* |
| Hepatitis B core antibody seropositivity | | | |
| Negative | 26,109 (85.4) | 7,859 (92.1) | 8,556 (92.6) |
| Not Done | 3,074 (10.1) | 301 (3.5) | 241 (2.6) |
| Positive | 1,376 (4.5) | 374 (4.4) | 447 (4.8) |
| *Missing* | *1,984 (6.1)* | *24 (0.3)* | *11 (0.1)* |
| Hepatitis C virus (HCV) seropositivity[a] | | | |
| Negative | 29,364 (92.1) | 8,206 (96.1) | 8,913 (96.4) |
| Not Done | 1,872 (5.9) | 131 (1.5) | 119 (1.3) |
| Positive | 642 (2.0) | 202 (2.4) | 218 (2.4) |
| *Missing* | *665 (2.0)* | *19 (0.2)* | *5 (0.1)* |
| **Donor Characteristics** | | | |
| Age[a] — yr | 30.0 [22.0, 41.0] | 31.0 [24.0, 40.0] | 32.0 [25.0, 39.0] |
| Female sex | 9,423 (29.0) | 2,532 (29.6) | 2,505 (27.1) |
| Race or ethnic group | | | |
| White | 21,616 (66.4) | 5,546 (64.8) | 5,732 (61.9) |
| Black | 4,776 (14.7) | 1,330 (15.5) | 1,512 (16.3) |
| Hispanic/Latino | 5,327 (16.4) | 1,434 (16.8) | 1,724 (18.6) |
| Asian | 541 (1.7) | 138 (1.6) | 150 (1.6) |
| Native American | 167 (0.5) | 46 (0.5) | 79 (0.9) |
| Pacific Islander | 50 (0.2) | 19 (0.2) | 16 (0.2) |
| Multi-Racial | 62 (0.2) | 45 (0.5) | 42 (0.5) |
| Body-Mass Index — kg/m$^2$ | 25.8 [23.0, 29.5] | 26.5 [23.4, 30.5] | 26.8 [23.6, 31.0] |
| *Missing* | *378 (1.2)* | *156 (1.8)* | *195 (2.1)* |
| Weight[a] — kg | 79.4 [69.0, 91.0] | 80.5 [70.0, 93.5] | 81.7 [70.7, 95.0] |
| *Missing* | *415 (1.3)* | *163 (1.9)* | *219 (2.4)* |
| Height — cm | 175.3 [167.6, 182.0] | 175.0 [167.6, 180.3] | 175.0 [168.0, 180.3] |
| ABO blood group | | | |

*Continued on next page*

*Table S15 continued*

| Characteristic | Training Cohort (*n*=32,543) | Tuning Cohort (*n*=8,558) | Temporal Validation Cohort (*n*=9,255) |
|---|---|---|---|
| A | 11,799 (36.3) | 3,124 (36.5) | 3,207 (34.7) |
| B | 3,539 (10.9) | 933 (10.9) | 981 (10.6) |
| AB | 716 (2.2) | 178 (2.1) | 163 (1.8) |
| O | 16,488 (50.7) | 4,323 (50.5) | 4,904 (53.0) |
| **Donor Clinical History & Status** | | | |
| Cause of death | | | |
| Anoxia | 5,646 (17.4) | 3,417 (39.9) | 4,289 (46.3) |
| Cerebrovascular event | 7,514 (23.1) | 1,228 (14.3) | 1,114 (12.0) |
| Head trauma | 18,465 (56.8) | 3,694 (43.2) | 3,617 (39.1) |
| CNS tumor | 260 (0.8) | 37 (0.4) | 33 (0.4) |
| Other | 652 (2.0) | 182 (2.1) | 202 (2.2) |
| History of diabetes mellitus | 937 (2.9) | 312 (3.7) | 361 (4.0) |
| *Missing* | *125 (0.4)* | *41 (0.5)* | *135 (1.5)* |
| History of hypertension | 4,465 (13.8) | 1,349 (15.9) | 1,368 (15.0) |
| *Missing* | *180 (0.6)* | *47 (0.5)* | *146 (1.6)* |
| History of heavy alcohol use[a] | 5,457 (17.0) | 1,515 (18.2) | 1,781 (20.0) |
| *Missing* | *511 (1.6)* | *211 (2.5)* | *363 (3.9)* |
| History of cocaine use | 4,899 (15.4) | 2,305 (27.4) | 2,553 (28.4) |
| *Missing* | *632 (1.9)* | *136 (1.6)* | *254 (2.7)* |
| Inotropic support | 13,704 (42.2) | 3,210 (37.6) | 2,826 (30.6) |
| Left ventricular ejection fraction — % | 60.0 [55.0, 65.0] | 60.0 [56.0, 65.0] | 60.0 [56.0, 65.0] |
| *Missing* | *860 (2.6)* | *37 (0.4)* | *44 (0.5)* |
| Coronary angiography result | | | |
| No | 24,021 (74.3) | 5,126 (60.0) | 5,116 (55.3) |
| Yes, normal | 7,686 (23.8) | 3,155 (36.9) | 3,747 (40.5) |
| Yes, not normal | 633 (2.0) | 268 (3.1) | 390 (4.2) |
| Total ischemic time[a] — hr | 3.2 [2.5, 3.8] | 3.2 [2.5, 3.8] | 3.5 [2.9, 4.0] |
| *Missing* | *1,428 (4.4)* | *139 (1.6)* | *80 (0.9)* |
| **Donor Laboratory Values — median [IQR]** | | | |
| Serum creatinine — mg/dL | 1.0 [0.8, 1.3] | 1.0 [0.8, 1.5] | 1.0 [0.7, 1.5] |
| *Missing* | *266 (0.8)* | *118 (1.4)* | *166 (1.8)* |
| Blood urea nitrogen — mg/dL | 13.0 [9.0, 20.0] | 20.0 [13.0, 30.0] | 22.0 [15.0, 35.0] |
| *Missing* | *340 (1.0)* | *116 (1.4)* | *227 (2.5)* |
| Total bilirubin — mg/dL | 0.8 [0.5, 1.2] | 0.7 [0.5, 1.1] | 0.7 [0.4, 1.1] |
| *Missing* | *490 (1.5)* | *116 (1.4)* | *110 (1.2)* |
| **Donor Serology — no. (%)** | | | |
| Cytomegalovirus (CMV) seropositivity | | | |
| Indeterminate | 69 (0.2) | 39 (0.5) | 39 (0.4) |
| Negative | 12,456 (38.3) | 3,261 (38.1) | 3,476 (37.6) |
| Not Done | 23 (0.1) | 7 (0.1) | 5 (0.1) |
| Positive | 19,968 (61.4) | 5,248 (61.3) | 5,733 (62.0) |
| Hepatitis B core antibody seropositivity | | | |
| Negative | 31,730 (97.5) | 8,366 (97.8) | 9,009 (97.3) |
| Not Done | 86 (0.3) | 10 (0.1) | 0 (0.0) |
| Positive | 708 (2.2) | 182 (2.1) | 245 (2.6) |
| **Outcomes** | | | |
| All-cause mortality at 1 year | 3,451 (10.6) | 732 (8.6) | 859 (9.3) |

Values are presented as median [Q1, Q3] for continuous variables and no. (%) for categorical variables. Missingness is reported for variables with ≥1% missing data in any cohort. [a]Variable included in trained models (LR, MLP, RF, XGB; 22 predictors). [b]Variable included in the IMPACT scoring model (12 predictors). Variables marked [a,b] are shared between the two sets. BiVAD denotes biventricular assist device; IMPACT, Index for Mortality Prediction After Cardiac Transplantation; LR, logistic regression; LVAD, left ventricular assist device; MLP, multilayer perceptron; Q1, first quartile; Q3, third quartile; RF, random forest; RVAD, right ventricular assist device; VAD, ventricular assist device; and XGB, XGBoost.

Table S16. Calibration and accuracy: external validation cohort, *n*=9,255, related to Table 2

| Model | Version | No. of Variables or Terms | CITL (95% CI) | Calibration Slope (95% CI) | Observed:Expected Ratio (95% CI) | Brier (95% CI) |
|---|---|---|---|---|---|---|
| | Source | 12[v] | −0.535 (−0.608, −0.463) | 0.602 (0.502, 0.701) | 0.642 (0.603, 0.683) | 0.089 (0.085, 0.093) |
| | PRiSM | 9 | −0.170 (−0.241, −0.099) | 0.988 (0.837, 1.139) | 0.861 (0.809, 0.915) | 0.083 (0.079, 0.088) |
| IMPACT | Sparse PRiSM | 5 | −0.137 (−0.207, −0.067) | 1.360 (1.119, 1.600) | 0.885 (0.831, 0.941) | 0.083 (0.079, 0.088) |
| | PRN | 9 | −0.158 (−0.230, −0.087) | 0.847 (0.726, 0.968) | 0.872 (0.818, 0.926) | 0.083 (0.079, 0.088) |
| | Sparse PRN | 5 | −0.117 (−0.188, −0.046) | 0.964 (0.812, 1.115) | 0.902 (0.846, 0.958) | 0.083 (0.079, 0.088) |
| | Source | 22[v] | −0.226 (−0.297, −0.154) | 0.806 (0.685, 0.927) | 0.823 (0.774, 0.875) | 0.083 (0.079, 0.088) |
| | PRiSM | 21 | −0.198 (−0.270, −0.126) | 0.703 (0.600, 0.805) | 0.845 (0.795, 0.899) | 0.084 (0.080, 0.088) |
| LR | Sparse PRiSM | 9 | −0.146 (−0.216, −0.075) | 1.179 (0.988, 1.371) | 0.878 (0.826, 0.935) | 0.083 (0.079, 0.088) |
| | PRN | 19 | −0.182 (−0.254, −0.111) | 0.839 (0.729, 0.948) | 0.855 (0.804, 0.909) | 0.083 (0.079, 0.087) |
| | Sparse PRN | 9 | −0.151 (−0.223, −0.080) | 0.905 (0.770, 1.040) | 0.876 (0.824, 0.932) | 0.083 (0.079, 0.087) |
| | Source | 22[v] | −0.253 (−0.326, −0.181) | 0.739 (0.643, 0.834) | 0.808 (0.760, 0.859) | 0.084 (0.079, 0.088) |
| | PRiSM | 30 | −0.199 (−0.271, −0.127) | 0.784 (0.682, 0.885) | 0.845 (0.795, 0.898) | 0.083 (0.079, 0.087) |
| MLP | Sparse PRiSM | 10 | −0.120 (−0.190, −0.049) | 1.246 (1.073, 1.419) | 0.899 (0.845, 0.956) | 0.082 (0.078, 0.087) |
| | PRN | 22 | −0.173 (−0.245, −0.101) | 0.856 (0.746, 0.967) | 0.862 (0.810, 0.917) | 0.082 (0.078, 0.087) |
| | Sparse PRN | 10 | −0.131 (−0.203, −0.060) | 0.972 (0.842, 1.101) | 0.891 (0.839, 0.948) | 0.082 (0.078, 0.087) |
| | Source | 22[v] | −0.248 (−0.320, −0.176) | 0.730 (0.624, 0.836) | 0.810 (0.760, 0.862) | 0.084 (0.080, 0.089) |
| | PRiSM | 22 | −0.099 (−0.171, −0.028) | 0.891 (0.768, 1.014) | 0.917 (0.861, 0.975) | 0.083 (0.078, 0.087) |
| RF | Sparse PRiSM | 8 | −0.142 (−0.212, −0.071) | 0.977 (0.824, 1.129) | 0.883 (0.829, 0.938) | 0.083 (0.079, 0.088) |
| | PRN | 15 | −0.126 (−0.197, −0.054) | 0.977 (0.852, 1.101) | 0.896 (0.842, 0.953) | 0.082 (0.078, 0.087) |
| | Sparse PRN | 8 | −0.174 (−0.245, −0.103) | 0.907 (0.772, 1.041) | 0.859 (0.808, 0.913) | 0.083 (0.079, 0.088) |
| | Source | 22[v] | −0.156 (−0.228, −0.084) | 0.811 (0.705, 0.917) | 0.875 (0.824, 0.929) | 0.083 (0.079, 0.087) |
| | PRiSM | 26 | −0.158 (−0.231, −0.086) | 0.781 (0.678, 0.884) | 0.873 (0.822, 0.929) | 0.083 (0.079, 0.087) |
| XGB | Sparse PRiSM | 11 | −0.112 (−0.183, −0.041) | 1.053 (0.906, 1.200) | 0.906 (0.851, 0.962) | 0.082 (0.078, 0.087) |
| | PRN | 17 | −0.160 (−0.231, −0.088) | 0.906 (0.788, 1.023) | 0.871 (0.819, 0.926) | 0.082 (0.078, 0.087) |
| | Sparse PRN | 11 | −0.117 (−0.188, −0.046) | 1.019 (0.887, 1.151) | 0.902 (0.849, 0.959) | 0.082 (0.078, 0.087) |
| EBM | | 42 | −0.191 (−0.263, −0.119) | 0.759 (0.660, 0.858) | 0.851 (0.800, 0.904) | 0.083 (0.079, 0.088) |
| GAM | | 22 | −0.190 (−0.262, −0.118) | 0.747 (0.649, 0.845) | 0.852 (0.801, 0.905) | 0.083 (0.079, 0.088) |
| NAM | | 22 | −0.161 (−0.233, −0.089) | 0.707 (0.597, 0.818) | 0.871 (0.818, 0.926) | 0.084 (0.079, 0.088) |

CITL = calibration-in-the-large; ideal = 0; negative values indicate overprediction. Slope = calibration slope (ideal = 1); <1 indicates prediction compression. O:E = observed-to-expected ratio (ideal = 1). Brier Score: lower values indicate better overall accuracy; null model Brier (predicting the cohort prevalence for all patients) = 0.084. 95% CIs from logistic calibration model (CITL, Slope) and bootstrap percentile method (O:E, Brier; 1,000 resamples). [v] No. of variables or terms denotes input variables for source models, fitted terms for the comparators, and nomogram terms for all other rows. Comparator and nomogram terms are counted over the grouped clinical variables, so a categorical variable contributes one term rather than one per level. EBM = Explainable Boosting Machine; GAM = Generalized Additive Model (spline-based); NAM = Neural Additive Model. IMPACT denotes Index for Mortality Prediction After Cardiac Transplantation; LR, logistic regression; MLP, multilayer perceptron; PRiSM, Partial Responses in Structured Models; PRN, Partial Response Network; RF, random forest; and XGB, XGBoost.

Table S17. Comparison to additive baseline models: difference in AUROC vs EBM, external validation *n*=9,255, related to Table 3

| Model | Version | No. of Variables or Terms | AUROC (95% CI) | Difference in AUROC (95% CI) |
|---|---|---|---|---|
| EBM (ref) | | 42 | 0.647 (0.627, 0.667) | — |
| IMPACT | Source | 12[v] | 0.619 (0.599, 0.639) | −0.028 (−0.044, −0.013) |
| | PRiSM | 9 | 0.630 (0.610, 0.649) | −0.017 (−0.032, −0.003) |
| | Sparse PRiSM | 5 | 0.601 (0.581, 0.621) | −0.046 (−0.063, −0.029) |
| | PRN | 9 | 0.635 (0.616, 0.655) | −0.012 (−0.025, +0.001) |
| | Sparse PRN | 5 | 0.620 (0.600, 0.639) | −0.028 (−0.043, −0.011) |
| LR | Source | 22[v] | 0.620 (0.600, 0.640) | −0.027 (−0.037, −0.017) |
| | PRiSM | 21 | 0.625 (0.605, 0.645) | −0.022 (−0.031, −0.013) |
| | Sparse PRiSM | 9 | 0.611 (0.591, 0.631) | −0.036 (−0.050, −0.021) |
| | PRN | 19 | 0.645 (0.625, 0.665) | −0.002 (−0.008, +0.004) |
| | Sparse PRN | 9 | 0.625 (0.605, 0.645) | −0.022 (−0.035, −0.010) |
| MLP | Source | 22[v] | 0.645 (0.625, 0.665) | −0.002 (−0.009, +0.005) |
| | PRiSM | 30 | 0.646 (0.626, 0.666) | −0.001 (−0.006, +0.005) |
| | Sparse PRiSM | 10 | 0.636 (0.616, 0.655) | −0.011 (−0.023, +0.001) |
| | PRN | 22 | 0.646 (0.626, 0.665) | −0.002 (−0.007, +0.004) |
| | Sparse PRN | 10 | 0.641 (0.621, 0.661) | −0.006 (−0.016, +0.004) |
| RF | Source | 22[v] | 0.642 (0.622, 0.661) | −0.005 (−0.015, +0.004) |
| | PRiSM | 22 | 0.641 (0.621, 0.660) | −0.006 (−0.016, +0.003) |
| | Sparse PRiSM | 8 | 0.621 (0.601, 0.641) | −0.026 (−0.039, −0.013) |
| | PRN | 15 | 0.649 (0.629, 0.668) | +0.002 (−0.006, +0.009) |
| | Sparse PRN | 8 | 0.624 (0.604, 0.644) | −0.023 (−0.036, −0.011) |
| XGB | Source | 22[v] | 0.646 (0.627, 0.666) | −0.001 (−0.006, +0.005) |
| | PRiSM | 26 | 0.644 (0.624, 0.663) | −0.003 (−0.009, +0.003) |
| | Sparse PRiSM | 11 | 0.635 (0.615, 0.654) | −0.012 (−0.022, −0.002) |
| | PRN | 17 | 0.645 (0.626, 0.665) | −0.002 (−0.008, +0.004) |
| | Sparse PRN | 11 | 0.646 (0.627, 0.666) | −0.001 (−0.011, +0.009) |

Difference in AUROC = PRiSM model − EBM (AUROC 0.647). 95% CIs by DeLong method for correlated ROC curves. EBM = Explainable Boosting Machine. [v] No. of variables or terms denotes input variables for source models, fitted terms for the comparator, and nomogram terms for all other rows. Comparator and nomogram terms are counted over the grouped clinical variables, so a categorical variable contributes one term rather than one per level. PRiSM = Phase I only (ANOVA decomposition with LASSO selection); PRN = Phase I + Phase II (PRN refinement). Baseline: lambda preserving $\geq$99.8% of maximum tuning AUROC. Sparse: the sparse lambda, minimizing $n$-terms while targeting noninferiority on tuning AUROC; Sparse PRN applies PRN refinement to the sparse selection. AUROC denotes area under the receiver operating characteristic curve; IMPACT, Index for Mortality Prediction After Cardiac Transplantation; LR, logistic regression; MLP, multilayer perceptron; PRiSM, Partial Responses in Structured Models; PRN, Partial Response Network; RF, random forest; and XGB, XGBoost.

Table S18. Comparison to additive baseline models: difference in AUROC vs GAM, external validation *n*=9,255, related to Table 3

| Model | Version | No. of Variables or Terms | AUROC (95% CI) | Difference in AUROC (95% CI) |
|---|---|---|---|---|
| GAM (ref) | | 22 | 0.645 (0.625, 0.664) | — |
| IMPACT | Source | 12[v] | 0.619 (0.599, 0.639) | −0.026 (−0.042, −0.010) |
| | PRiSM | 9 | 0.630 (0.610, 0.649) | −0.015 (−0.030, +0.000) |
| | Sparse PRiSM | 5 | 0.601 (0.581, 0.621) | −0.044 (−0.061, −0.026) |
| | PRN | 9 | 0.635 (0.616, 0.655) | −0.010 (−0.023, +0.003) |
| | Sparse PRN | 5 | 0.620 (0.600, 0.639) | −0.025 (−0.041, −0.009) |
| LR | Source | 22[v] | 0.620 (0.600, 0.640) | −0.025 (−0.035, −0.016) |
| | PRiSM | 21 | 0.625 (0.605, 0.645) | −0.020 (−0.028, −0.012) |
| | Sparse PRiSM | 9 | 0.611 (0.591, 0.631) | −0.034 (−0.047, −0.020) |
| | PRN | 19 | 0.645 (0.625, 0.665) | +0.000 (−0.004, +0.004) |
| | Sparse PRN | 9 | 0.625 (0.605, 0.645) | −0.020 (−0.032, −0.008) |
| MLP | Source | 22[v] | 0.645 (0.625, 0.665) | +0.000 (−0.007, +0.007) |
| | PRiSM | 30 | 0.646 (0.626, 0.666) | +0.001 (−0.003, +0.005) |
| | Sparse PRiSM | 10 | 0.636 (0.616, 0.655) | −0.009 (−0.020, +0.002) |
| | PRN | 22 | 0.646 (0.626, 0.665) | +0.001 (−0.003, +0.004) |
| | Sparse PRN | 10 | 0.641 (0.621, 0.661) | −0.004 (−0.013, +0.005) |
| RF | Source | 22[v] | 0.642 (0.622, 0.661) | −0.003 (−0.014, +0.007) |
| | PRiSM | 22 | 0.641 (0.621, 0.660) | −0.004 (−0.014, +0.006) |
| | Sparse PRiSM | 8 | 0.621 (0.601, 0.641) | −0.024 (−0.037, −0.010) |
| | PRN | 15 | 0.649 (0.629, 0.668) | +0.004 (−0.003, +0.011) |
| | Sparse PRN | 8 | 0.624 (0.604, 0.644) | −0.021 (−0.033, −0.009) |
| XGB | Source | 22[v] | 0.646 (0.627, 0.666) | +0.002 (−0.005, +0.008) |
| | PRiSM | 26 | 0.644 (0.624, 0.663) | −0.001 (−0.007, +0.005) |
| | Sparse PRiSM | 11 | 0.635 (0.615, 0.654) | −0.010 (−0.020, +0.000) |
| | PRN | 17 | 0.645 (0.626, 0.665) | +0.000 (−0.004, +0.005) |
| | Sparse PRN | 11 | 0.646 (0.627, 0.666) | +0.001 (−0.007, +0.010) |

Difference in AUROC = PRiSM model − GAM (AUROC 0.645). 95% CIs by DeLong method for correlated ROC curves. GAM = Generalized Additive Model (spline-based). [v] No. of variables or terms denotes input variables for source models, fitted terms for the comparator, and nomogram terms for all other rows. Comparator and nomogram terms are counted over the grouped clinical variables, so a categorical variable contributes one term rather than one per level. PRiSM = Phase I only (ANOVA decomposition with LASSO selection); PRN = Phase I + Phase II (PRN refinement). Baseline: lambda preserving $\geq$99.8% of maximum tuning AUROC. Sparse: the sparse lambda, minimizing $n$-terms while targeting noninferiority on tuning AUROC; Sparse PRN applies PRN refinement to the sparse selection. AUROC denotes area under the receiver operating characteristic curve; IMPACT, Index for Mortality Prediction After Cardiac Transplantation; LR, logistic regression; MLP, multilayer perceptron; PRiSM, Partial Responses in Structured Models; PRN, Partial Response Network; RF, random forest; and XGB, XGBoost.

Table S19. Comparison to additive baseline models: difference in AUROC vs NAM, external validation *n*=9,255, related to Table 3

| Model | Version | No. of Variables or Terms | AUROC (95% CI) | Difference in AUROC (95% CI) |
|---|---|---|---|---|
| NAM (ref) | | 22 | 0.615 (0.595, 0.635) | — |
| IMPACT | Source | 12[v] | 0.619 (0.599, 0.639) | +0.004 (−0.014, +0.022) |
| | PRiSM | 9 | 0.630 (0.610, 0.649) | +0.015 (−0.003, +0.033) |
| | Sparse PRiSM | 5 | 0.601 (0.581, 0.621) | −0.014 (−0.032, +0.005) |
| | PRN | 9 | 0.635 (0.616, 0.655) | +0.021 (+0.004, +0.037) |
| | Sparse PRN | 5 | 0.620 (0.600, 0.639) | +0.005 (−0.013, +0.023) |
| LR | Source | 22[v] | 0.620 (0.600, 0.640) | +0.005 (−0.009, +0.019) |
| | PRiSM | 21 | 0.625 (0.605, 0.645) | +0.010 (−0.003, +0.023) |
| | Sparse PRiSM | 9 | 0.611 (0.591, 0.631) | −0.003 (−0.020, +0.014) |
| | PRN | 19 | 0.645 (0.625, 0.665) | +0.030 (+0.018, +0.043) |
| | Sparse PRN | 9 | 0.625 (0.605, 0.645) | +0.010 (−0.006, +0.027) |
| MLP | Source | 22[v] | 0.645 (0.625, 0.665) | +0.030 (+0.017, +0.044) |
| | PRiSM | 30 | 0.646 (0.626, 0.666) | +0.032 (+0.019, +0.044) |
| | Sparse PRiSM | 10 | 0.636 (0.616, 0.655) | +0.021 (+0.004, +0.038) |
| | PRN | 22 | 0.646 (0.626, 0.665) | +0.031 (+0.018, +0.044) |
| | Sparse PRN | 10 | 0.641 (0.621, 0.661) | +0.026 (+0.011, +0.042) |
| RF | Source | 22[v] | 0.642 (0.622, 0.661) | +0.027 (+0.013, +0.041) |
| | PRiSM | 22 | 0.641 (0.621, 0.660) | +0.026 (+0.012, +0.040) |
| | Sparse PRiSM | 8 | 0.621 (0.601, 0.641) | +0.007 (−0.010, +0.023) |
| | PRN | 15 | 0.649 (0.629, 0.668) | +0.034 (+0.020, +0.048) |
| | Sparse PRN | 8 | 0.624 (0.604, 0.644) | +0.009 (−0.006, +0.025) |
| XGB | Source | 22[v] | 0.646 (0.627, 0.666) | +0.032 (+0.020, +0.044) |
| | PRiSM | 26 | 0.644 (0.624, 0.663) | +0.029 (+0.017, +0.042) |
| | Sparse PRiSM | 11 | 0.635 (0.615, 0.654) | +0.020 (+0.005, +0.035) |
| | PRN | 17 | 0.645 (0.626, 0.665) | +0.031 (+0.018, +0.044) |
| | Sparse PRN | 11 | 0.646 (0.627, 0.666) | +0.032 (+0.016, +0.047) |

Difference in AUROC = PRiSM model − NAM (AUROC 0.615). 95% CIs by DeLong method for correlated ROC curves. NAM = Neural Additive Model. [v] No. of variables or terms denotes input variables for source models, fitted terms for the comparator, and nomogram terms for all other rows. Comparator and nomogram terms are counted over the grouped clinical variables, so a categorical variable contributes one term rather than one per level. PRiSM = Phase I only (ANOVA decomposition with LASSO selection); PRN = Phase I + Phase II (PRN refinement). Baseline: lambda preserving $\geq$99.8% of maximum tuning AUROC. Sparse: the sparse lambda, minimizing $n$-terms while targeting noninferiority on tuning AUROC; Sparse PRN applies PRN refinement to the sparse selection. AUROC denotes area under the receiver operating characteristic curve; IMPACT, Index for Mortality Prediction After Cardiac Transplantation; LR, logistic regression; MLP, multilayer perceptron; PRiSM, Partial Responses in Structured Models; PRN, Partial Response Network; RF, random forest; and XGB, XGBoost.

## Table S20. Seed stability summary across three random seeds (external validation), related to Table 2

| Model | Step | AUROC Range | No. of Terms | Jaccard | NI Result | Consistent |
|---|---|---|---|---|---|---|
| IMPACT | PRiSM | 0.002 | 9 | 1.00 | PASS | Yes |
| | PRN | 0.000 | 9 | 1.00 | PASS | Yes |
| | Sparse PRiSM | 0.002 | 5 | 1.00 | FAIL | Yes |
| | Sparse PRN | 0.001 | 5 | 1.00 | FAIL | Yes |
| LR | PRiSM | 0.000 | 21 | 1.00 | PASS | Yes |
| | PRN | 0.001 | 17–19 | 0.89 | PASS | Yes |
| | Sparse PRiSM | 0.001 | 9–10 | 0.90 | FAIL | Yes |
| | Sparse PRN | 0.002 | 9–10 | 0.90 | PASS | Yes |
| MLP | PRiSM | 0.000 | 30 | 1.00 | PASS | Yes |
| | PRN | 0.002 | 19–26 | 0.67 | PASS | Yes |
| | Sparse PRiSM | 0.000 | 10 | 1.00 | FAIL | Yes |
| | Sparse PRN | 0.001 | 10 | 1.00 | FAIL | Yes |
| RF | PRiSM | 0.000 | 22 | 1.00 | PASS | Yes |
| | PRN | 0.003 | 15–18 | 0.83 | PASS | Yes |
| | Sparse PRiSM | 0.000 | 8 | 1.00 | FAIL | Yes |
| | Sparse PRN | 0.001 | 8 | 1.00 | FAIL | Yes |
| XGB | PRiSM | 0.000 | 26 | 1.00 | PASS | Yes |
| | PRN | 0.002 | 17 | 1.00 | PASS | Yes |
| | Sparse PRiSM | 0.000 | 11 | 1.00 | FAIL | Yes |
| | Sparse PRN | 0.001 | 11 | 1.00 | PASS | Yes |

AUROC Range = maximum minus minimum AUROC point estimate across seeds 42, 123, and 257. No. of Terms = range of selected nomogram terms across seeds (single value if identical). Jaccard = Jaccard similarity coefficient of the selected term sets across all three seeds (1.00 = identical terms). NI Result = non-inferiority test outcome (identical across all seeds for every one of the 20 nomograms). Noninferiority was assessed on the retention-adjusted contrast $T = \mathrm{AUROC}_{\mathrm{new}} - 0.9 \times \mathrm{AUROC}_{\mathrm{source}} - 0.05$, and met when the lower bound of its 95% confidence interval exceeds 0. Consistent = whether NI conclusion is the same across all seeds and the 3-seed ensemble. AUROC denotes area under the receiver operating characteristic curve; IMPACT, Index for Mortality Prediction After Cardiac Transplantation; LR, logistic regression; MLP, multilayer perceptron; NI, noninferiority; PRiSM, Partial Responses in Structured Models; PRN, Partial Response Network; RF, random forest; and XGB, XGBoost.

## Table S21. Model performance on external validation (seed 42), related to Table 2

| Model | Version | No. of Variables or Terms | AUROC (95% CI) | Retention Contrast $T$ (95% CI) | Calibration Slope (95% CI) | CITL (95% CI) | Observed:Expected Ratio (95% CI) |
|---|---|---|---|---|---|---|---|
| IMPACT | Source | 12[v] | 0.619 (0.599, 0.639) | — | 0.602 (0.502, 0.701) | −0.535 (−0.608, −0.463) | 0.642 (0.603, 0.683) |
| | PRiSM | 9 | 0.628 (0.608, 0.647) | +0.021† (+0.009, +0.033) | 1.046 (0.885, 1.207) | −0.163 (−0.233, −0.092) | 0.866 (0.814, 0.921) |
| | Sparse PRiSM | 5 | 0.603 (0.583, 0.623) | −0.004 (−0.019, +0.012) | 1.360 (1.119, 1.600) | −0.137 (−0.207, −0.067) | 0.885 (0.831, 0.941) |
| | Sparse PRN | 5 | 0.619 (0.599, 0.639) | +0.012 (−0.004, +0.029) | 0.972 (0.819, 1.125) | −0.116 (−0.187, −0.045) | 0.903 (0.847, 0.959) |
| LR | Source | 22[v] | 0.620 (0.600, 0.640) | — | 0.806 (0.685, 0.927) | −0.226 (−0.297, −0.154) | 0.823 (0.774, 0.875) |
| | PRiSM | 21 | 0.625 (0.605, 0.645) | +0.017† (+0.012, +0.022) | 0.703 (0.600, 0.805) | −0.198 (−0.270, −0.126) | 0.845 (0.795, 0.899) |
| | Sparse PRiSM | 10 | 0.612 (0.592, 0.632) | +0.004 (−0.006, +0.015) | 1.183 (0.992, 1.375) | −0.144 (−0.215, −0.074) | 0.880 (0.827, 0.936) |
| | Sparse PRN | 10 | 0.627 (0.607, 0.646) | +0.019† (+0.007, +0.030) | 0.891 (0.760, 1.022) | −0.129 (−0.200, −0.058) | 0.893 (0.841, 0.951) |
| MLP | Source | 22[v] | 0.645 (0.625, 0.665) | — | 0.739 (0.643, 0.834) | −0.253 (−0.326, −0.181) | 0.808 (0.760, 0.859) |
| | PRiSM | 30 | 0.646 (0.626, 0.666) | +0.016† (+0.010, +0.021) | 0.784 (0.682, 0.885) | −0.199 (−0.271, −0.127) | 0.845 (0.795, 0.898) |
| | Sparse PRiSM | 10 | 0.636 (0.616, 0.655) | +0.005 (−0.007, +0.017) | 1.246 (1.073, 1.419) | −0.120 (−0.190, −0.049) | 0.899 (0.845, 0.956) |
| | Sparse PRN | 10 | 0.640 (0.621, 0.660) | +0.010 (−0.001, +0.021) | 0.970 (0.840, 1.099) | −0.130 (−0.201, −0.058) | 0.893 (0.840, 0.950) |
| RF | Source | 22[v] | 0.642 (0.622, 0.661) | — | 0.730 (0.624, 0.836) | −0.248 (−0.320, −0.176) | 0.810 (0.760, 0.862) |
| | PRiSM | 22 | 0.641 (0.621, 0.660) | +0.013† (+0.006, +0.020) | 0.891 (0.768, 1.014) | −0.099 (−0.171, −0.028) | 0.917 (0.861, 0.975) |
| | Sparse PRiSM | 8 | 0.621 (0.601, 0.641) | −0.006 (−0.016, +0.004) | 0.977 (0.824, 1.129) | −0.142 (−0.212, −0.071) | 0.883 (0.829, 0.938) |
| | Sparse PRN | 8 | 0.624 (0.604, 0.644) | −0.004 (−0.015, +0.008) | 0.916 (0.780, 1.052) | −0.179 (−0.250, −0.107) | 0.855 (0.805, 0.910) |
| XGB | Source | 22[v] | 0.646 (0.627, 0.666) | — | 0.811 (0.705, 0.917) | −0.156 (−0.228, −0.084) | 0.875 (0.824, 0.929) |
| | PRiSM | 26 | 0.644 (0.624, 0.663) | +0.012† (+0.008, +0.017) | 0.781 (0.678, 0.884) | −0.159 (−0.231, −0.086) | 0.873 (0.822, 0.929) |
| | Sparse PRiSM | 11 | 0.635 (0.615, 0.654) | +0.003 (−0.005, +0.012) | 1.053 (0.906, 1.200) | −0.112 (−0.183, −0.041) | 0.906 (0.851, 0.963) |
| | Sparse PRN | 11 | 0.645 (0.625, 0.664) | +0.013† (+0.004, +0.022) | 1.008 (0.877, 1.140) | −0.109 (−0.180, −0.038) | 0.909 (0.854, 0.966) |

The retention contrast is $T = \mathrm{AUROC}_{\mathrm{nomogram}} - 0.90 \times \mathrm{AUROC}_{\mathrm{source}} - 0.05$, which exceeds zero exactly when the nomogram retains more than 90% of its source model's discrimination above chance. †Noninferiority met, defined as the lower bound of the 95% confidence interval for $T$ exceeding 0. [v] No. of variables or terms denotes input variables for source models and nomogram terms for all other rows. 95% CIs for AUROC by DeLong method, and for $T$ from the same paired DeLong covariance matrix of the two AUROCs; calibration slope and CITL by Wald method; O:E ratio by bootstrap percentile method (1,000 resamples). A calibration slope of 1, CITL of 0, and an observed-to-expected ratio of 1 indicate perfect calibration. PRiSM = Phase I only (ANOVA decomposition with LASSO selection) (baseline lambda preserving $\geq$99.8% of maximum tuning AUROC). Sparse PRiSM = Phase I under the sparse lambda; the comparator each sparse PRN nomogram was refined from. Sparse PRN = Phase I + Phase II (PRN refinement) with the sparse lambda, minimizing $n$-terms while targeting noninferiority on tuning AUROC. AUROC denotes area under the receiver operating characteristic curve; CITL, calibration-in-the-large; IMPACT, Index for Mortality Prediction After Cardiac Transplantation; LR, logistic regression; MLP, multilayer perceptron; O:E, observed-to-expected; PRiSM, Partial Responses in Structured Models; PRN, Partial Response Network; RF, random forest; and XGB, XGBoost.

## Table S22. Model performance on external validation (seed 123), related to Table 2

| Model | Version | No. of Variables or Terms | AUROC (95% CI) | Retention Contrast $T$ (95% CI) | Calibration Slope (95% CI) | CITL (95% CI) | Observed:Expected Ratio (95% CI) |
|---|---|---|---|---|---|---|---|
| IMPACT | Source | 12[v] | 0.619 (0.599, 0.639) | — | 0.602 (0.502, 0.701) | −0.535 (−0.608, −0.463) | 0.642 (0.603, 0.683) |
| | PRiSM | 9 | 0.630 (0.610, 0.650) | +0.023[†] (+0.012, +0.035) | 0.988 (0.837, 1.139) | −0.170 (−0.241, −0.099) | 0.861 (0.809, 0.915) |
| | Sparse PRiSM | 5 | 0.602 (0.581, 0.622) | −0.005 (−0.021, +0.011) | 1.360 (1.119, 1.600) | −0.137 (−0.208, −0.067) | 0.884 (0.831, 0.941) |
| | Sparse PRN | 5 | 0.619 (0.599, 0.639) | +0.013 (−0.004, +0.029) | 0.962 (0.811, 1.113) | −0.116 (−0.187, −0.045) | 0.903 (0.847, 0.960) |
| LR | Source | 22[v] | 0.620 (0.600, 0.640) | — | 0.806 (0.685, 0.927) | −0.226 (−0.297, −0.154) | 0.823 (0.774, 0.875) |
| | PRiSM | 21 | 0.625 (0.605, 0.645) | +0.017[†] (+0.012, +0.022) | 0.703 (0.600, 0.805) | −0.198 (−0.270, −0.126) | 0.845 (0.795, 0.899) |
| | Sparse PRiSM | 9 | 0.611 (0.591, 0.631) | +0.004 (−0.008, +0.015) | 1.180 (0.988, 1.372) | −0.146 (−0.216, −0.075) | 0.878 (0.826, 0.935) |
| | Sparse PRN | 9 | 0.625 (0.606, 0.645) | +0.018[†] (+0.005, +0.030) | 0.906 (0.772, 1.041) | −0.142 (−0.213, −0.070) | 0.883 (0.832, 0.940) |
| MLP | Source | 22[v] | 0.645 (0.625, 0.665) | — | 0.739 (0.643, 0.834) | −0.253 (−0.326, −0.181) | 0.808 (0.760, 0.859) |
| | PRiSM | 30 | 0.646 (0.626, 0.666) | +0.016[†] (+0.010, +0.021) | 0.784 (0.682, 0.885) | −0.199 (−0.271, −0.127) | 0.845 (0.795, 0.898) |
| | Sparse PRiSM | 10 | 0.636 (0.616, 0.655) | +0.005 (−0.007, +0.017) | 1.246 (1.073, 1.419) | −0.120 (−0.190, −0.049) | 0.899 (0.845, 0.956) |
| | Sparse PRN | 10 | 0.641 (0.622, 0.661) | +0.011 (0.000, +0.021) | 0.974 (0.845, 1.104) | −0.130 (−0.201, −0.058) | 0.893 (0.840, 0.950) |
| RF | Source | 22[v] | 0.642 (0.622, 0.661) | — | 0.730 (0.624, 0.836) | −0.248 (−0.320, −0.176) | 0.810 (0.760, 0.862) |
| | PRiSM | 22 | 0.641 (0.621, 0.660) | +0.013[†] (+0.006, +0.020) | 0.891 (0.768, 1.014) | −0.099 (−0.171, −0.028) | 0.917 (0.861, 0.975) |
| | Sparse PRiSM | 8 | 0.621 (0.601, 0.641) | −0.006 (−0.016, +0.004) | 0.977 (0.825, 1.129) | −0.142 (−0.212, −0.071) | 0.883 (0.829, 0.938) |
| | Sparse PRN | 8 | 0.623 (0.603, 0.643) | −0.004 (−0.016, +0.007) | 0.902 (0.768, 1.037) | −0.176 (−0.247, −0.105) | 0.857 (0.807, 0.912) |
| XGB | Source | 22[v] | 0.646 (0.627, 0.666) | — | 0.811 (0.705, 0.917) | −0.156 (−0.228, −0.084) | 0.875 (0.824, 0.929) |
| | PRiSM | 26 | 0.644 (0.624, 0.663) | +0.012[†] (+0.008, +0.017) | 0.781 (0.678, 0.884) | −0.159 (−0.231, −0.086) | 0.873 (0.822, 0.928) |
| | Sparse PRiSM | 11 | 0.635 (0.615, 0.654) | +0.003 (−0.005, +0.012) | 1.053 (0.906, 1.200) | −0.112 (−0.183, −0.041) | 0.906 (0.851, 0.962) |
| | Sparse PRN | 11 | 0.645 (0.625, 0.665) | +0.013[†] (+0.004, +0.022) | 1.010 (0.879, 1.142) | −0.106 (−0.177, −0.035) | 0.911 (0.857, 0.969) |

The retention contrast is $T = \mathrm{AUROC}_{\mathrm{nomogram}} - 0.90 \times \mathrm{AUROC}_{\mathrm{source}} - 0.05$, which exceeds zero exactly when the nomogram retains more than 90% of its source model's discrimination above chance. [†]Noninferiority met, defined as the lower bound of the 95% confidence interval for $T$ exceeding 0. For the MLP Sparse PRN nomogram the lower bound does not exceed 0 but rounds to 0.000 at the three decimal places shown. [v] No. of variables or terms denotes input variables for source models and nomogram terms for all other rows. 95% CIs for AUROC by DeLong method, and for $T$ from the same paired DeLong covariance matrix of the two AUROCs; calibration slope and CITL by Wald method; O:E ratio by bootstrap percentile method (1,000 resamples). A calibration slope of 1, CITL of 0, and an observed-to-expected ratio of 1 indicate perfect calibration. PRiSM = Phase I only (ANOVA decomposition with LASSO selection) (baseline lambda preserving $\geq$99.8% of maximum tuning AUROC). Sparse PRiSM = Phase I under the sparse lambda; the comparator each sparse PRN nomogram was refined from. Sparse PRN = Phase I + Phase II (PRN refinement) with the sparse lambda, minimizing $n$-terms while targeting noninferiority on tuning AUROC. AUROC denotes area under the receiver operating characteristic curve; CITL, calibration-in-the-large; IMPACT, Index for Mortality Prediction After Cardiac Transplantation; LR, logistic regression; MLP, multilayer perceptron; O:E, observed-to-expected; PRiSM, Partial Responses in Structured Models; PRN, Partial Response Network; RF, random forest; and XGB, XGBoost.

## Table S23. Model performance on external validation (ensemble of 3 seeds), related to Table 2

| Model | Version | No. of Variables or Terms | AUROC (95% CI) | Retention Contrast $T$ (95% CI) | Calibration Slope (95% CI) | CITL (95% CI) | Observed:Expected Ratio (95% CI) |
|---|---|---|---|---|---|---|---|
| IMPACT | Source | — | 0.619 (0.599, 0.639) | — | 0.602 (0.502, 0.701) | −0.535 (−0.608, −0.463) | 0.642 (0.603, 0.683) |
| | PRiSM | — | 0.629 (0.609, 0.648) | +0.022[†] (+0.010, +0.034) | 1.007 (0.852, 1.161) | −0.168 (−0.238, −0.097) | 0.863 (0.811, 0.917) |
| | Sparse PRiSM | — | 0.603 (0.583, 0.623) | −0.004 (−0.020, +0.012) | 1.360 (1.119, 1.600) | −0.137 (−0.207, −0.067) | 0.885 (0.831, 0.941) |
| | Sparse PRN | — | 0.619 (0.599, 0.639) | +0.013 (−0.004, +0.029) | 0.967 (0.815, 1.119) | −0.116 (−0.187, −0.045) | 0.902 (0.847, 0.959) |
| LR | Source | — | 0.620 (0.600, 0.640) | — | 0.806 (0.685, 0.927) | −0.226 (−0.297, −0.154) | 0.823 (0.774, 0.875) |
| | PRiSM | — | 0.625 (0.605, 0.645) | +0.017[†] (+0.012, +0.022) | 0.703 (0.600, 0.805) | −0.198 (−0.270, −0.126) | 0.845 (0.795, 0.899) |
| | Sparse PRiSM | — | 0.612 (0.592, 0.631) | +0.004 (−0.007, +0.015) | 1.181 (0.989, 1.373) | −0.145 (−0.216, −0.075) | 0.879 (0.826, 0.935) |
| | Sparse PRN | — | 0.626 (0.606, 0.646) | +0.018[†] (+0.006, +0.030) | 0.904 (0.770, 1.038) | −0.141 (−0.212, −0.070) | 0.884 (0.832, 0.941) |
| MLP | Source | — | 0.645 (0.625, 0.665) | — | 0.739 (0.643, 0.834) | −0.253 (−0.326, −0.181) | 0.808 (0.760, 0.859) |
| | PRiSM | — | 0.646 (0.626, 0.666) | +0.016[†] (+0.010, +0.021) | 0.784 (0.682, 0.885) | −0.199 (−0.271, −0.127) | 0.845 (0.795, 0.898) |
| | Sparse PRiSM | — | 0.636 (0.616, 0.655) | +0.005 (−0.007, +0.017) | 1.246 (1.073, 1.419) | −0.120 (−0.190, −0.049) | 0.899 (0.845, 0.956) |
| | Sparse PRN | — | 0.641 (0.621, 0.661) | +0.010 (−0.001, +0.021) | 0.972 (0.843, 1.102) | −0.130 (−0.201, −0.059) | 0.892 (0.840, 0.949) |
| RF | Source | — | 0.642 (0.622, 0.661) | — | 0.730 (0.624, 0.836) | −0.248 (−0.320, −0.176) | 0.810 (0.760, 0.862) |
| | PRiSM | — | 0.641 (0.621, 0.660) | +0.013[†] (+0.006, +0.020) | 0.891 (0.768, 1.014) | −0.099 (−0.171, −0.028) | 0.917 (0.861, 0.975) |
| | Sparse PRiSM | — | 0.621 (0.601, 0.641) | −0.006 (−0.016, +0.004) | 0.977 (0.824, 1.129) | −0.142 (−0.212, −0.071) | 0.883 (0.829, 0.938) |
| | Sparse PRN | — | 0.624 (0.604, 0.643) | −0.004 (−0.015, +0.008) | 0.909 (0.774, 1.044) | −0.176 (−0.247, −0.105) | 0.857 (0.806, 0.912) |
| XGB | Source | — | 0.646 (0.627, 0.666) | — | 0.811 (0.705, 0.917) | −0.156 (−0.228, −0.084) | 0.875 (0.824, 0.929) |
| | PRiSM | — | 0.644 (0.624, 0.663) | +0.012[†] (+0.008, +0.017) | 0.781 (0.678, 0.884) | −0.159 (−0.231, −0.086) | 0.873 (0.822, 0.929) |
| | Sparse PRiSM | — | 0.635 (0.615, 0.654) | +0.003 (−0.005, +0.012) | 1.053 (0.906, 1.200) | −0.112 (−0.183, −0.041) | 0.906 (0.851, 0.962) |
| | Sparse PRN | — | 0.645 (0.626, 0.665) | +0.014[†] (+0.004, +0.023) | 1.014 (0.882, 1.145) | −0.111 (−0.182, −0.040) | 0.907 (0.853, 0.965) |

The retention contrast is $T = \mathrm{AUROC}_{\mathrm{nomogram}} - 0.90 \times \mathrm{AUROC}_{\mathrm{source}} - 0.05$, which exceeds zero exactly when the nomogram retains more than 90% of its source model's discrimination above chance. [†]Noninferiority met, defined as the lower bound of the 95% confidence interval for $T$ exceeding 0. [v] No. of variables or terms denotes input variables for source models and nomogram terms for all other rows. 95% CIs for AUROC by DeLong method, and for $T$ from the same paired DeLong covariance matrix of the two AUROCs; calibration slope and CITL by Wald method; O:E ratio by bootstrap percentile method (1,000 resamples). A calibration slope of 1, CITL of 0, and an observed-to-expected ratio of 1 indicate perfect calibration. PRiSM = Phase I only (ANOVA decomposition with LASSO selection) (baseline lambda preserving $\geq$99.8% of maximum tuning AUROC). Sparse PRiSM = Phase I under the sparse lambda; the comparator each sparse PRN nomogram was refined from. Sparse PRN = Phase I + Phase II (PRN refinement) with the sparse lambda, minimizing $n$-terms while targeting noninferiority on tuning AUROC. AUROC denotes area under the receiver operating characteristic curve; CITL, calibration-in-the-large; IMPACT, Index for Mortality Prediction After Cardiac Transplantation; LR, logistic regression; MLP, multilayer perceptron; O:E, observed-to-expected; PRiSM, Partial Responses in Structured Models; PRN, Partial Response Network; RF, random forest; and XGB, XGBoost.

Table S24. Effect of including transplant year on model performance (external validation), related to Table 2

| Model | Step | AUROC (95% CI) Without Tx Year | AUROC (95% CI) With Tx Year | Difference in No. of Terms | No-Yr NI | With-Yr NI | Agree |
|---|---|---|---|---|---|---|---|
| IMPACT | PRiSM | 0.630 (0.610, 0.649) | 0.630 (0.610, 0.649) | +0 | PASS | PASS | Yes |
| | PRN | 0.635 (0.616, 0.655) | 0.635 (0.616, 0.655) | +0 | PASS | PASS | Yes |
| | Sparse PRiSM | 0.601 (0.581, 0.621) | 0.601 (0.581, 0.621) | +0 | FAIL | FAIL | Yes |
| | Sparse PRN | 0.620 (0.600, 0.639) | 0.620 (0.600, 0.639) | +0 | FAIL | FAIL | Yes |
| LR | PRiSM | 0.625 (0.605, 0.645) | 0.632 (0.612, 0.651) | +0 | PASS | PASS | Yes |
| | PRN | 0.645 (0.625, 0.665) | 0.649 (0.629, 0.669) | −1 | PASS | PASS | Yes |
| | Sparse PRiSM | 0.611 (0.591, 0.631) | 0.623 (0.603, 0.642) | +2 | FAIL | FAIL | Yes |
| | Sparse PRN | 0.625 (0.605, 0.645) | 0.644 (0.624, 0.663) | +2 | PASS | PASS | Yes |
| MLP | PRiSM | 0.646 (0.626, 0.666) | 0.646 (0.627, 0.666) | −5 | PASS | PASS | Yes |
| | PRN | 0.646 (0.626, 0.665) | 0.644 (0.624, 0.664) | +1 | PASS | PASS | Yes |
| | Sparse PRiSM | 0.636 (0.616, 0.655) | 0.637 (0.618, 0.657) | +1 | FAIL | FAIL | Yes |
| | Sparse PRN | 0.641 (0.621, 0.661) | 0.643 (0.623, 0.663) | +1 | FAIL | PASS | **No** |
| RF | PRiSM | 0.641 (0.621, 0.660) | 0.640 (0.621, 0.660) | +2 | PASS | PASS | Yes |
| | PRN | 0.649 (0.629, 0.668) | 0.648 (0.628, 0.667) | −1 | PASS | PASS | Yes |
| | Sparse PRiSM | 0.621 (0.601, 0.641) | 0.622 (0.602, 0.641) | +1 | FAIL | FAIL | Yes |
| | Sparse PRN | 0.624 (0.604, 0.644) | 0.624 (0.604, 0.644) | +1 | FAIL | FAIL | Yes |
| XGB | PRiSM | 0.644 (0.624, 0.663) | 0.648 (0.628, 0.667) | +9 | PASS | PASS | Yes |
| | PRN | 0.645 (0.626, 0.665) | 0.651 (0.632, 0.671) | +6 | PASS | PASS | Yes |
| | Sparse PRiSM | 0.635 (0.615, 0.654) | 0.639 (0.620, 0.659) | +2 | FAIL | FAIL | Yes |
| | Sparse PRN | 0.646 (0.627, 0.666) | 0.646 (0.627, 0.666) | +2 | PASS | PASS | Yes |

AUROC 95% CIs by DeLong method. All confidence intervals overlap between configurations for every nomogram. Difference in No. of Terms = change in number of selected nomogram terms (with transplant year −without). NI = non-inferiority test result (PASS/FAIL), met when the lower bound of the 95% confidence interval for the retention-adjusted contrast $T = \mathrm{AUROC}_{\mathrm{new}} - 0.9 \times \mathrm{AUROC}_{\mathrm{source}} - 0.05$ exceeds 0. Each configuration is judged against its own source model. Agree = whether NI conclusion is the same with and without transplant year. AUROC denotes area under the receiver operating characteristic curve; IMPACT, Index for Mortality Prediction After Cardiac Transplantation; LR, logistic regression; MLP, multilayer perceptron; NI, noninferiority; PRiSM, Partial Responses in Structured Models; PRN, Partial Response Network; RF, random forest; Tx, transplant; and XGB, XGBoost.

Table S25. Performance of source models and translated nomograms with transplant year included in the temporal external-validation cohort, related to Table 2

| Model | Version | No. of Variables or Terms | AUROC (95% CI) | Retention Contrast $T$ (95% CI) | Calibration Slope (95% CI) | CITL (95% CI) | Observed:Expected Ratio (95% CI) |
|---|---|---|---|---|---|---|---|
| IMPACT | Source | 12[v] | 0.619 (0.599, 0.639) | — | 0.602 (0.502, 0.701) | −0.535 (−0.608, −0.463) | 0.642 (0.603, 0.683) |
| | PRiSM | 9 | 0.630 (0.610, 0.649) | +0.023[†] (+0.011, +0.034) | 0.988 (0.837, 1.139) | −0.170 (−0.241, −0.099) | 0.861 (0.809, 0.915) |
| | Sparse PRiSM | 5 | 0.601 (0.581, 0.621) | −0.006 (−0.021, +0.010) | 1.360 (1.119, 1.600) | −0.137 (−0.207, −0.067) | 0.885 (0.831, 0.941) |
| | Sparse PRN | 5 | 0.620 (0.600, 0.639) | +0.013 (−0.004, +0.029) | 0.964 (0.812, 1.115) | −0.117 (−0.188, −0.046) | 0.902 (0.846, 0.958) |
| LR | Source | 23[v] | 0.628 (0.608, 0.647) | — | 0.725 (0.621, 0.829) | 0.021 (−0.051, 0.093) | 1.018 (0.958, 1.084) |
| | PRiSM | 21 | 0.632 (0.612, 0.651) | +0.017[†] (+0.013, +0.021) | 0.710 (0.610, 0.811) | 0.146 (0.073, 0.218) | 1.134 (1.067, 1.208) |
| | Sparse PRiSM | 11 | 0.623 (0.603, 0.642) | +0.008 (−0.002, +0.018) | 1.215 (1.030, 1.400) | 0.019 (−0.052, 0.089) | 1.017 (0.956, 1.081) |
| | Sparse PRN | 11 | 0.644 (0.624, 0.663) | +0.029[†] (+0.018, +0.040) | 0.979 (0.850, 1.108) | 0.113 (0.041, 0.184) | 1.105 (1.039, 1.174) |
| MLP | Source | 23[v] | 0.639 (0.619, 0.659) | — | 0.699 (0.607, 0.792) | 0.059 (−0.013, 0.132) | 1.052 (0.988, 1.118) |
| | PRiSM | 25 | 0.646 (0.627, 0.666) | +0.021[†] (+0.013, +0.029) | 0.794 (0.691, 0.897) | 0.031 (−0.041, 0.103) | 1.027 (0.967, 1.093) |
| | Sparse PRiSM | 11 | 0.637 (0.618, 0.657) | +0.012 (−0.001, +0.025) | 1.245 (1.073, 1.417) | 0.031 (−0.040, 0.102) | 1.028 (0.967, 1.092) |
| | Sparse PRN | 11 | 0.643 (0.623, 0.663) | +0.018[†] (+0.006, +0.029) | 0.996 (0.864, 1.127) | 0.115 (0.044, 0.186) | 1.108 (1.042, 1.177) |
| RF | Source | 23[v] | 0.642 (0.623, 0.662) | — | 0.760 (0.647, 0.872) | −0.168 (−0.240, −0.096) | 0.866 (0.813, 0.921) |
| | PRiSM | 24 | 0.640 (0.621, 0.660) | +0.012[†] (+0.005, +0.019) | 0.924 (0.797, 1.051) | 0.021 (−0.050, 0.092) | 1.019 (0.957, 1.083) |
| | Sparse PRiSM | 9 | 0.622 (0.602, 0.641) | −0.006 (−0.017, +0.004) | 0.955 (0.807, 1.102) | −0.116 (−0.187, −0.045) | 0.903 (0.849, 0.960) |
| | Sparse PRN | 9 | 0.624 (0.604, 0.644) | −0.004 (−0.016, +0.008) | 0.974 (0.829, 1.118) | −0.005 (−0.076, 0.066) | 0.996 (0.937, 1.060) |
| XGB | Source | 23[v] | 0.647 (0.628, 0.667) | — | 0.804 (0.701, 0.908) | 0.006 (−0.066, 0.078) | 1.005 (0.945, 1.068) |
| | PRiSM | 35 | 0.648 (0.628, 0.667) | +0.015[†] (+0.009, +0.021) | 0.801 (0.697, 0.905) | 0.021 (−0.051, 0.093) | 1.018 (0.958, 1.083) |
| | Sparse PRiSM | 13 | 0.639 (0.620, 0.659) | +0.007 (−0.003, +0.016) | 1.097 (0.948, 1.247) | 0.018 (−0.053, 0.089) | 1.016 (0.954, 1.079) |
| | Sparse PRN | 13 | 0.646 (0.627, 0.666) | +0.013[†] (+0.003, +0.024) | 1.027 (0.893, 1.161) | 0.147 (0.076, 0.219) | 1.140 (1.073, 1.212) |

The retention contrast is $T = \mathrm{AUROC}_{\mathrm{nomogram}} - 0.90 \times \mathrm{AUROC}_{\mathrm{source}} - 0.05$, which exceeds zero exactly when the nomogram retains more than 90% of its source model's discrimination above chance. [†]Noninferiority met, defined as the lower bound of the 95% confidence interval for $T$ exceeding 0. [v] No. of variables or terms denotes input variables for source models and nomogram terms for all other rows. 95% CIs for AUROC by DeLong method, and for $T$ from the same paired DeLong covariance matrix of the two AUROCs; calibration slope and CITL by Wald method; O:E ratio by bootstrap percentile method (1,000 resamples). A calibration slope of 1, CITL of 0, and an observed-to-expected ratio of 1 indicate perfect calibration. PRiSM = Phase I only (ANOVA decomposition with LASSO selection) (baseline lambda preserving $\geq$99.8% of maximum tuning AUROC). Sparse PRiSM = Phase I under the sparse lambda; the comparator each sparse PRN nomogram was refined from. Sparse PRN = Phase I + Phase II (PRN refinement) with the sparse lambda, minimizing $n$-terms while targeting noninferiority on tuning AUROC. AUROC denotes area under the receiver operating characteristic curve; CITL, calibration-in-the-large; IMPACT, Index for Mortality Prediction After Cardiac Transplantation; LR, logistic regression; MLP, multilayer perceptron; O:E, observed-to-expected; PRiSM, Partial Responses in Structured Models; PRN, Partial Response Network; RF, random forest; and XGB, XGBoost.

Table S26. Discrimination by recipient sex and race or ethnic group: external validation cohort, *n*=9,255

| Model | Version | White *n*=5,414 496 deaths | Black *n*=2,409 222 deaths | Hispanic *n*=968 87 deaths | Other *n*=464 54 deaths | Male *n*=6,820 628 deaths | Female *n*=2,435 231 deaths |
|---|---|---|---|---|---|---|---|
| IMPACT | Source | 0.632 (0.605, 0.658) | 0.601 (0.562, 0.640) | 0.614 (0.550, 0.677) | 0.637 (0.559, 0.714) | 0.622 (0.598, 0.645) | 0.621 (0.583, 0.659) |
| | PRiSM | 0.642 (0.617, 0.668) | 0.612 (0.573, 0.650) | 0.594 (0.528, 0.659) | 0.646 (0.567, 0.725) | 0.630 (0.607, 0.654) | 0.633 (0.597, 0.670) |
| | Sparse PRiSM | 0.613 (0.587, 0.640) | 0.574 (0.536, 0.613) | 0.587 (0.524, 0.649) | 0.629 (0.548, 0.710) | 0.602 (0.578, 0.625) | 0.600 (0.561, 0.638) |
| | PRN | 0.646 (0.621, 0.672) | 0.611 (0.573, 0.650) | 0.612 (0.547, 0.676) | 0.661 (0.587, 0.736) | 0.637 (0.614, 0.660) | 0.632 (0.595, 0.669) |
| | Sparse PRN | 0.630 (0.604, 0.656) | 0.586 (0.546, 0.625) | 0.625 (0.563, 0.686) | 0.648 (0.568, 0.729) | 0.621 (0.598, 0.645) | 0.614 (0.576, 0.652) |
| LR | Source | 0.636 (0.610, 0.662) | 0.586 (0.546, 0.626) | 0.583 (0.518, 0.649) | 0.665 (0.582, 0.748) | 0.621 (0.598, 0.645) | 0.613 (0.575, 0.652) |
| | PRiSM | 0.643 (0.617, 0.668) | 0.585 (0.546, 0.624) | 0.596 (0.531, 0.661) | 0.667 (0.585, 0.749) | 0.628 (0.604, 0.651) | 0.615 (0.577, 0.653) |
| | Sparse PRiSM | 0.622 (0.596, 0.648) | 0.581 (0.541, 0.621) | 0.598 (0.533, 0.662) | 0.659 (0.578, 0.740) | 0.614 (0.591, 0.637) | 0.604 (0.566, 0.643) |
| | PRN | 0.660 (0.634, 0.685) | 0.616 (0.577, 0.656) | 0.616 (0.551, 0.681) | 0.664 (0.583, 0.745) | 0.646 (0.623, 0.669) | 0.640 (0.603, 0.678) |
| | Sparse PRN | 0.637 (0.611, 0.663) | 0.603 (0.564, 0.642) | 0.594 (0.527, 0.660) | 0.659 (0.578, 0.739) | 0.625 (0.602, 0.648) | 0.624 (0.586, 0.662) |
| MLP | Source | 0.657 (0.631, 0.683) | 0.624 (0.584, 0.663) | 0.611 (0.544, 0.677) | 0.672 (0.593, 0.752) | 0.649 (0.625, 0.672) | 0.635 (0.596, 0.673) |
| | PRiSM | 0.662 (0.636, 0.687) | 0.615 (0.576, 0.654) | 0.620 (0.555, 0.685) | 0.663 (0.582, 0.745) | 0.649 (0.626, 0.673) | 0.636 (0.598, 0.673) |
| | Sparse PRiSM | 0.647 (0.622, 0.673) | 0.607 (0.568, 0.646) | 0.622 (0.558, 0.685) | 0.667 (0.588, 0.746) | 0.637 (0.614, 0.660) | 0.632 (0.594, 0.670) |
| | PRN | 0.660 (0.634, 0.685) | 0.616 (0.576, 0.655) | 0.627 (0.562, 0.691) | 0.663 (0.582, 0.744) | 0.647 (0.624, 0.670) | 0.640 (0.603, 0.678) |
| | Sparse PRN | 0.654 (0.629, 0.680) | 0.609 (0.570, 0.648) | 0.630 (0.567, 0.694) | 0.662 (0.583, 0.742) | 0.642 (0.619, 0.665) | 0.639 (0.601, 0.677) |
| RF | Source | 0.654 (0.628, 0.679) | 0.609 (0.571, 0.646) | 0.633 (0.570, 0.695) | 0.676 (0.596, 0.756) | 0.641 (0.618, 0.664) | 0.643 (0.607, 0.679) |
| | PRiSM | 0.655 (0.629, 0.680) | 0.609 (0.571, 0.648) | 0.619 (0.556, 0.681) | 0.676 (0.599, 0.753) | 0.640 (0.617, 0.664) | 0.641 (0.605, 0.678) |
| | Sparse PRiSM | 0.632 (0.606, 0.658) | 0.591 (0.553, 0.630) | 0.605 (0.543, 0.668) | 0.666 (0.587, 0.745) | 0.619 (0.595, 0.642) | 0.627 (0.590, 0.663) |
| | PRN | 0.662 (0.637, 0.688) | 0.621 (0.582, 0.661) | 0.621 (0.558, 0.684) | 0.674 (0.596, 0.751) | 0.650 (0.627, 0.673) | 0.644 (0.607, 0.681) |
| | Sparse PRN | 0.636 (0.610, 0.662) | 0.598 (0.558, 0.637) | 0.597 (0.532, 0.663) | 0.661 (0.579, 0.743) | 0.624 (0.601, 0.648) | 0.622 (0.584, 0.660) |
| XGB | Source | 0.660 (0.635, 0.685) | 0.621 (0.583, 0.658) | 0.628 (0.565, 0.690) | 0.653 (0.571, 0.736) | 0.645 (0.622, 0.668) | 0.648 (0.612, 0.685) |
| | PRiSM | 0.658 (0.633, 0.684) | 0.615 (0.577, 0.653) | 0.630 (0.568, 0.693) | 0.647 (0.563, 0.730) | 0.645 (0.622, 0.668) | 0.641 (0.604, 0.678) |
| | Sparse PRiSM | 0.648 (0.623, 0.674) | 0.604 (0.566, 0.643) | 0.625 (0.564, 0.685) | 0.649 (0.567, 0.730) | 0.635 (0.612, 0.658) | 0.634 (0.597, 0.671) |
| | PRN | 0.660 (0.634, 0.685) | 0.615 (0.576, 0.654) | 0.627 (0.563, 0.691) | 0.661 (0.580, 0.742) | 0.645 (0.622, 0.669) | 0.644 (0.606, 0.681) |
| | Sparse PRN | 0.660 (0.634, 0.685) | 0.618 (0.578, 0.657) | 0.627 (0.564, 0.690) | 0.666 (0.588, 0.745) | 0.647 (0.624, 0.670) | 0.644 (0.606, 0.681) |
| EBM | | 0.661 (0.636, 0.687) | 0.619 (0.580, 0.657) | 0.619 (0.556, 0.683) | 0.667 (0.588, 0.746) | 0.647 (0.623, 0.670) | 0.646 (0.609, 0.682) |
| GAM | | 0.660 (0.635, 0.686) | 0.615 (0.575, 0.654) | 0.620 (0.556, 0.684) | 0.654 (0.573, 0.734) | 0.645 (0.622, 0.668) | 0.643 (0.606, 0.680) |
| NAM | | 0.636 (0.610, 0.662) | 0.568 (0.528, 0.607) | 0.601 (0.536, 0.666) | 0.623 (0.540, 0.706) | 0.614 (0.590, 0.638) | 0.614 (0.577, 0.651) |

Values are AUROC (95% CI) in the temporal external-validation cohort, by DeLong method. Models were evaluated without recalibration. Neither recipient sex nor race or ethnic group is among the 22 predictors, except in the IMPACT score, whose published definition includes both. Race or ethnic group follows the categories of the baseline characteristics table; Other comprises Asian, American Indian/Alaska Native, Native Hawaiian/Pacific Islander and multiracial recipients, and is underpowered at 464 patients and 54 deaths. Subgroup analyses were exploratory and were not adjusted for multiplicity. PRiSM = Phase I only (ANOVA decomposition with LASSO selection); PRN = Phase I + Phase II (PRN refinement). Baseline: lambda preserving $\geq$99.8% of maximum tuning AUROC. Sparse: the sparse lambda. AUROC denotes area under the receiver operating characteristic curve; EBM, explainable boosting machine; GAM, spline-based generalized additive model; IMPACT, Index for Mortality Prediction After Cardiac Transplantation; LR, logistic regression; MLP, multilayer perceptron; NAM, neural additive model; PRiSM, Partial Responses in Structured Models; PRN, Partial Response Network; RF, random forest; and XGB, XGBoost.

## S4 Supplementary References

[S1] G. S. Collins, K. G. M. Moons, P. Dhiman, et al. “TRIPOD+AI Statement: Updated Guidance for Reporting Clinical Prediction Models That Use Regression or Machine Learning Methods”. In: *BMJ* 385 (Apr. 2024), e078378. ISSN: 1756-1833. DOI: 10.1136/bmj-2023-078378.

[S2] D. J. Stekhoven and P. Bühlmann. “MissForest—Non-Parametric Missing Value Imputation for Mixed-Type Data”. In: *Bioinformatics* 28.1 (Jan. 2012), pp. 112–118. ISSN: 1367-4803. DOI: 10.1093/bioinformatics/btr597. (Visited on 02/27/2026).

[S3] E. S. Weiss, J. G. Allen, G. J. Arnaoutakis, et al. “Creation of a Quantitative Recipient Risk Index for Mortality Prediction After Cardiac Transplantation (IMPACT)”. In: *The Annals of Thoracic Surgery* 92.3 (Sept. 2011), pp. 914–922. ISSN: 0003-4975. DOI: 10.1016/j.athoracsur.2011.04.030. (Visited on 03/25/2024).

[S4] D. W. Hosmer, S. Lemeshow, and R. X. Sturdivant. *Applied Logistic Regression*. 3rd ed. John Wiley & Sons, Feb. 2013, p. 528. ISBN: 978-1-118-54835-6.

[S5] D. E. Rumelhart, G. E. Hinton, and R. J. Williams. “Learning Representations by Back-Propagating Errors”. In: *Nature* 323.6088 (Oct. 1986), pp. 533–536. ISSN: 1476-4687. DOI: 10.1038/323533a0. (Visited on 02/27/2026).

[S6] L. Breiman. “Random Forests”. In: *Machine Learning* 45.1 (Oct. 2001), pp. 5–32. ISSN: 1573-0565. DOI: 10.1023/A:1010933404324. (Visited on 02/27/2026).

[S7] T. Chen and C. Guestrin. “XGBoost: A Scalable Tree Boosting System”. In: *Proceedings of the 22nd ACM SIGKDD International Conference on Knowledge Discovery and Data Mining*. KDD ’16. New York, NY, USA: Association for Computing Machinery, Aug. 2016, pp. 785–794. ISBN: 978-1-4503-4232-2. DOI: 10.1145/2939672.2939785. (Visited on 02/27/2026).

[S8] T. Akiba, S. Sano, T. Yanase, T. Ohta, and M. Koyama. “Optuna: A Next-generation Hyperparameter Optimization Framework”. In: *Proceedings of the 25th ACM SIGKDD International Conference on Knowledge Discovery & Data Mining*. KDD ’19. New York, NY, USA: Association for Computing Machinery, July 2019, pp. 2623–2631. ISBN: 978-1-4503-6201-6. DOI: 10.1145/3292500.3330701. (Visited on 02/27/2026).

[S9] T. J. Hastie and R. J. Tibshirani. *Generalized Additive Models*. CRC Press, June 1990. ISBN: 978-0-412-34390-2.

[S10] B. Walters, S. Ortega-Martorell, I. Olier, and P. J. G. Lisboa. “How to Open a Black Box Classifier for Tabular Data”. In: *Algorithms* 16.4 (Apr. 2023), p. 181. ISSN: 1999-4893. DOI: 10.3390/a16040181. (Visited on 12/02/2024).

[S11] R. Tibshirani. “Regression Shrinkage and Selection Via the Lasso”. In: *Journal of the Royal Statistical Society: Series B (Methodological)* 58.1 (Jan. 1996), pp. 267–288. ISSN: 0035-9246. DOI: 10.1111/j.2517-6161.1996.tb02080.x. (Visited on 02/27/2026).

[S12] X. Robin, N. Turck, A. Hainard, et al. “pROC: An Open-Source Package for R and S+ to Analyze and Compare ROC Curves”. In: *BMC Bioinformatics* 12.1 (Mar. 2011), p. 77. ISSN: 1471-2105. DOI: 10.1186/1471-2105-12-77. (Visited on 02/27/2026).

[S13] A. J. Vickers and E. B. Elkin. “Decision Curve Analysis: A Novel Method for Evaluating Prediction Models”. In: *Medical Decision Making* 26.6 (Nov. 2006), pp. 565–574. ISSN: 0272-989X. DOI: 10.1177/0272989X06295361. (Visited on 02/27/2026).

[S14] H. Pigot, P. Lisboa, and J. Nilsson. *PRiSM (Partial Responses in Structured Models): A Python Framework for Converting Probabilistic Binary Classifiers into Auditable Nomograms*. Version 0.1.1. 2026. DOI: 10.5281/zenodo.19632957.

[S15] R. Agarwal, L. Melnick, N. Frosst, et al. “Neural Additive Models: Interpretable Machine Learning with Neural Nets”. In: *Advances in Neural Information Processing Systems 34 (NeurIPS 2021)*. Vol. 34. Curran Associates, Inc., 2021, pp. 4699–4711. (Visited on 02/27/2026).

[S16] Y. Lou, R. Caruana, J. Gehrke, and G. Hooker. “Accurate Intelligible Models with Pairwise Interactions”. In: *Proceedings of the 19th ACM SIGKDD International Conference on Knowledge Discovery and Data Mining*. KDD ’13. New York, NY, USA: Association for Computing Machinery, Aug. 2013, pp. 623–631. ISBN: 978-1-4503-2174-7. DOI: 10.1145/2487575.2487579. (Visited on 02/27/2026).

[S17] H. Nori, S. Jenkins, P. Koch, and R. Caruana. *InterpretML: A Unified Framework for Machine Learning Interpretability*. Sept. 2019. DOI: 10.48550/arXiv.1909.09223. arXiv: 1909.09223. (Visited on 12/02/2024).

[S18] S. N. Wood. *Generalized Additive Models: An Introduction with R, Second Edition*. 2nd ed. New York: Chapman and Hall/CRC, May 2017. ISBN: 978-1-315-37027-9. DOI: 10.1201/9781315370279.